\documentclass{article}

\usepackage{latexsym}
\usepackage{graphicx}
\usepackage{graphics}
\usepackage{subfigure}
\usepackage{algorithm}
\usepackage{hyperref}

\usepackage{amsfonts}
\usepackage{amsmath}
\usepackage{amstext}
\usepackage{latexsym}
\usepackage{amssymb}
\usepackage{url}
\usepackage{color}
\usepackage{wrapfig}
\usepackage{rotating}
\usepackage{fullname}

\newcommand{\ignore}[1]{ }

\newfont{\msym}{msbm10}
\newcommand{\reals}{\mathbb{R}}
\newcommand{\half}{\frac{1}{2}}

\newcommand{\paren}[1]{\left({#1}\right)}

\newcommand{\braces}[1]{\left\{{#1}\right\}}

\newcommand{\pr}[1]{{\rm Pr}\left[{#1}\right]}

\newcommand{\comdots}{, \ldots ,}

\newcommand{\mcal}[1]{{\mathcal{#1}}}

\newcommand{\KL}{{\textrm{D}_{\textrm{KL}}}}

\newcommand{\nolineskips}{
\setlength{\parskip}{0pt}
\setlength{\parsep}{0pt}
\setlength{\topsep}{0pt}
\setlength{\partopsep}{0pt}
\setlength{\itemsep}{0pt}}

\newcommand{\beq}[1]{\begin{equation}\label{#1}}
\newcommand{\eeq}{\end{equation}}
\newcommand{\beqa}{\begin{eqnarray}}
\newcommand{\eeqa}{\end{eqnarray}}
\renewcommand{\eqref}[1]{Eq.~(\ref{#1})}
\newcommand{\secref}[1]{Sec.~\ref{#1}}
\newcommand{\figref}[1]{Fig.~\ref{#1}}

\newcommand{\tabref}[1]{Table~\ref{#1}}

\newcommand{\norm}{\mcal{N}}

\newcommand{\algref}[1]{Alg.~\ref{#1}}
\newcommand{\mb}[1]{{\boldsymbol{#1}}}

\newcommand{\vx}{\mb{x}}
\newcommand{\vxi}[1]{\vx_{#1}}
\newcommand{\vxii}{\vxi{i}}

\newcommand{\yi}[1]{y_{#1}}

\newcommand{\hyi}[1]{\hat{y}_{#1}}

\newcommand{\vy}{\mb{y}}
\newcommand{\vyi}[1]{\vy_{#1}}
\newcommand{\vyii}{\vyi{i}}

\newcommand{\hvy}{\hat{\vy}}
\newcommand{\hvyi}[1]{\hvy_{#1}}
\newcommand{\hvyii}{\hvyi{i}}

\newcommand{\vmu}{\mb{\mu}}

\newcommand{\vmui}[1]{\vmu_{#1}}
\newcommand{\vmuii}{\vmui{i}}

\newcommand{\msigma}{\Sigma}

\newcommand{\msigmai}[1]{\msigma_{#1}}
\newcommand{\msigmaii}{\msigmai{i}}

\newcommand{\vw}{\mb{w}}
\newcommand{\vwi}[1]{\vw_{#1}}
\newcommand{\vwii}{\vwi{i}}

\newcommand{\mi}{I}

\newcommand{\X}{\mcal{X}}
\newcommand{\Y}{\mcal{Y}}

\newcommand{\alphai}[1]{\alpha_{#1}}
\newcommand{\alphaii}{\alphai{i}}

\newcommand{\eps}{\epsilon}

\newcommand{\newstufffrom}[1]{}

\newcommand{\oldnote}[2]{}

\newcommand{\commentout}[1]{}

\newcounter {mySubCounter}
\newcommand {\twocoleqn}[4]{
  \setcounter {mySubCounter}{0} %
  \let\OldTheEquation \theequation %
  \renewcommand {\theequation }{\OldTheEquation \alph {mySubCounter}}%
  \noindent \hfill%
  \begin{minipage}{.40\textwidth}
\vspace{-0.6cm}
    \begin{equation}\refstepcounter{mySubCounter}
      #1
    \end {equation}
  \end {minipage}
~~~~~~
  \addtocounter {equation}{ -1}%
  \begin{minipage}{.40\textwidth}
\vspace{-0.6cm}
    \begin{equation}\refstepcounter{mySubCounter}
      #3
    \end{equation}
  \end{minipage}%
  \let\theequation\OldTheEquation
}

\newcommand{\vzero}{\mb{0}}

\newcommand{\betai}[1]{\beta_{#1}}
\newcommand{\betaii}{\betai{i}}

\newcommand{\varb}{v}
\newcommand{\varbi}[1]{\varb_{#1}}
\newcommand{\varbii}{\varbi{i}}

\newcommand{\vara}{v^+}
\newcommand{\varai}[1]{\vara_{#1}}
\newcommand{\varaii}{\varai{i}}

\newcommand{\marb}{m}
\newcommand{\marbi}[1]{\marb_{#1}}
\newcommand{\marbii}{\marbi{i}}

\newcommand{\phia}{\phi'}
\newcommand{\phib}{\phi''}

\newcommand{\vz}{\mb{z}}
\newcommand{\vzi}[1]{\vz^{(#1)}}
\newcommand{\vzii}{\vzi{i}}

\newcommand{\fun}{\mb \Phi}

\newcommand{\gun}{\mb \Delta}

\newcommand{\guniba}{\gun_{i,\vy,\vz}}
\newcommand{\gunibp}{\gun_{i,\vy,\hvy}}

\title{Confidence Estimation in Structured Prediction}

\author{Avihai Mejer\\
  Department of Computer Science\\
  Technion-Israel Institute of Technology\\
  Haifa 32000, Israel\\
  {\tt amejer@tx.technion.ac.il}\\
  \and
  Koby Crammer\\
  Department of Electrical Engineering\\
  Technion-Israel Institute of Technology\\
  Haifa 32000, Israel\\
  {\tt koby@ee.technion.ac.il}}
\date{}

\begin{document}
\maketitle

\begin{abstract}
  Structured classification tasks such as sequence labeling and
  dependency parsing have seen much interest by the Natural Language
  Processing and the machine learning communities. Several online
  learning algorithms were adapted for structured tasks such as
  Perceptron, Passive-Aggressive and the recently introduced
  Confidence-Weighted learning . These online algorithms are easy to
  implement, fast to train and yield state-of-the-art
  performance. However, unlike probabilistic models like Hidden Markov
  Model and Conditional random fields, these methods generate models
  that output merely a prediction with no additional information
  regarding confidence in the correctness of the output. In this work
  we fill the gap proposing few alternatives to compute the confidence
  in the output of non-probabilistic algorithms. We show how to
  compute confidence estimates in the prediction such that the
  confidence reflects the probability that the word is labeled
  correctly. We then show how to use our methods to detect mislabeled
  words, trade recall for precision and active learning. We evaluate
  our methods on four noun-phrase chunking and named entity
  recognition sequence labeling tasks, and on dependency parsing for 14
  languages.

\end{abstract}

\section{Introduction}

Large scale natural language processing systems are often composed of
few components each designed for solving a specific task.
Example tasks are part-of-speech (POS) tagging (annotate words with
their grammatical role), noun-phrase (NP) chunking (identify
noun-phrases), information-extraction (IE) or named-entity-recognition
(NER) (identify entities such as persons, organizations, locations,
amounts and dates) and dependency parsing (grammatical or semantical
relations are identified between words of a sentence). In many cases
the input of such systems is typed text, but in some cases it is
the output of an a automatic speech recognition component, or of an
optical character recognition (OCR) layer that converts an image of
printed text or hand writing to symbolic text.

Although the major tasks of algorithms designed to solve these
problems is to make the right decision or prediction, often it is not
enough as the output of one component is fed as input to a second
one. The second component may act differently on inputs of various
quality. Therefor it is desired in many situations to have not only
an output, but to accompany it with a confidence estimation score,
either for the entire prediction, or even per element (or word).
For example, an interactive machine translation system can highlight
low confidence translated segments for the user to inspect.  Another
example is an information-extraction algorithm that can use the
confidence scores to detect low-confidence field candidates and apply
a more restrictive or more aggressive extraction policy in order to
allow the user for higher precision or higher recall.

The NLP applications mentioned above, among others,
are also known as \emph{structured prediction} tasks. The input is a
general object, often with high-regularity, such sentences which are a
{\em sequences} of words. The required output is also complex and structured, for example, the role of words in sentences
are dependent and correlated. In the past decade structured prediction
has gained increased interest by the machine learning community. After
the introduction of conditional random fields
(CRFs)~\cite{Lafferty01conditionalrandom}, and maximum margin Markov
networks~\cite{Taskar03max-marginmarkov}, which are batch algorithms,
new online method were introduced. For example, the passive-aggressive
algorithm~\cite{CrammerDeKeShSi07} originally designed for binary-classification, was
adapted to NP chunking~\cite{1273171}, dependency
parsing~\cite{McDonaldCP05}, learning preferences~\cite{wick09sample}
and text segmentation~\cite{McDonaldCP05a} and so on. These new online
algorithms are fast to train and simple to implement, yet they
generate models that output merely a prediction with no additional
confidence information, as opposed to probabilistic models like CRFs
or HMMs that naturally provide confidence estimation in the form of
probability distribution over the outputs.

In this work we fill this gap proposing few methods to estimate
confidence in the output of discriminative non-probabilistic
algorithms. Some of our algorithms are very general and can be used in
many structured prediction problems, and in combination of a
wide-range of learning algorithms and models. We focus and exemplify
our methods in two tasks: sequence labeling, and in particular
named-entity recognition and noun-phrase chunking, and dependency
parsing. In both tasks we show how to compute  \emph{per word} confidence estimates in the
predicted label, such that the confidence reflects the
probability that the label is correct.

Inspired by the recently introduced confidence-weighted
learning~\cite{DredzeCrPe08,CrammerKuDr09} we develop methods that are
based on its representation, and in particular we induce a
distribution over labelings from the distribution maintained over
weight-vectors. Additionally, we provide among the first results of
applying CW algorithms to the tasks mentioned above.

After describing our proposed methods for confidence estimation, we
provide comprehensive empirical results for evaluating various
aspects of these algorithms. First, we evaluate the ability to estimate
the confidence well in a relative setting where confidence is
meaningful only when compared to another confidence quantity. Here,
the goal of an algorithm is to rank all labeled words (in all
sentences), that can be thought of as a retrieval of the erroneous
predictions, which can then be passed to human annotator for an
examination.

Second, some of the confidence measures are in the range $[0,1]$ and
thus can be thought of as probabilities. Our second set of experiments
evaluates the accuracy of these confidence measures. The goal of the
algorithm is that the frequency of correct inputs with some confidence
value would be close to this confidence value. That is the predicted values
between $0$ and $1$ would correspond their statistical properties over
the data.

Next we describe two applications of using confidence. The first one
is using the confidence to trade recall and precision, we show that
our methods can be used to increase precision at cost of decreasing recall,
where the overall f-measure is not dropping.
The second application is active learning, where we use confidence to
chose the sentences to be annotated.

A short version of this paper was presented in the conference on
empirical methods in natural language
processing~\cite{MejerCr10}. This long version contains the following
{\em additional material} (1) Evaluation on a second task of
dependency parsing of $14$ languages. (2) Additional application of
trading recall and precision. (3) Evaluation of additional algorithm,
related to CRF~\cite{Lafferty:2001:CRF}. (4) Study of
sensitivity to parameters in the main algorithm.

\begin{figure}[!t!]
\begin{center}
\begin{tabular}{c}
\subfigure[Parsing]{\includegraphics[width=0.7\textwidth]{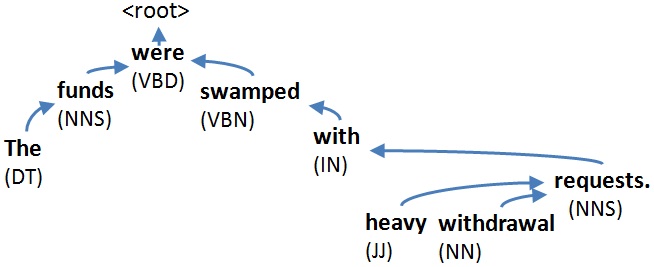}\label{fig:Seq_Parsing_Example_parsing}}\\~\\
\subfigure[Name Entity Recognition]{\includegraphics[width=0.7\textwidth]{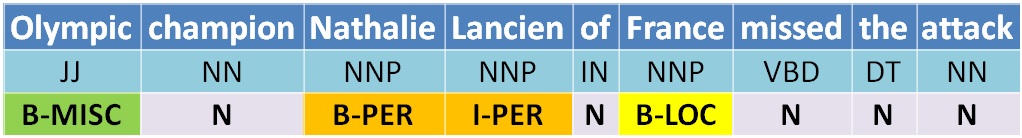}\label{fig:Seq_Parsing_Example_ner}}
\end{tabular}
\caption{Structured Prediction Examples (a) Dependency parsing:
  Connect words to a directed connected tree. (b) Name entity
  recognition: annotate words that are part of persons, locations and
  so on.}
\label{fig:Seq_Parsing_Example}
\end{center}
\end{figure}

\section{Structured Prediction}
Structured prediction problems involve complex input and complex
output, where both are composed of smaller atoms.  Consider for
example part-of-speech (POS) tagging. Given a sentence the goal is to
annotate each word with a tag reflecting its grammatical role in the
sentence. Here the input is a sentence composed of words and the
output is a label of each word. Humans often perform this task by
inspecting both the word identity and its context in the sentence.

Two additional related problems are Noun-Phrase (NP) chunking and
Named Entity Recognition (NER). In both problems the input is a
sentence as well, yet the goal is to annotate {\em segments} of words
which are either noun-phrases or specific named entities. We model
these problems in a similar way to POS, where the goal is to annotate
each word whether it belongs to a noun-phrase (or a named-entity) or
not.

There are only two possible categories in NP chunking, one indicating
that the word belongs to a noun-phrase and one indicating that it is
not the case. NER is slightly more involved as typically there are
several categories. For example a system developed about a decade
ago~\cite{Sang03introductionto} uses four categories: Person,
Location, Organization and Misc, yet additional categories are also
used such as Date, Address, Amount and so on. As mentioned above the
total number of categories is often larger by one to indicate that
{\em none} of the possible named-entities is described by that word. A
NER example is shown in \figref{fig:Seq_Parsing_Example} where the
sentence and POS are given in the two top rows and the task is to find
the proper labeling as shown in the bottom row.

Both of these problems are special case of the general problem of
sequence labeling. Here given inputs $\vx\in\X$ - e.g. sentences -
with finite number $n$ of atoms - e.g. words. The goal is to annotate
each atom with a label $y \in Y$, where we assume that the number $L$
of possible labels $Y$ is finite and known. We denote by $\vy$ the
concatenation of the labels of all atoms which belongs to the product
of the label set, that is

\(
\vy\in \Y(\vx)
\)
where $\Y(\vx)= {\overbrace {Y\times Y\dots Y}^n}$.

Dependency parsing of a sentence is one form of grammatical-reasoning
of text. Given a sentence the goal is to output a directed-tree over
the words, where an edge from one word to another indicate a
dependency relation between the words. The output of dependency
parsing is more complex than of shallow-parsing, in which the output
is a sequence of grammatical roles per sentence, and is similar to the
output defined by a context-free grammar.  Concretely, given a
sentence $w_1 \dots w_n$ of length $n$ the goal is to connect each
word to some other word or a special word called {\em root}, that is
the labels are the words of that input sentence or the {\em root}. In
other words, all words depended exactly in some other words (called
head), yet an head may have many dependents. Exactly one word must
depend in the root, and all other words depend on it directly of
indirectly. As our goal is to construct a parse {\em tree} over
sentences there is one global constraint: the graph induced must by
acyclic with no loops. That is, a word can not be dependent of
another word, which in turn is a dependent of the first word, or a
word the is a dependent of it (directly or indirectly).  Formally a
parse tree $\vy$ for a sentence with $n$ words is a relation $\vy
\subset \{1 \dots n\} \times \{0 \dots n\}$, where the special index
$0$ indicates the root of the tree. For each word $i$ the set $\{
(i,t) \in\vy\}$ is of size $1$ (a function). The set $\{ (i,0)
\in\vy\}$ is of size $1$, as only a single word is connected to the
root. Finally, the induced graphs has no loops.

An illustration of a dependency parse is shown in
\figref{fig:Seq_Parsing_Example} .

We use a unified notation for both sequential labeling and dependency
parsing (as well as other problems) and define a scoring function
\(
s(\vx,\vz),
\)
which assign a real scalar value scoring how well the complex label $\vy$ should be the
label of the input $\vx$. Given such a function a prediction is
defined to be the labeling with maximal score, that is,
\begin{equation}
\hvy = \arg\max_{\vz\in\Y(\vx)} s(\vx,\vz),
\label{inference}
\end{equation}
where $\Y(\vx)$ are all possible labelings of the input $\vx$,
e.g. all possible parse trees for a given sentence $\vx$.

In this work, we restrict ourself to linear functions $s(\cdot,\cdot)$
of some parameters~\cite{collins-02,CrammerDrKu09}, that is,
\begin{align}
s(\vx,\vz) =\vmu\cdot\fun(\vx,\vz)~,
\label{linear_scoring}
\end{align}
where $\fun(\vx,\vy)\in\reals^d$ is a joint feature mapping of an
instance $\vx$ and a labeling $\vy$ into a common vector
space. Features are derived from combinations of words (unigrams or
n-grams), part-of-speech, orthographic features such as capital
letters, hyphens and so on. The vector $\vmu\in\reals^d$ parameterize
the function $s(\cdot,\cdot)$, where $\vmu$ is chosen by a learning algorithm such that for all sentences in the training set, the label $\hvy$ output by the system is close or similar to
the correct label (or gold label) of these sentences, that is
\begin{align}
\hvy=\arg\max_{\vz\in\Y(\vx)}\vmu\cdot\fun(\vx,\vz)
\label{linear_model}
\end{align}
should be close to the best label.

A brute-force approach for computing the best label
$\hvy=\arg\max_{\vz\in\Y(\vx)}\vmu\cdot\fun(\vx,\vz)$ is not feasible
as, typically, the size of the set $\Y(\vx)$ grows very fast with
the size $n$ of the input $\vx$. For example, in sequence labeling
there are $L^n$ possible labeling of a sentence of length $n$, and
there are $n^{n-2}$  directed acyclic trees over $n$ words (or nodes).

We thus employ common approach and factor the scoring function
$s(\vx,\vy)$ by factoring the appropriate feature function
$\fun(\vx,\vy)$.
For sequence labeling we allow only functions over a single label or
pairs of consecutive labels,
\begin{align}
\fun(\vx,\vy) = \sum_{p=1}^n \fun(\vx, \yi{p}) + \sum_{q=2}^n \fun(\vx, \yi{q}, \yi{q-1})~.
\label{features}
\end{align}
Such factorization allows to perform the search for the best labeling
$\hvy$ time linear in $n$ and quadratic in $L$ using
the Viterbi dynamic-programming algorithm.

\begin{figure}[!t!]
\begin{center}
{\includegraphics[width=0.7\textwidth]{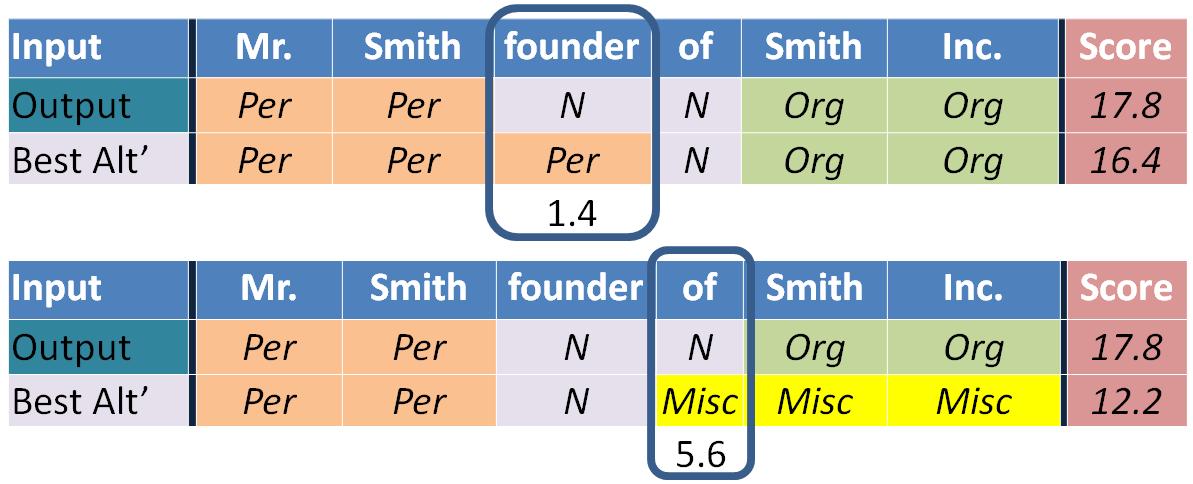}
\caption{Illustration of the margin based method to estimate confidence. Score difference between the highest and second highest scoring label per word is defined as the confidence. In this example the word "of" is labeled with higher confidence than the word "founder".}
\label{fig:Margin_Confidence}}
\end{center}
\end{figure}

For dependency parsing we build on MSTParser of \namecite{McDonaldCP05b}
and focus on non-projective parsing (tree edges may cross) with
non-typed (unlabeled) edges. MSTParser factors the score for each
parse to be a sum of the score over its edges, that is,
\[
\fun(\vx,\vy)=\sum_{(i,j)\in\vy }\fun(\vx,i,j)~,
\]
where as mentioned above every pair $(i,j)\in\vy$ represents a single
edge between word $w_i$ and word $w_j$. Example features are the distance between the two words, words
identity and words part-of-speech.  Using this factorization, the
search for the best tree can be computed efficiently by first
constructing a full directed graph over the words of the sentence with
weighted edges and then outputting the maximal spanning tree (MST) of
the graph, which can be computed for dense graphs in quadratic time in the length of the
sentence using Chu-Liu-Edmonds (CLE) algorithm~\cite{Chu-Liu/65,E16,T33}.

\section{Confidence Estimation}
\label{sec:confidence}

Many large-margin-based training algorithms maintain and output linear
models in the form of \eqref{linear_model}. Linear models are easy to
train, yet in the end-of-the-day such models are designed merely to
make a prediction which is a single labeling $\hvy$ given an input
$\vx$, with no additional information about the quality or correctness
of that prediction. This behavior assumes that the predicted labeling
$\hvy$ will be used ignoring the quality of each specific output (as
opposed to global quality of the system, such as average accuracy,
precision or recall).

\begin{table}[!t!]
\begin{center}
{
\begin{tabular}{|l|c|c|c|c|}
\hline
& Margin    & Marginal    & \emph{K}-Best   & \emph{K}-Alternatives   \tabularnewline
& Based     & Probability & Alternatives    & by Stochastic           \tabularnewline
&           &             &                 & Models                  \tabularnewline
\hline
\hline
{\em Absolute}      & No  &      Yes &      Yes &    Yes   \tabularnewline
confidence score    & (only relative)&   &   & 	\tabularnewline
\hline
Hyper-parameters    &       &       &       &       \tabularnewline
tuning              & No    & Yes	& Yes	& Yes	\tabularnewline
\hline
Complexity  &                       &                       &                              &       \tabularnewline
Sequences   & $O(n\cdot |\Y|^2)$	& $O(n\cdot |\Y|^2)$	& $O(K\cdot n\cdot |\Y|^2)$	   & $O(K\cdot n\cdot |\Y|^2)$	  \tabularnewline
\hline
Complexity  &                       &                       &                              &       \tabularnewline
Parsing     & $O(n^3)$	            & -	                    & $O(K\cdot n^2)$	           & $O(K\cdot n^2)$	  \tabularnewline
\hline
\end{tabular}
\label{table:conf_methods_comparison}
\caption{Properties of confidence estimation methods.}
}
\end{center}
\end{table}

There are situations for which additional information per labeling
$\hvy$ is useful, for example when the user of the prediction system
has the option to ignore $\hvy$, or when using erroneous $\hvy$ is
worse than not using it at all. One possible scenario is when the
output $\hvy$ is used as an input of another system that integrates
various input sources or is sensitive to the correctness of the specific
prediction. In such cases, additional confidence information about
the correctness of these feeds for specific input can be used to
improve the total output quality. For example, data mining systems use
NER predictors as sub-components. Such systems may use few NER
predictors which all provide {\em confidence} information about their
output, and allow the data-mining system to integrate better the output
from all NER predictors, and overcome possible mistakes.

Another case where confidence information is useful, is when there is
an additional agent that validates the output $\hvy$ (as was done in
building RCV1~\cite{lewis-04}). Confidence information associated with output
$\hvy$ can be used to direct the agent to small subset of ``suspect''
outputs rather than use a random sample or evaluate all outputs.

We now describe three methods to compute confidence in prediction of
the form of a single number per prediction.
One method only provide {\em relative} confidence information. This
numeric information can be only used to compare two outputs and decide
which is of better quality. One use of such relative confidence
information is to {\em rank} all predictions or outputs according to
their confidence numeric-score, and validate the output of the outputs
which are assigned with the most low-confident score values. One
property of this {\em relative} score is that any monotonic
transformation of the confidence values yields equivalent confidence
information (and ranking).

Other two methods described below provide {\em absolute} numeric
confidence information in the prediction. Conceptually, the numeric
confidence information is given as the probability of a prediction to
be correct. We interpret these probabilistic outputs in a frequentists
approach.  A large set of events (predictions) all assigned with
similar probability confidence value $\nu$ of being correct, indeed
should contain about $\nu$ fraction of the predictions of that group
correct. Clearly, any absolute information is relative as well, as two
absolute confidence values may also be compared to each other to
determine which output is of better quality.

In \secref{sec:margin} we describe our relative confidence
method which is based on extending the notion of {\em margin}
originally used to design support-vector machines in the context of
binary-classification~\cite{BoserGuVa92,BartlettScScSm00}. Intuitively, we define the confidence
in a prediction to be the difference between the score of the (best)
labeling $s(\vx,\hvy)$, and an additional prediction.

Next, in \secref{sec:marginal} we define a probability distribution by
using the score values $s(\vx,\vz)$ as arguments of a
suitable-function yielding non-negative values which sum to unit, and
then compute the marginals induced from this distribution. Such method
is used for example in conditional-random fields (CRF)~\cite{Lafferty:2001:CRF}.

Finally, our third approach, which output absolute confidence
information as well, is described in \secref{sec:k_draw}.  This method
generates few alternative outputs additional to $\hvy$, and evaluates
the confidence by computing the agreement between the output $\hvy$
and the alternatives. We use two methods to generate the alternatives:
one deterministic based on extension of the prediction algorithms to
produce $K$-best predictions, and one stochastic based on sampling
models.  A comparison of the confidence estimation methods properties
is summarized in Table~1
, each row is described in details in the appropriate place below.

\subsection{A Margin-Based Method}
\label{sec:margin}
\begin{figure}[!t!]
\begin{center}
{\includegraphics[width=0.9\textwidth]{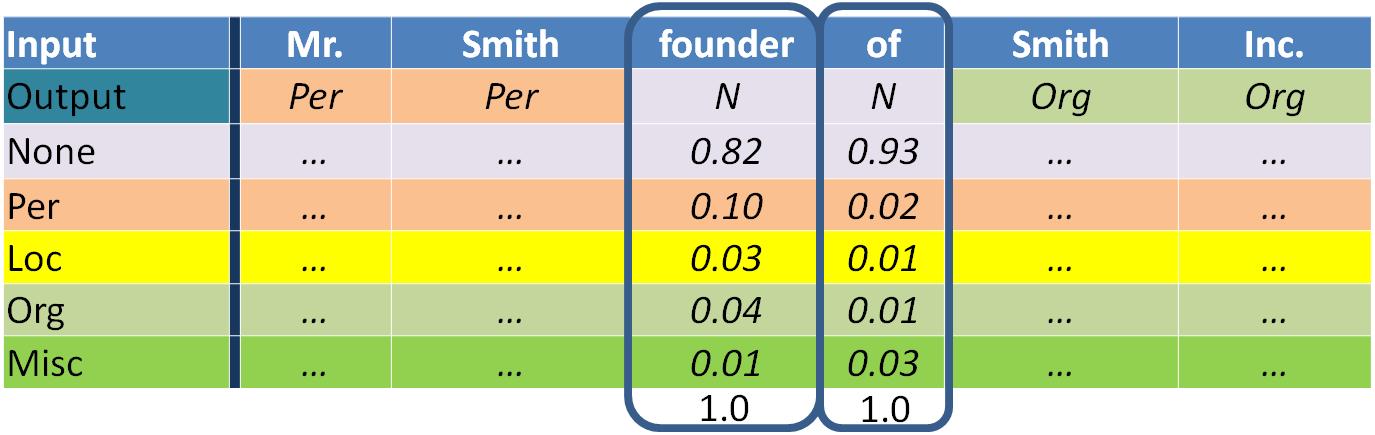}
\caption{Illustration of the marginal-probability based method to estimate confidence. The confidence in the chosen label is defined to be the marginal-probability of the label. In this example the word "of" is labeled as \emph{None} with higher confidence than the word "founder".}
\label{fig:Marginals_Confidence}}
\end{center}
\end{figure}
Our first method extends the notion of margin known mainly in the
context of support vector
machines~\cite{BartlettScScSm00}. Originally, margin is a geometrical
concept and is defined to be the distance of an input point embedded
in some vector space to the separating
hyperplane. Later~(e.g. \cite{BredensteinerBe99,CrammerSi01a,HarPeledRoZi02,WestonWa99,Taskar03max-marginmarkov})
it was extended in the context of multiclass problems. For example, in
sequential labeling it is defined to be the difference between the
best scoring labeling $\hvy$ and the second best,
\[
\vmu\cdot\fun(\vx,\hvy) - \max_{\vz\neq\hvy} \vmu\cdot\fun(\vx,\vz) ~.
\]
This definition is too crude for our purpose as we need a measure of
confidence per unit, or word in our tasks. Thus, we refine the above
definition and define the margin of the p$th$ word
to be the difference in the score of the best labeling score and the
score of the best labeling
where we set the label of that word to anything but the label
with the highest score. Formally, as before we define the best
labeling $\hvy=\arg\max_\vz \vmu\cdot\fun(\vx,\vz)$, then the margin
of the p$th$ word is defined to be,

\begin{equation}
\delta_p = \vmu\cdot\fun(\vx,\hvy) - \max_{\vz_{\vert p}
  \neq\hvy_{\vert p}} \vmu\cdot\fun(\vx,\vz)~,
\label{delta}
\end{equation}
where $\vz_{\vert p}$ is the labeling of the p$th$ word according to
the labeling $\vz$ of the entire input. Since labeling of consecutive
words are dependent according to the model the labeling of additional
words may change from $\hvy$ by the restriction that the labeling of
the p$th$ word is not its labeling according to the best labeling
$\hat{y}_p$.  In parsing, for example, changing the parent of a single
word in a parse tree may cause a loop in the graph that require
changing the parents of additional words to resolve the loop.

We refer to this method as \texttt{Delta} where the confidence
information is the margin which is a difference or delta between two
score values. Clearly, the absolute margin value $\delta_p$ provides
confidence score that is only relative and not absolute, namely it can
be used to compare the confidence in two labeling, yet there is no
semantics defined over the scores as it is not calibrated to be in
$[0,1]$.

An illustration of the margin method is given in
\figref{fig:Margin_Confidence} for the sentence fragment {\em Mr Smith
  founder of Smith Inc.}. The top row of both panels shows the highest
scoring labeling $\hvy$ which attains a score of $17.8$ in our
case. The second row of each panel shows the best labeling where the
label of some word is restricted. The top panel shows the best
labeling where the label of the word {\em founder} is restricted not
to be {\em N} - its labeling according to $\hvy$. Clearly its score of
$16.4$ is not higher than the score of the best labeling, as the
$\max$ operator is performed over a strict subset of possible
labelings $\vz$. The bottom panel shows similar process for the word
{\em of} where its labeling is restricted from being {\em N}, and the
one that has the highest score is {\em Misc}, the difference in score
is $5.6$ which is defined to be the confidence value. Thus, in our
example, the confidence in the labeling of the word {\em of} is higher
than the confidence in the labeling of the word {\em founder}.

A straightforward implementation of the {\tt Delta} method requires
repeating the inference process $n$ times, once per word of the input,
each time with a single constraint over the labeling, that is, the
label $\hat{y}_p$ is not allowed for the p$th$ word - the word for
which the confidence is being evaluated. Such implementation costs
$O(n\cdot\textrm{(inference~cost)})$. We used this approach for
dependency parsing task, and thus the computation complexity is $O(n\cdot
n^2)$ which is cubic in $n$.

For the sequence labeling task the computation
of {\tt Delta} confidence can be improved by using the
forward-backward-Viterbi algorithm (similar to standard
forward-backward algorithm~\cite{Rabiner89atutorial}). This dynamic
programming algorithm allows to efficiently compute the score of the
best sequence labeling that includes a specific label constraint, therefore
the computation of the {\tt Delta} confidence scores can performed in
$O(n\cdot |\Y|^2)$, the same complexity as the standard Viterbi
algorithm used for sequence labeling prediction.

\subsection{Marginal-Probability Method} \label{sec:marginal}
The second method we describe is based on specific function converting
score values into probabilities.
We follow the same modeling of conditional random fields (CRF)~\cite{Lafferty:2001:CRF}
and define the conditional probability,
\begin{equation}
P(\vy\vert\vx) = \frac{\exp\braces{c\paren{\vmu\cdot\fun(\vx,\vy)}}}{Z_x}
\end{equation}
where $Z_x$ is a normalization factor over all possible labeling to the input $\vx$,
\begin{equation}
Z_\vx = \sum_{\vz\in\Y(\vx)}{\exp\braces{c\paren{\vmu\cdot\fun(\vx,\vz)}}},
\end{equation}
and $c>0$ is a scaling parameter. While in some learning algorithms,
such as CRFs, the model parameters are trained to maximize the
log-likelihood of data using the above conditional probability, this
is not the case for other algorithms such as ones that are based on
large-margin such as the passive-aggressive~\cite{CrammerDeKeShSi07} algorithm or the
confidence-weighted algorithm~\cite{DredzeCrPe08,CrammerDrPe08}, both described below. For this
reason the coefficient $c$ is used to allow tuning of the model score
to confidence score.

We define the confidence in a prediction of the p$th$ word to be the
marginal probability of that prediction, that is,
\[
P(\hvy_{\vert p} \vert \vx) = \sum_{\vz ~:~ \vz_{\vert p} = \hvy_{\vert p}} P(\vz\vert\vx) ~.
\]
Using the dynamic-programming forward-backward
algorithm~\cite{Rabiner89atutorial} for sequence labeling the marginal
probability can be efficiently computed in $O(n\cdot |\Y|^2)$ for all
assigned labels $P(\hvy_{\vert p} \vert \vx) $ and $p=1 \dots n$. It
is common to refer to the quantities computed by the forward-backward
algorithm by $\gamma$, so we also refer to this method as
\texttt{Gamma}. We report the results of this method only in the
context of sequence labeling and not dependency parsing, as we found
empirically that it is was shown not to perform well compared to {\tt
  Delta} described above and some of the methods we describe next.

An illustration of this process appear in
\figref{fig:Marginals_Confidence} where the best labeling appears in the
top row, and the marginals for two words appear in the following rows.
In this example the confidence in the prediction of the word {\em
  founder} is defined to be $0.82$ and the confidence in the
prediction for the word {\em of} is defined to be $0.93$, the marginal
value in both cases. One notable property is that the confidence
values are close to $1$, this is because of the exponent-function used
to convert scores to probabilities, a phenomena that was shown to
appear in other contexts~\cite{malkin2009-mlrsc-icassp}.

\subsection{Confidence by Alternatives}
\label{sec:k_draw}

\begin{figure}[!t!]
\begin{center}
{\includegraphics[width=0.9\textwidth]{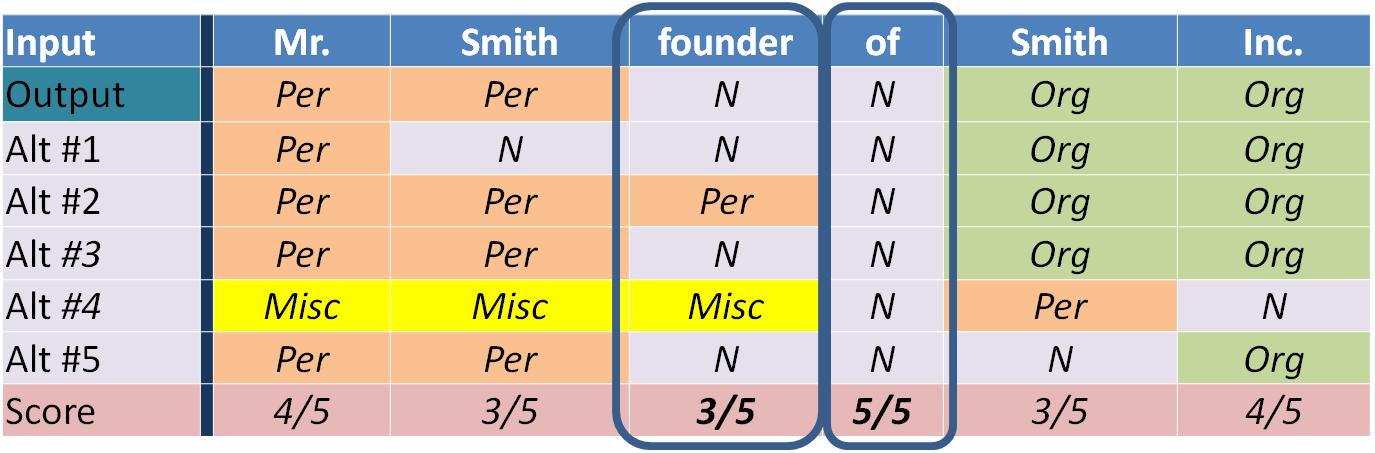}}
\caption{Illustration of the alternatives method to estimate confidence. Confidence is defined as the fraction of alternative labeling that agree with the prediction.}
\label{fig:K_Alternatives_Confidence}
\end{center}
\end{figure}

This method works in two stages.  First, a set of $K$ alternative
labeling for a given input sentence are generated, where the
predicted labeling $\hvy$ is not necessarily one of the $k$ labeling.  Then,
the confidence in the predicted labeling is computed by evaluating the
agreement (or disagreement) between $\hvy$ and the $K$ alternatives.
In other words, the confidence in the prediction for a specific word
is defined to be the proportion of labeling which are consistent with
the predicted label. Formally, let $\vzii$ for $i=1 \dots K$ be the
$K$ labeling for some input $\vx$, and let $\hvy$ be the actual
prediction for the input. (We do not assume that $\hvy=\vzii$ for some
$i$). The confidence in the label $\hat{y}_p$ of word $p = 1 \dots
\vert\vx\vert$ is defined to be
\begin{equation}
\nu_p = \frac{\left\vert\left\{i~:~ \hat{y}_p = z^{(i)}_p
    \right\}\right\vert}{K} ~.
\label{k_draw_eq}
\end{equation}

The process is illustrated in \figref{fig:K_Alternatives_Confidence}.
The input is given in the top row and the prediction is given in the next row. The example
output $\hvy$ includes a person name {\em Mr Smith} and a company name
{\em Smith Inc}. In  the next five rows there are five alternative labelings to the input sentence.
Numeric confidence scores are given in the bottom line, and
are the fraction of alternatives which agree with the labeling of the
output. For example, in the alternative just after the output only the
word {\em Mr} is part of the person name while the following word {\em
  Smith} is labeled as a none entity. The confidence for the word {\em
  founder} is $0.6=3/5$ as in three alternatives out of five the
labeling of it ({\em None}) agrees with the labeling of the output. The confidence for the word {\em
  of} is $1.0=5/5$ as all alternatives agree with the labeling of the output ({\em None}).

We tried two major approaches to generate $K$ possible alternatives:
deterministic and stochastic.

\subsubsection{\emph{K}-Best Predictions}
\label{sec:kbest_pred}
The inference procedures are returning the
labeling $\hvy$ that achieves the highest score in
\eqref{inference},
$\hvy=\arg\max_{\vz\in\Y(\vx)} s(\vx,\vz)$.  We modify the
inference algorithm to return not a single labeling but the best $K$
distinct labelings with highest score. Formally, we pick the top-$K$
{\em distinct} labelings that satisfy,
\begin{align*}
s\paren{\vx,\vzi{1}} \geq s\paren{\vx,\vzi{2}} \comdots \geq s\paren{\vx,\vzi{K}} \geq s\paren{\vx,\vz},\\
\textrm{~  for all  ~~}
\vz \notin Z = \braces{\vzi{1} \comdots \vzi{K}}.
\end{align*}
By definition we have that the first labeling is the predicted output
$\vzi{1} = \hvy$ and that all $K$ labelings differ from each other
$\vzii \neq \vzi{j}$ for $i\neq j$.
Specifically, we use the k-best Viterbi algorithm~\cite{ChoSch-SNLW-89} to find the k-best
sequence labeling in $O(K\cdot n\cdot |\Y|^2)$, and for dependency
parsing we used the k-best Maximum Spanning Trees
algorithm~\cite{Hall07,CamFraMaf-Nw-80} to produce the $K$ parse trees with the
highest score in $O(K\cdot n^2)$.

We use two variants of this approach. The first variant assigns
uniform importance to each of the $K$ labelings ignoring the actual
score values. We call this method \texttt{KB}, for $K$-best. The
second variant assigns a specific importance weight $\omega_i$ to each
labeling $\vzii$, and evaluate confidence using the weights,
where we set the weights to be their score value clipped at zero from
below $\omega_i = \max\{0,\vmu\cdot \fun(\vx,\vzii)\}$. (In practice,
top scores were always positive.) We call this method \texttt{WKB} for
weighted $K$-best. \eqref{k_draw_eq} is naturally extended to a
weighted set,
\begin{equation}
\nu_p = \frac{\displaystyle ~~~~~ \sum_{ i~:~ \hat{y}_p = z^{(i)}_p}
  \omega_i ~~~~~}{\displaystyle \sum_i
  \omega_i} ~.
\label{k_draw_weighted_eq}
\end{equation}

\subsubsection{Stochastic Models}
\label{sec:stochastic_models}
Previous method used a single model to generate few alternatives. This
approach is complimentary, we use a single model to generate
(stochastically) few additional models, each is used to generate a
single alternative (best according to itself) labeling. Concretely,
given a model $\vmu$ learned by some algorithm we induce a probability
distribution over weight-vectors given $\vmu$, denoted by
$\pr{\vw \vert \vmu}$. We then draw a set of $K$ weight-vectors $\vwii
\sim \pr{\vw \vert \vmu}$, and use each to output a best labeling
according to it to get a set of alternatives,
\begin{align}
Z =  \{ \vzii ~:~  \vzii=\arg\max_\vz \vwii\cdot\fun(\vx,\vz)
 \textrm{ where } \vwii\sim \pr{\vw \vert \vmu} \} ~.
\label{z}
\end{align}

Again we use two variants of this approach. The first variant is more
generic we define the probability distribution over weights to be
Gaussian with an isotropic covariance matrix,
$\msigma=s\mi$ for some positive scale information $s$, that is we
have that,
\(
\vw\sim\norm\paren{\vmu,\msigma}~,
\)
where $\vmu$ are the parameters returned by the learning algorithm.
The value of $s$ was tuned on the training set.  We denote this method
\texttt{KD-Fix} for $K$-draws with fixed standard deviation. This
method is especially appealing, since it can be used in combination
with training algorithms that do {\em not} maintain confidence
information, such as the Perceptron or PA.

The second variant is used with classifiers that not only maintain a single
weight vector $\vmu$ but also a distribution weight vectors. For example,
the confidence weighted classifier (CW) described below in \secref{sec:learning}
maintains by definition a Gaussian distribution over weights,
$\vw\sim\norm\paren{\vmu,\msigma}$. We proceed as the previous
variant, given an input the algorithm draws $K$ alternative weight vectors
according the distribution maintained by CW scaled by scalar $s$, that is $s\msigma$, and output the
best labeling with respect to each weight vector. Again the value of $s$ was tuned on the training set.
We denote this method
\texttt{KD-PC} for $K$-draws by parameter confidence.

We stress that the two variants generating the set of $K$ alternatives
are inherently different from each other. The first deterministic
approach generates few labelings from the same model, all of the
labelings are different from each other, and one of them is always the
prediction $\hvy$. Additionally, although different from each other
the labelings are close (or similar) to each other (in term of Hamming
difference), as all attain very close (to the best) score values,
because the score over sequences decompose to scores over words and
pairs of adjacent words. The second stochastic approach generates one
alternative per sampled model. With high probability these models
differ from the model $\vmu$ used to make the prediction, and thus the
output labeling $\hvy$ may be different from each of the
alternatives. However, if the covariance matrix is close to zero, the
sampled weight-vectors are likely to be close (in terms of Euclidean
distance) to $\vmu$ and thus produce similar labeling to $\hvy$, that
is, there may be overlap between alternative labeling.

Using the stochastic models approach the $K$ alternative predictions
are generated by executing the inference algorithm $K$ times, each
time with different sampled model, therefore the computing complexity
is $O(K\cdot(\texttt{inference cost}))$, that is $O(K\cdot n^2)$ for parsing
and $O(K\cdot n\cdot |\Y|^2)$ for sequence labeling.

\section{Learning}
\label{sec:learning}

The algorithms described above to compute confidence are designed for
linear models decomposed over units, and work with any algorithm in
such setting. We evaluate the confidence estimation methods described
in \secref{sec:confidence} using online learning
algorithms~\cite{CesaBiGa06}. These algorithms are fast, efficient,
simple to implement and work well in practice. The passive-aggressive
(PA) algorithms~\cite{CrammerDeKeShSi07} were shown to achieve
state-of-the-art performance in many tasks such as NP
chunking~\cite{1273171}, dependency parsing~\cite{McDonaldCP05},
learning preferences~\cite{wick09sample} and text
segmentation~\cite{McDonaldCP05a}. Recently, confidence weighted (CW)
algorithms were introduced for binary
classification~\cite{DredzeCrPe08,CrammerDrPe08} and multi-class
problems~\cite{CrammerDrKu09} and were shown to outperform many
competitors.  We next describe both PA and CW algorithms for
structured prediction.
Both versions are reductions from structured problems to binary
classification in a manner similar to the reduction performed for
parsing using the Perceptron algorithm~\cite{collins-02}.

Online algorithms work in rounds. On the i$th$ round the online
algorithm receives an input $\vxi{i}\in\X$ and applies its current
rule to make a prediction $\hvyi{i}\in\Y(\vxii)$, it then receives the
correct label $\vyi{i}\in\Y(\vxii)$ and suffers a loss
$\ell(\vyi{i},\hvyi{i})$.  At this point, the algorithm updates its
prediction rule with the pair $(\vxi{i},\vyi{i})$ and proceeds to the
next round.  A summary of online algorithms can be found in the book
written by~\namecite{CesaBiGa06}. As noted above, in structured
prediction we assume a joint feature representation, $s(\vx,\vz)
=\vmu\cdot\fun(\vx,\vz)$ for $\fun(\vx,\vy)\in\reals^d$ (see
\eqref{linear_scoring} and the text after it).

\begin{algorithm}[tb]
{\bf Input: }
\vspace{-0.4cm}
\begin{itemize}
\nolineskips
\item Joint feature mapping $\fun(\vx,\vy)\in\reals^d$
\item Tradeoff parameter $C$
\end{itemize}
\vspace{-0.4cm}

{\bf Initialize:}
\vspace{-0.4cm}
\begin{itemize}
\nolineskips
\item $\vmui{0}=\vzero$
\end{itemize}
\vspace{-0.4cm}

{\bf For $i=1,2 \comdots T$}
\vspace{-0.4cm}
\begin{itemize}
\nolineskips
\item   Get input $\vxii\in\X$
\item   Predict best labeling $\displaystyle \hvyii=\arg\max_\vz\vmui{i-1}\cdot\fun(\vxii,\vz)$
\item   Get correct labeling $\vyii\in\Y(\vxii)$
\item   Define $\gunibp = \fun(\vx,\vyii) - \fun(\vx,\hvyii)$
\item   Compute (from \eqref{paii})
\[
\alphaii = \min\braces{C, \frac{\max\braces{0, \ell(\yi{i},\hyi{i}) - \vmui{i}\cdot\gunibp }}{\Vert \gunibp\Vert^{2}}}~.
\]\item   Set
\begin{align*}
\vmui{i+1}  &= \vmui{i} + \alphaii \gunibp
\end{align*}
\end{itemize}
\vspace{-0.2cm}

{\bf Output:} Weight vector $\vmui{T+1}$
 \caption{Sequence Labeling PA}
   \label{alg:paar_algorithm}
\end{algorithm}

\paragraph{Passive-Aggressive Learning}
We first review a version of the passive-aggressive (PA) algorithms
for structured prediction~\cite[Sec. 10]{CrammerDeKeShSi07}.  The
algorithm maintains a weight vector $\vmuii\in\reals^d$ and updates it
on each round using the current input $\vxii$ and label $\vyii$,
by optimizing:
\begin{align}
\vmui{i+1} =\arg\min_{\vmu} &~\half \Vert \vmu-\vmuii\Vert^{2} + C
\xi\nonumber\\
\textrm{s.t.} &~ \vmu\cdot\fun(\vxii,\vyii) - \vmu\cdot\fun(\vxii,\hvyii) \geq
\ell(\vyii, \hvyii) - \xi\quad,\quad \xi \geq 0 ~,
\label{PA}
\end{align}
where the loss $\ell(\yi{i},\hyi{i})$ is taken as the Hamming distance
between the two labeling which is number of incorrect edges in the
parse tree or incorrect labels in the sequence labeling tasks, and
$C>0$ controls the tradeoff between optimizing the current loss and
being close to the old weight vector.  To solve \eqref{PA} we define
the difference between the feature vector associated with the true
labeling $\vyii$ and the feature vector associated with some labeling
$\vz$ to be,
\(
\guniba = \fun(\vxii,\vyii) - \fun(\vxii,\vz) ~,
\)
~ and in particular, when we use the current model's prediction $\hvyii$ we get,
\begin{align*}
\gunibp = \fun(\vxii,\vyii) - \fun(\vxii,\hvyii) ~.
\end{align*}
The update of \eqref{PA} can be computed analytically to get a
structured version of PA-I :
\begin{align}
\vmui{i+1}  &= \vmui{i} + \alphaii \gunibp~,\nonumber\\
\alphaii &= \min\braces{C, \frac{\max\braces{0, \ell(\yi{i},\hyi{i}) - \vmui{i}\cdot\gunibp }}{\Vert \gunibp\Vert^{2}}}~.
\label{paii}
\end{align}
The algorithm is summarized in \algref{alg:paar_algorithm}.  The theoretical properties
of this algorithm were analyzed by \namecite{CrammerDeKeShSi07}, and it was
demonstrated on a variety of tasks (e.g. \cite{ChechikShShBe09}).

\begin{algorithm}[tb]
{\bf Input: }
\vspace{-0.4cm}
\begin{itemize}
\nolineskips
\item Joint feature mapping $\fun(\vx,\vy)\in\reals^d$
\item Initial variance $a>0$
\item Confidence parameter $\phi$
\end{itemize}
\vspace{-0.4cm}

{\bf Initialize:}
\vspace{-0.4cm}
\begin{itemize}
\nolineskips
\item $\vmui{0}=\vzero$ , $\msigmai{0}=a \mi$
\end{itemize}
\vspace{-0.4cm}

{\bf For $i=1,2 \comdots T$}
\vspace{-0.4cm}
\begin{itemize}
\nolineskips
\item   Get input $\vxii\in\X$
\item   Predict best labeling $\displaystyle \hvyii=\arg\max_\vz\vmui{i-1}\cdot\fun(\vxii,\vz)$
\item   Get correct labeling $\vyii\in\Y(\vxii)$
\item   Define $\gunibp = \fun(\vx,\vyii) - \fun(\vx,\hvyii)$
\item   Compute $\alphaii$ and $\betaii$ using \eqref{mar_var} and \eqref{alpha_beta}
\item   Set
\begin{align*}
\vmui{i} &= \vmui{i-1} + \alphaii \msigmai{i-1}\gunibp\\
\msigmaii^{-1} &= \msigmai{i-1}^{-1} + \betaii \,\textrm{diag}\paren{  \gunibp\gunibp^\top}
\end{align*}
\end{itemize}
\vspace{-0.2cm}

{\bf Output:} Weight vector $\vmui{T+1}$ and confidence
$\msigmai{T+1}$
 \caption{Sequence Labeling CW}
   \label{alg:cwar_algorithm}
\end{algorithm}

\paragraph{Confidence-Weighted Learning}

Online confidence-weighted (CW) learning
\cite{DredzeCrPe08,CrammerDrPe08} generalizes the passive-aggressive
(PA) update principle to multivariate Gaussian distributions over the
weight vectors - $\vw\sim\norm\paren{\vmu, \msigma}$. Originally, it was
designed for binary classification and later was extended to
multi-class problems~\cite{CrammerDrKu09}, speech
recognition~\cite{Crammer:2010fk} and sequence prediction~\cite{MejerCr10}.
We now sketch a generalization for
structured problems which contains all previous versions as special
cases.

The mean $\vmu\in\reals^d$ contains the current estimate for the best
weight vector, whereas the diagonal Gaussian covariance matrix
$\msigma\in\reals^{d \times d}$ captures the confidence in this
estimate. More precisely, the diagonal elements $\msigma_{p,p,}$
capture the confidence in the value of the corresponding weight
$\vmu_{p}$ ; the smaller the value of $\msigma_{p,p,}$ is, the more
confident is the model in the value of $\vmu_{p}$. Full matrices are
not feasible as the dimension is in the order of millions.

CW classifiers are trained according to a PA rule that is modified to
track differences in Gaussian distributions.  At each round, the new
mean and covariance of the weight vector distribution is chosen to be
the solution of the following optimization problem,
\begin{align}
(\vmui{i+1},\msigmai{i+1}) = \arg\min_{\vmu,\msigma} \KL\paren{ \norm\paren{\vmu, \msigma} \,\Vert\, \norm\paren{\vmui{i}, \msigmai{i}}}\nonumber\\
s.t.\ Pr[ \gunibp\cdot\vw \geq0]\geq  \Psi\paren{\phi \ell(\yi{i},\hyi{i}) }
\label{CW_optimization}
\end{align}
where $\Psi$ is the cumulative function of the normal distribution and
$\phi>0$ controls the tradeoff between adjusting the model according
to last example and being close to the old weight vector
distribution. The larger the loss is, the larger probability we
require for the event
$\gunibp\cdot\vw\geq0$.

The solution for the CW updates is of the form,
\begin{align}
\vmui{i} &= \vmui{i-1} + \alphaii \msigmai{i-1}\gunibp\nonumber\\
\msigmai{i}^{-1} &= \msigmai{i-1}^{-1} + \betaii\textrm{diag}\paren{\gunibp\gunibp^\top} \label{CW_update}
\end{align}
where $\textrm{diag}(A)$ return a diagonal matrix which equals to the
diagonal elements of the matrix $A$. The two scalars $\alphaii$ and
$\betaii$ are computed using the mean and variance of the margin,
\begin{equation}
\varbii = \gunibp^{\top}\msigmaii \gunibp ~,~ \marbii = \vmuii\cdot
\gunibp ~,
\label{mar_var}
\end{equation}
and are,
\begin{align}
\phi_\ell &=\phi \ell(\yi{i},\hyi{i}) ~,~ \phia=1+\phi_\ell^2/2 ~,~ \phib=1+\phi_\ell^2 \nonumber\\
\alphaii &= \max\left\{0,\frac{1}{\varbii\phib}\paren{ - \marbii \phia +
  \sqrt{\marbii^2 \frac{\phi_\ell^4}{4} + \varbii \phi_\ell^2 \phib}}\right\}\nonumber\\
\betaii &= \frac{\alphaii \phi_\ell}{\sqrt{\varaii}}\quad,\quad {\varaii} =
\frac{1}{4}\paren{-\alphaii \varbii \phi_\ell    +\sqrt{\alphaii^2
    \varbii^2 \phi_\ell^2 + 4\varbii}}^2 \label{alpha_beta} ~.
\end{align}

The method presented here is called \emph{1-best} binary reduction since the binary example for the update step at each round was generated as the difference between a single prediction, the model's best prediction, and the true labeling. There are variants of this method that at each round utilize multiple predictions, usually the \emph{n-best} predictions, to generate multiple binary examples for updating the model. Please see \cite{Crammer05scalablelarge-margin,McDonaldCP05} for more details.

Finally, we used parameter averaging with both algorithms. That is,
during test time we are not using the final parameter vector
$\vmui{T+1}$, but instead using its average $(\sum_t \vmui{t} ) /
(T+1)$. It was shown to improve performance in other settings, and for
us it either improved performance a bit, or did not make it worse.

\section{Data}

\begin{table}[!t!]
\begin{center}
{
\begin{tabular}{|l|r|r|r|}
\hline
Dataset & Sentences  & Words  & Features \tabularnewline
\hline
\hline
NP chunking & 11.0\,K  & 259.0\,K  & 1.35\,M\tabularnewline
\hline
NER English & 17.5\,K  & 250.0\,K  & 1.76\,M\tabularnewline
\hline
NER Spanish & 10.2\,K  & 317.6\,K  & 1.85\,M\tabularnewline
\hline
NER Dutch & 21.0\,K  & 271.5\,K  & 1.76\,M\tabularnewline
\hline
\end{tabular}
\label{table:sequences_datasets}
\caption{Properties of sequences labeling datasets.}
}
\end{center}
\end{table}

\begin{table}[!t!]
\begin{center}
{
\begin{tabular}{|l|r|r|r|}
\hline
Dataset & Sentences  & Words  & Features \tabularnewline
\hline
\hline
Arabic & 1.5\,K	& 54.3\,K& 1.03\,M\tabularnewline
\hline
Bulgarian & 12.8\,K & 	190.2\,K& 2.64\,M\tabularnewline
\hline
Chinese & 56.0\,K & 	337.1\,K& 4.92\,M\tabularnewline
\hline
Czech & 72.7\,K	& 1,249.0\,K& 12.69\,M\tabularnewline
\hline
Danish & 5.2\,K	& 94.3\,K& 1.22\,M\tabularnewline
\hline
Dutch & 13.3\,K	& 195.0\,K& 2.36\,M\tabularnewline
\hline
English & 39.8\,K	& 950.0\,K& 7.00\,M\tabularnewline
\hline
German & 39.2\,K	& 699.6\,K& 6.99\,M\tabularnewline
\hline
Japanese & 17.4\,K	& 151.4\,K& 0.85\,M\tabularnewline
\hline
Portuguese & 9.0\,K	& 206.6\,K& 2.50\,M\tabularnewline
\hline
Slovene & 1.5\,K	& 28.7\,K& 0.55\,M\tabularnewline
\hline
Spanish & 3.3\,K	& 89.3\,K& 1.40\,M\tabularnewline
\hline
Swedish & 11.0\,K	& 191.4\,K& 2.50\,M\tabularnewline
\hline
Turkish & 5.0\,K	& 57.5\,K& 1.10\,M\tabularnewline
\hline
\end{tabular}
\label{table:parsing_datasets}
\caption{Properties of dependency parsing datasets.}
}
\end{center}
\end{table}

We evaluated our algorithms on two types of structured predictions
problems: sequence labeling and dependency parsing. For the sequence
labeling experiments we used four large sequential classification
datasets taken from the CoNLL-2000, 2002 and 2003 shared tasks:
noun-phrase (NP) chunking~\cite{Sang00introductionto}, and
named-entity recognition (NER) in Spanish,
Dutch~\cite{Sang02introductionto} and
English~\cite{Sang03introductionto}.  The properties of the four
datasets are summarized in Table~2 
For the task of NP chunking we used the BIO system of labeling marking the
first word (beginning) of a phrase (B), additional words of a phrase
(in a phrase; I) and other words (O).  For NER problems we used the
same system for the four categories ending up with nine
possible labels. Eight labels are the beginning of a name-entity (B)
or being in it (I) for every the four categories: Location, Organization,
Person and Miscellaneous. The ninth label is marking O(ther) words.

We followed previous feature generation
process~\cite{Sha03shallowparsing}. For NP chunking we used word and
part-of-speech over a window of size (5) centered at the word to be
labeled. For NER we used word and standard sub-word features including
word, part-of-speech, suffix and prefix identity as well as standard
orthographic features (e.g.  word is capitalized), with all features
over a window of size five (5) centered around the word at
investigation.

To evaluate the task of dependency parsing we used $14$ datasets: $13$ languages used in CoNLL 2006 shared task (Arabic,
 Bulgarian, Chinese, Czech, Danish, Dutch, German, Japanese,
 Portuguese, Slovene, Spanish, Swedish and Turkish\,\footnote{See
 \url{http://nextens.uvt.nl/~conll/} for details}, and the English Penn
 Treebank. The properties of the datasets are summarized in Table~3
The feature representation of edges between words is generated as a combination of the connected words, the part-of-speech of the words and their local context, that is words before, after and between the connected words, the direction of the dependency (left or right) and the distance between the words (for more details see ~\cite{McDonaldCP05}).
 For dependency parsing evaluation we
 used a single split of training, development and testing sets for
 each language.  The number of sentences in the training datasets is
 ranging between $1.5-72K$, with an average of $20K$ sentences, $30K-1.2M$
 words and $0.5-12.7M$ features. The test sets contain $\sim\!\!400$
 sentences and $\sim\!\!6K$ words for all datasets, except English
 with $2.3K$ sentences and $55K$ words.

\begin{table}[!t!]
\begin{center}
{
\begin{tabular}{|l|r|r|r|}
\hline
 & CW  & 5-best PA & Perceptron\tabularnewline
\hline
\hline
NP chunking & 0.947 & 0.946 & **0.944 \tabularnewline
\hline
NER English & 0.877 & * 0.870  & * 0.862\tabularnewline
\hline
NER Dutch & 0.787 &  0.784 & * 0.761\tabularnewline
\hline
NER Spanish & 0.774 & 0.773  & * 0.756 \tabularnewline
\hline
\end{tabular}
\label{table:sequences_results}
\caption{Averaged F-measure of methods. Statistical significance (t-test) are with respect to CW, where {*} indicates 0.001 and  ** indicates 0.01}
}
\end{center}
\end{table}

\section{Prediction Performance Evaluation}
\label{sec:perf_eval}
Our primary goal is developing new methods to estimate confidence in
prediction, not prediction perse. Yet prediction itself is the end goal
of learning. Furthermore, we could not find published performance
evaluation of CW on structured prediction tasks, and the results below
are among such first results.  We thus briefly report for completeness
the performance of CW comparing it with previous state-of-the-art
online algorithms. Below we report in details the results of our
evaluation of various confidence estimation algorithms.  Our goal was
to evaluate whether CW improves performance for these structured
prediction tasks as it does for binary classification~\cite{DredzeCrPe08,CrammerDrPe08} and
multiclass prediction~\cite{CrammerDrKu09}.  As our goal is confidence and not
achieving the best accuracy we note that the performance results we
report now are not necessarily the best published in the literature as
they are obtained by using existing tools, and specifically MSTParser "out of
the box" not incorporating recent parsing advancements.

We compared the performance of
CW \algref{alg:cwar_algorithm} with the passive-aggressive algorithm,
which was shown to be a state-of-the art in both tasks. Specifically,
we used $5$-best PA (the value of five was shown to be optimal for
various tasks~\cite{Crammer05scalablelarge-margin}) for sequence
labeling and $1$-best PA which is the training algorithm MSTParser
uses for non-projective parsing~\cite{McDonaldCP05b}.
Additionally, we include the performance of the Averaged-Perceptron
algorithm on the sequence prediction tasks to show various properties
of all algorithms. It is omitted for parsing as it was shown to be
inferior to PA~\cite{McDonaldCP05b}.  We ran CW with a diagonal
covariance matrix, as full matrix is not feasible.  Specifically, we
used the update rule for full matrices and then removed the
off-diagonal elements\footnote{This was shown to perform the best
  compared with two other alternatives: update a full {\em inverse}
  $\msigma$ and remove its off-diagonal elements, and compute an exact
  update for a diagonal covariance matrix.}. We used parameter
averaging with all methods, including CW, as it improved performance
for all algorithms, especially on parsing.

We used 10-fold cross validation for sequence labeling and existing
split of data into training, development and test set for parsing.
Hyper-parameters ($\phi$ for CW, $C$ for PA) were tuned for each
sequence prediction task using a single run over a random split of the
data into a three-fourths training set and a one-fourth test set and
using a development set with $200$ sentences per language for
parsing. All algorithms were executed for ten (10) iterations over the training set.

\subsection{Sequence Labeling Performance Evaluation}

\begin{figure}[!t!]
\begin{centering}
\begin{tabular}{cc}
\subfigure[NP Chunking]{\includegraphics[width=0.45\textwidth]{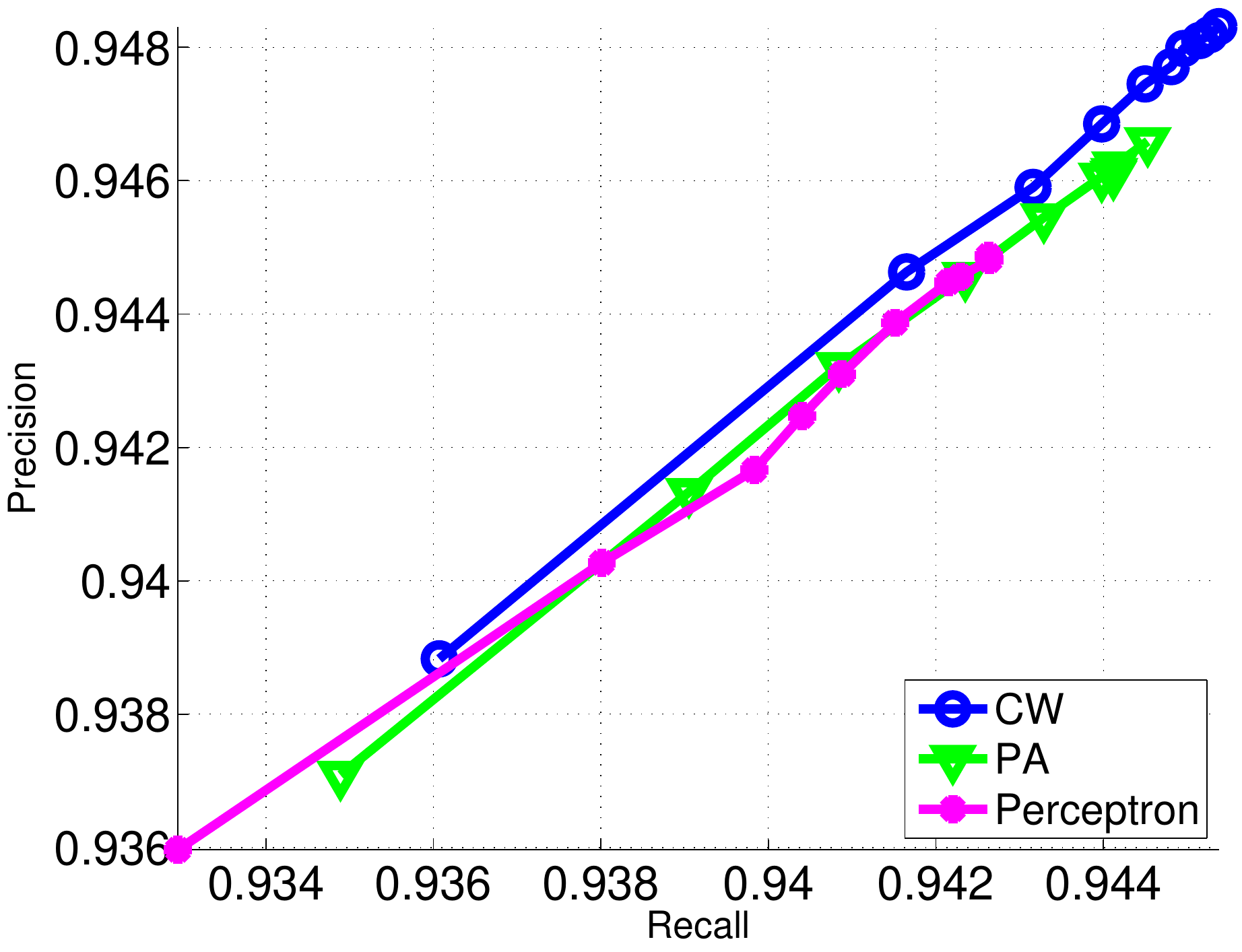}}&
\subfigure[NER English]{\includegraphics[width=0.45\textwidth]{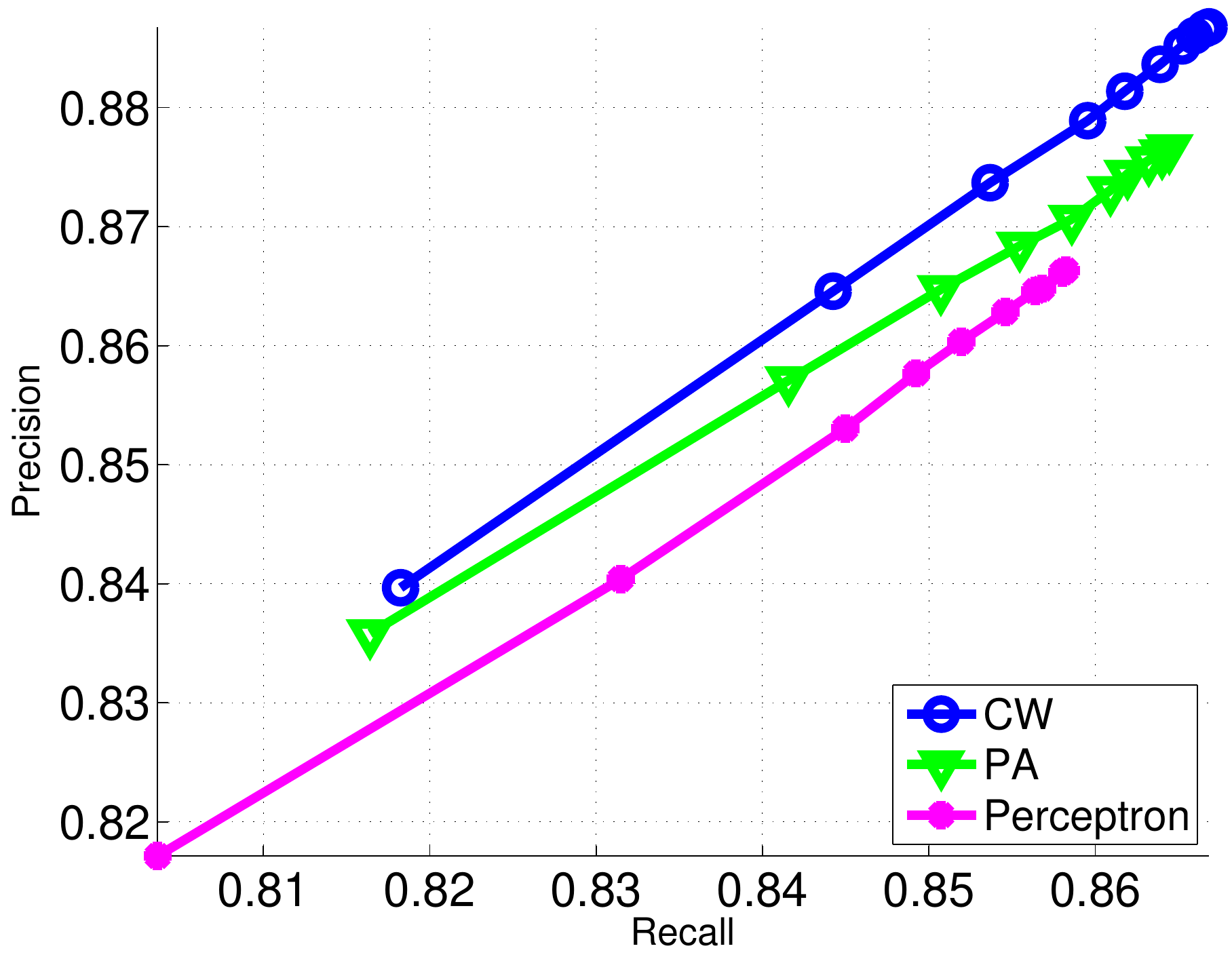}}\\
\subfigure[NER Dutch]{\includegraphics[width=0.45\textwidth]{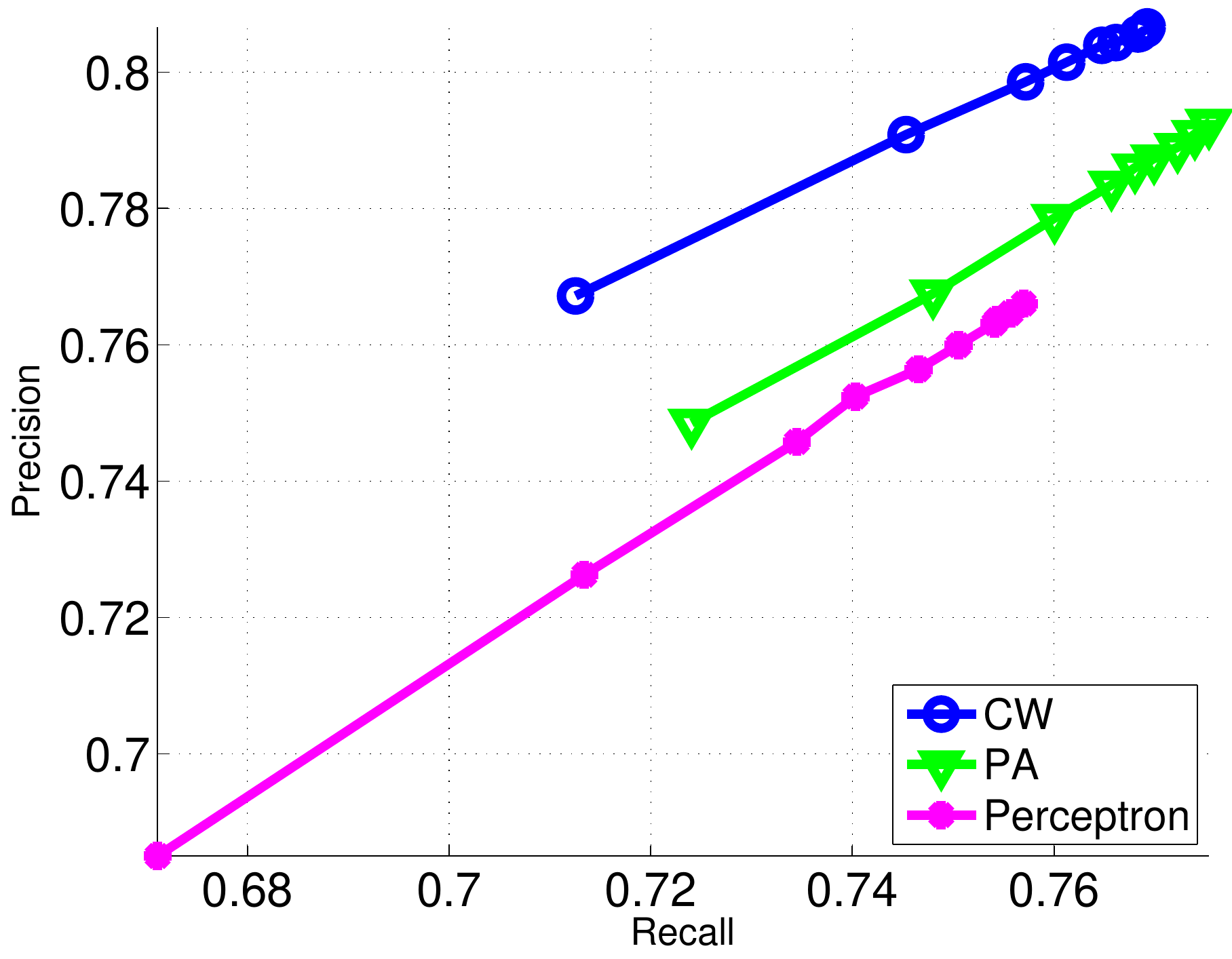}}&
\subfigure[NER Spanish]{\includegraphics[width=0.45\textwidth]{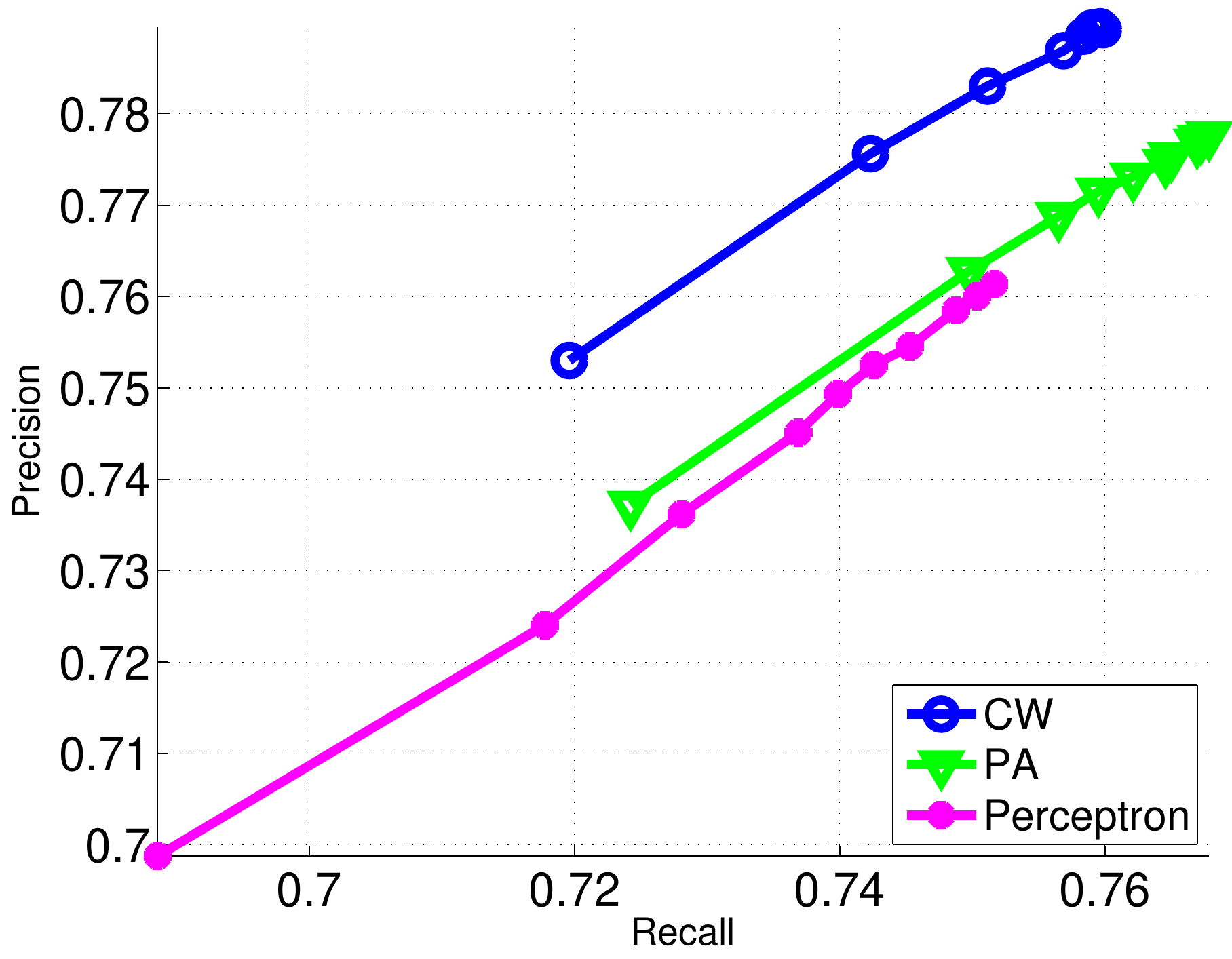}}
\end{tabular}
\caption{Precision and Recall on four datasets (four panels). Each
  connected set of ten points corresponds to the performance of a
  specific algorithm after each of the 10 iterations, increasing from bottom-left to top-right.}
\label{fig:sequences_inc_training}
\end{centering}
\end{figure}

The F-measure of all algorithms after 10 iterations
is summarized in Table~4.
In all four datasets CW algorithm outperforms PA that outperforms the
Perceptron algorithm. The difference between CW and the Perceptron is statistically
significant using paired t-test over the 10 folds, and between CW and PA it is significant in one dataset.

We further investigate the convergence properties of the algorithms in
\figref{fig:sequences_inc_training} in which we plot the recall and precision
evaluated after each training round averaged across the $10$ folds. Each
panel summarizes the results for a single dataset, and in each panel a
single set of connected points corresponds to one
algorithm. Points in the left-bottom of the plot correspond to early
iterations and points in the right-top correspond to later
iterations. Long segments indicate a big improvement in performance
between two consecutive iterations.

High (in the y-axis) values indicate better precision and right (in
the x-axis) values indicate better recall. The performance of all
algorithms is converging in about $10$ iterations as indicated by the
fact the points in the top-right of the plot are close to each
other. The long segments in the bottom-left for the Perceptron
algorithm indicate that this algorithm benefits more from more than
one pass compared with the other algorithms.  Interestingly, in NER
Dutch and NER Spanish (two bottom panels), PA achieves slightly better
recall than CW but is paying in terms of precision and overall
F-measure performance.

\subsection{Dependency Parsing Performance Evaluation}

\begin{table}[!t!]
\begin{center}
{
\begin{tabular}{|l|c|c|c|c|}
\hline
&\multicolumn{2}{c|}{Edge Accuracy}&\multicolumn{2}{c|}{Complete Trees}\tabularnewline
Dataset & CW  & 1-best PA & CW  & 1-best PA \tabularnewline
\hline
\hline
Arabic 	& 0.777	& 0.772 & 0.110	& 0.123\tabularnewline
\hline
Bulgarian 	& 0.899	& 0.898 & 0.410	& 0.397\tabularnewline
\hline
Chinese 	& 0.901	& 0.900 & 0.739	& 0.737\tabularnewline
\hline
Czech 	& 0.845	& 0.844 & 0.340	& 0.323\tabularnewline
\hline
Danish 	& 0.878	& 0.871 & 0.326	& 0.304\tabularnewline
\hline
Dutch 	& 0.830	& 0.831 & 0.282	& 0.282\tabularnewline
\hline
English 	& 0.888	& 0.889 & 0.292	& 0.287\tabularnewline
\hline
German 	& 0.886	& 0.888 & 0.412	& 0.409\tabularnewline
\hline
Japanese 	& 0.936	& 0.940 & 0.769	& 0.779\tabularnewline
\hline
Portuguese 	& 0.863	& 0.863 & 0.302	& 0.302\tabularnewline
\hline
Slovene 	& 0.782	& 0.777 & 0.266	& 0.256\tabularnewline
\hline
Spanish 	& 0.820	& 0.813 & 0.189	& 0.180\tabularnewline
\hline
Swedish 	& 0.865	& 0.866 & 0.404	& 0.391\tabularnewline
\hline
Turkish 	& 0.781	& 0.776 & 0.281	& 0.276\tabularnewline
\hline
\hline
\textbf{Average}	& \textbf{0.854}	& 0.852 & \textbf{0.366}	& 0.360\tabularnewline
\hline
\end{tabular}
\label{table:parsing_results}
\caption{Accuracy of predicted edges (two left columns) and percentage of complete trees (two right columns) of parser trained with CW and PA.}
}
\end{center}
\end{table}

\begin{figure}[!t!]
\begin{centering}
\subfigure{\includegraphics[width=0.60\textwidth]{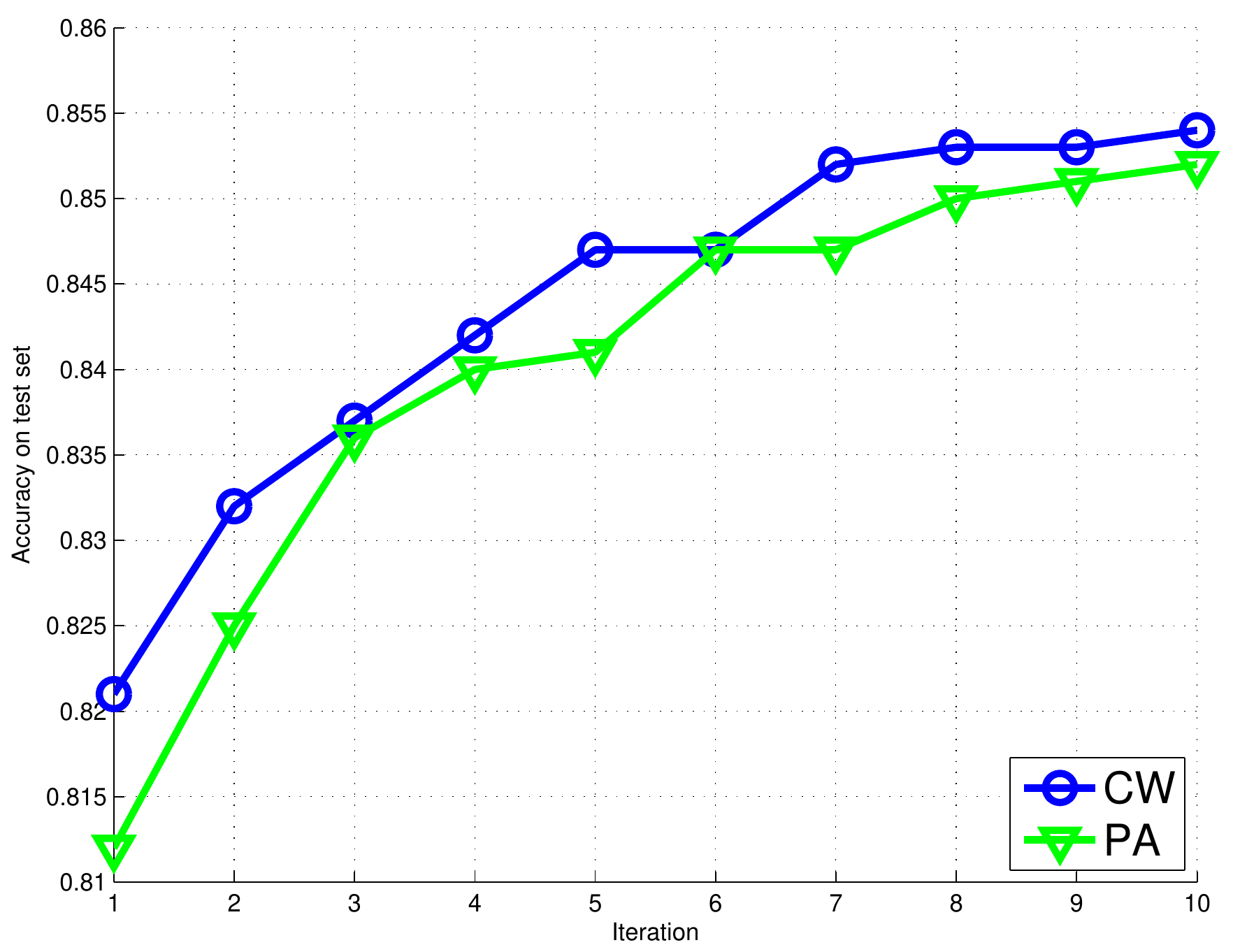}}
\caption{Parsing accuracy of PA and CW vs. iteration averaged over all
  $14$ languages.}
\label{fig:parsing_inc_training}
\end{centering}
\end{figure}

Predicted edges accuracy of PA and CW
are 
summarized in Table~5. 
The accuracy ranges from $77\%$ on Arabic to $94\%$ on Japanese, with an average of
$85\%$. Training the parser with CW algorithm compared to PA yield a
small accuracy improvement in 8 of 14 languages, with maximal
improvement of $0.7\%$ for Danish and Spanish and maximal degradation
of $0.4\%$ for Japanese. The accuracy averaged over all the languages
using CW is $85.4\%$ compared to $85.2\%$ achieved by PA. The
percentage of $complete-trees$, that is sentences where the parse tree
was completely correct, was also improved by CW compared to PA in $10$
of $14$ languages, and averaged over all the languages the parser
trained with CW got $36.6\%$ complete trees compared to $36.0\%$.

We investigate the convergence results of the two algorithm in
\figref{fig:parsing_inc_training} where we plot the accuracy results
(averaged over all the language) evaluated using the test set after
each training iteration. After a single pass over the training data CW
algorithm performs better than PA, achieving higher accuracy on 11 of
14 datasets and average accuracy of $82.1\%$ compared to
$81.2\%$. With additional passes over training data the performance
gap is reduced, until finally after ten iteration PA closes most of
the gap and achieves average accuracy lower than CW by only $0.2\%$.

To conclude, in all tasks of sequence labeling and most tasks of
dependency parsing, CW slightly improves over PA; which is also
reflected in its averaged performance which is slightly better than
the averaged performance of PA. Additionally, CW obtain
high-performance after a single round, and benefit less from multiple
iterations over data, as opposed to both PA and the Perceptron
algorithm. This relation are consistent with previous evaluation of CW
on both binary classification~\cite{DredzeCrPe08,CrammerDrPe08},
multi-class prediction~\cite{CrammerDrKu09} and
phoneme-recognition~\cite{Crammer:2010fk}. Yet, in all these previous
work the improvement of CW over both PA and Perceptron was higher than
the improvement we found here for sequence labeling and
parsing. Currently, it is not clear why CW improves more over PA in
the context of multi-class and binary prediction, and less in the
context of structured prediction. One possible explanation is that
implicitly CW exploits feature statistics. In the former simple
problems the features are orthogonal per class, while in structured
prediction, the features are sum over parts (as in \eqref{features})
and thus may have different statistical properties, such as
dependencies.

\section{Confidence Estimation Methods Evaluation}

We now report an evaluation of the confidence estimation methods
mentioned above. We trained a classifier using the CW algorithm
running for ten (10) iterations on the training set and applied it to
the testing set to obtain an initial labeling, the hype-parameter $\phi$ was set to its optimal value obtained in the
experiments reported in \secref{sec:perf_eval}. We then applied each
of the confidence estimation methods on all the testing set labels.
For sequences, a single split of four-fifths of the data was used as
training set and the remaining one-fifth as testing set. For parsing
we used the given split of the data into a training set and a test set
as described above.  For sequences, the fraction of words for
which the trained model made a mistake ranges between $2\%$ (for NER
Dutch) to $4.1\%$ (for NER Spanish).  While for parsing the fraction of
incorrect edges is between $23\%$ (for Arabic) to $6\%$ (for
Japanese), with an average of $15\%$.

Six algorithms and one baseline were evaluated.
The baseline is random confidence scores for all the labels.
The margin based method called
{\tt Delta} described in \secref{sec:margin}.  The marginal based
method called {\tt Gamma} described in \secref{sec:marginal}, and four
methods based on alternatives described in \secref{sec:k_draw}. Two of
these methods are based on top-K best prediction defined in
\secref{sec:kbest_pred}, one using the output of the prediction algorithm
as is called {\tt {\tt KB}} ($K$-best) defined in
\eqref{k_draw_eq}, and the other is based on weighting its output
called {\tt {\tt WKB}} (weighted $K$-best) defined in
\eqref{k_draw_weighted_eq}. The other two alternatives-based methods
are using stochastic models both described in
\secref{sec:stochastic_models}. The first by inducing a Gaussian
distribution over weights with covariance $s\mi$ called  {\tt {\tt KD-Fix}}
for $K$ draws with fixed covariance, and the second by using the
scaled covariance matrix learned by CW called {\tt KD-PC} for $K$-draws by
parameter confidence.

For the {\tt KD-PC} algorithm we note that the predictions of the CW
algorithm are based solely on the mean weight vector $\hvy=\arg\max_\vz
\vmu\cdot\fun(\vx,\vz)$ and are invariant to initial scale $a$ of the
covariance $a\mi$ (as was noted
elsewhere~\cite{CrammerDrPe08}). Nevertheless, for the purpose of
confidence estimation the scale of the covariance $\msigma$ has a huge
impact. Small eigenvalue of $\msigma$ yield that all the samples of
$Z$ in \eqref{z} will be the same, while large values yield almost
complete random vectors, ignoring the mean.

One possible simple option is to run CW few times with few possible
initializations of the covariance $\msigma$ and choose one copy based
on the confidence evaluated on the training set. However, since the
actual predictions of all these versions is the same and all the
resulting covariance matrices will be proportional to each other
\cite[Lemma 3]{CrammerDrPe08} in practice we run the algorithm only
once initializing the covariance with $\mi$. Then, after training is
completed, we pick the best covariance matrix of the form $s \msigma$
for a positive scalar $s$ where $\msigma$ is the covariance output
with initialization $\mi$ , and choose the best value $s$ using the
training set.

The parameters of the confidence estimation methods: size of $K$ of
the number of labelings used in the four first methods ({\tt KD-PC},
{\tt {\tt {\tt KD-Fix}}}, {\tt KB}, {\tt {\tt WKB}}), the weighting
scalar $s$ used in {\tt KD-PC} and {\tt {\tt {\tt KD-Fix}}}, and the
coefficient $c$ of the {\tt Gamma} method were tuned for each dataset
on a development set according to the best measured {\em average
  precision} achieved in the task of incorrect prediction detection,
described in \secref{sec:rel_con}.

We tried $20$ values in the range $0.01$ to $1.0$ for the parameter
$s$. For the number of labeling $K$ the values in $10,20 \dots 80$
were used. For the K-Draws methods, larger $K$ generally improved
performance up to about $K=50$ with flat performance beyond that, so
$K=50$ was set for all datasets (see also \secref{sec:effect_k}).  The
{\tt K-Best} and {\tt WK-Best} methods are more sensitive to value of
$K$ and values between $10$ to $30$ were used across the different
datasets. Performance degraded significantly for larger values of $K$.
For the $c$ parameter of {\tt Gamma} method $30$ values between $0.01$
to $3.0$ were tried.

We evaluate the algorithms in two aspects of confidence: relative
confidence (\secref{sec:rel_con}) and absolute confidence
(\secref{sec:abs_con}) and two application: precision-recall tradeoff
(\secref{sec:prec_recall_trade}) and active learning
(\secref{sec:active}).

\subsection{Relative Confidence}
\label{sec:rel_con}

In this experiment confidence estimation methods are evaluated in
accordance to their ability to rank all the words in the test set (per
dataset) having words for which there is a prediction mistake in the
top and the correct predictions in the bottom.  Conceptually, this
task can be thought of as a retrieval task of the erroneous words.
All words were ranked from low to high according to the confidence
score in the prediction associated with each word by the various
confidence methods. Then their performance in the task was evaluated
in a few ways. We split the results according to the task type.

\subsubsection{Sequence Labeling}
\begin{figure}[!t!]
\begin{centering}
\subfigure{\includegraphics[width=0.48\textwidth]{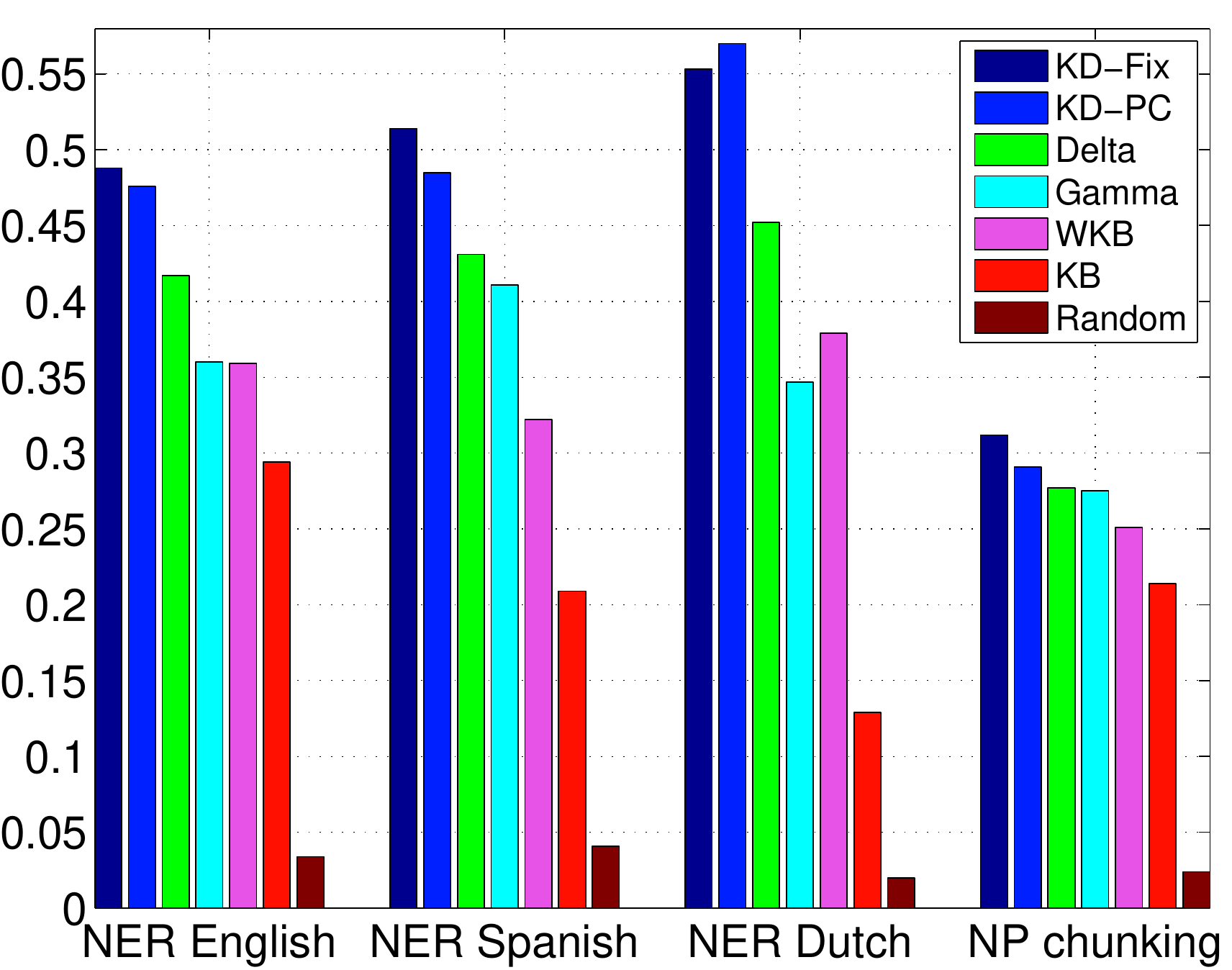}}
\subfigure{\includegraphics[width=0.48\textwidth]{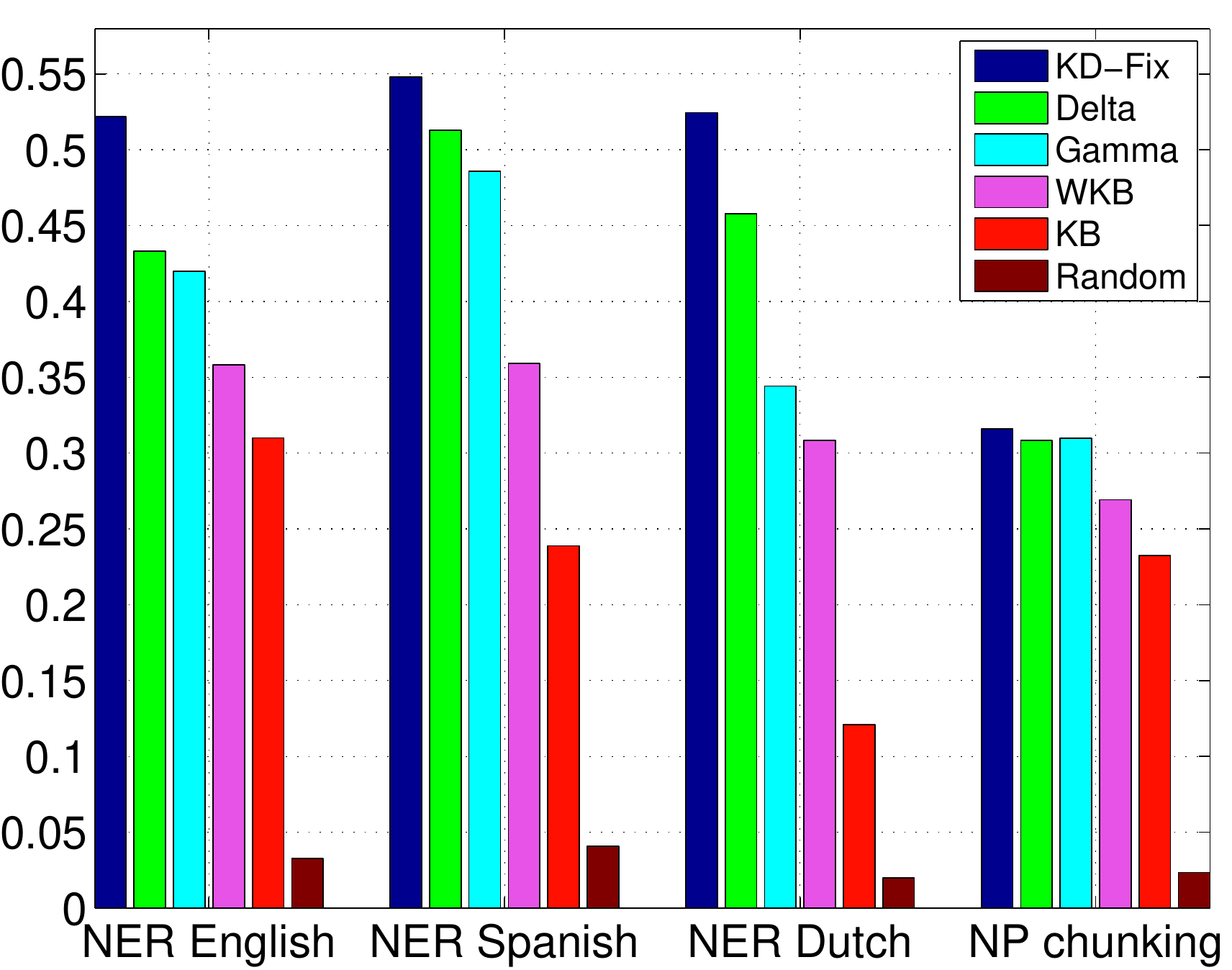}}
\caption{Average precision of rankings of the words
  of the test-set according to confidence in the prediction of all
  methods (left to right bars in each group): {\tt {\tt KD-Fix}}, {\tt KD-PC}, {\tt Delta}, {\tt Gamma},
  {\tt WKB}, {\tt KB} and random ordering, when training with the CW algorithm
  (left) and the PA algorithm (right).}
\label{fig:sequences_avg_precision_CW_PA}
\end{centering}
\end{figure}

\begin{figure}[!t!]
\begin{centering}
\begin{tabular}{cc}
\subfigure{\includegraphics[width=0.48\textwidth]{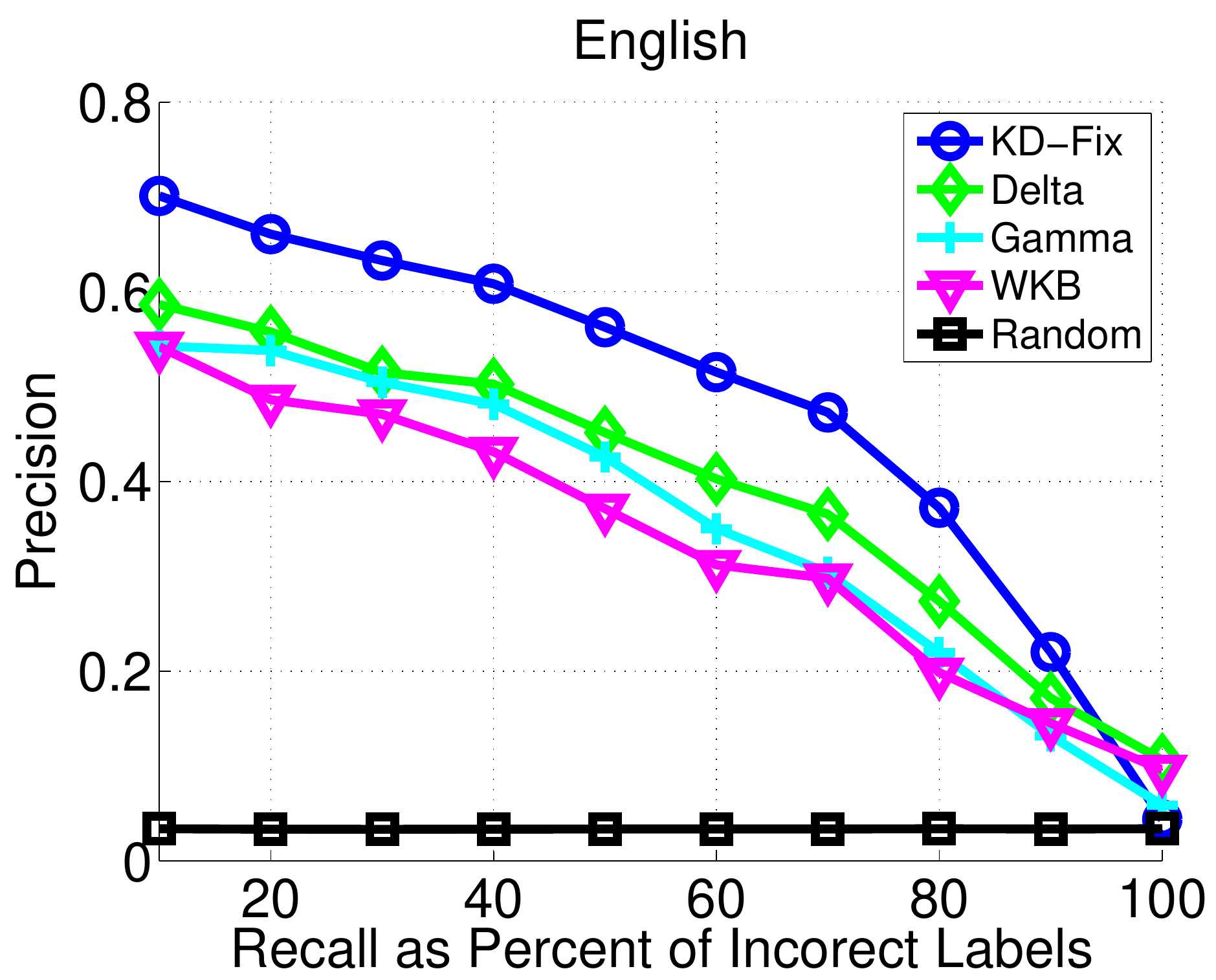}}&
\subfigure{\includegraphics[width=0.48\textwidth]{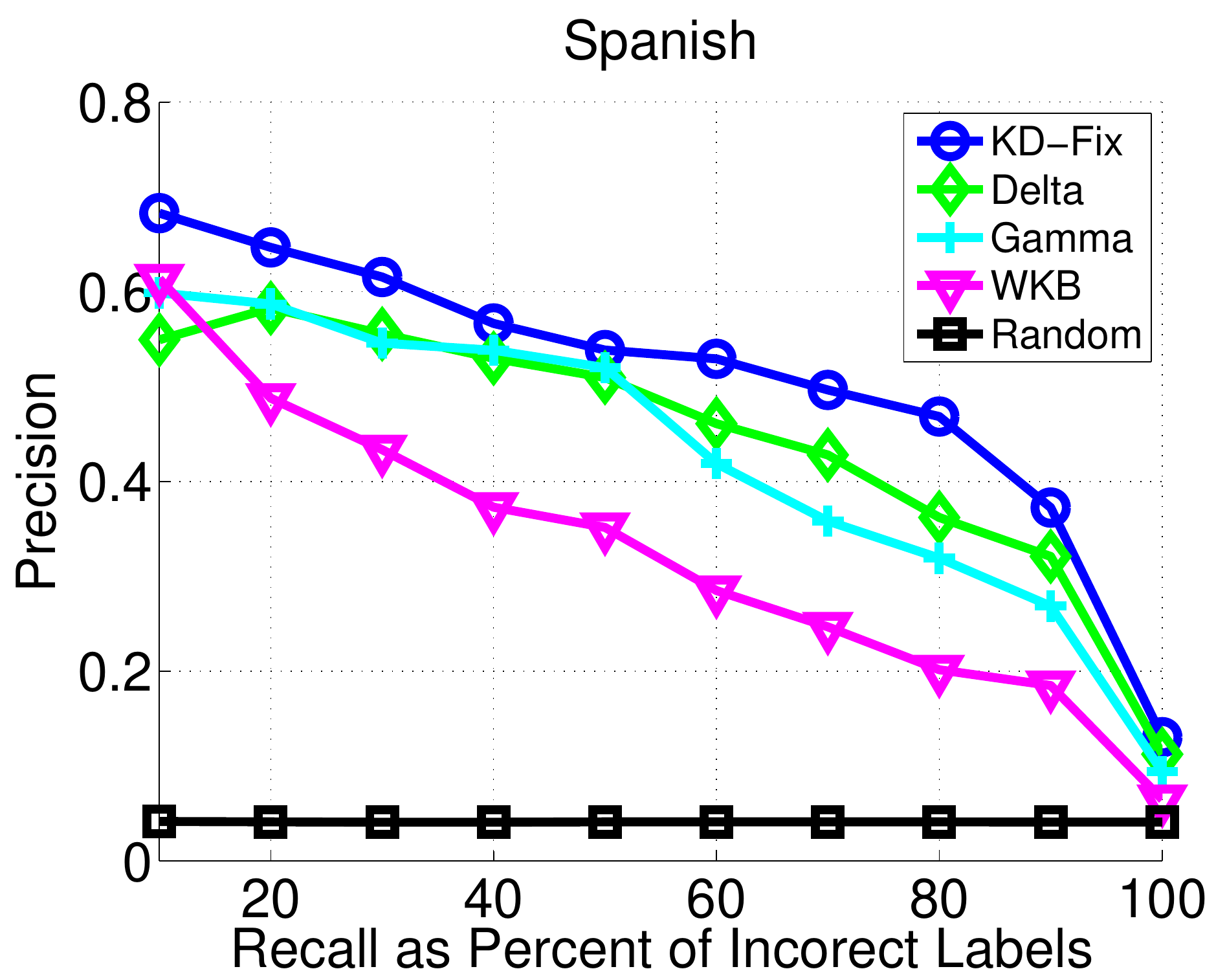}}\\
\subfigure{\includegraphics[width=0.48\textwidth]{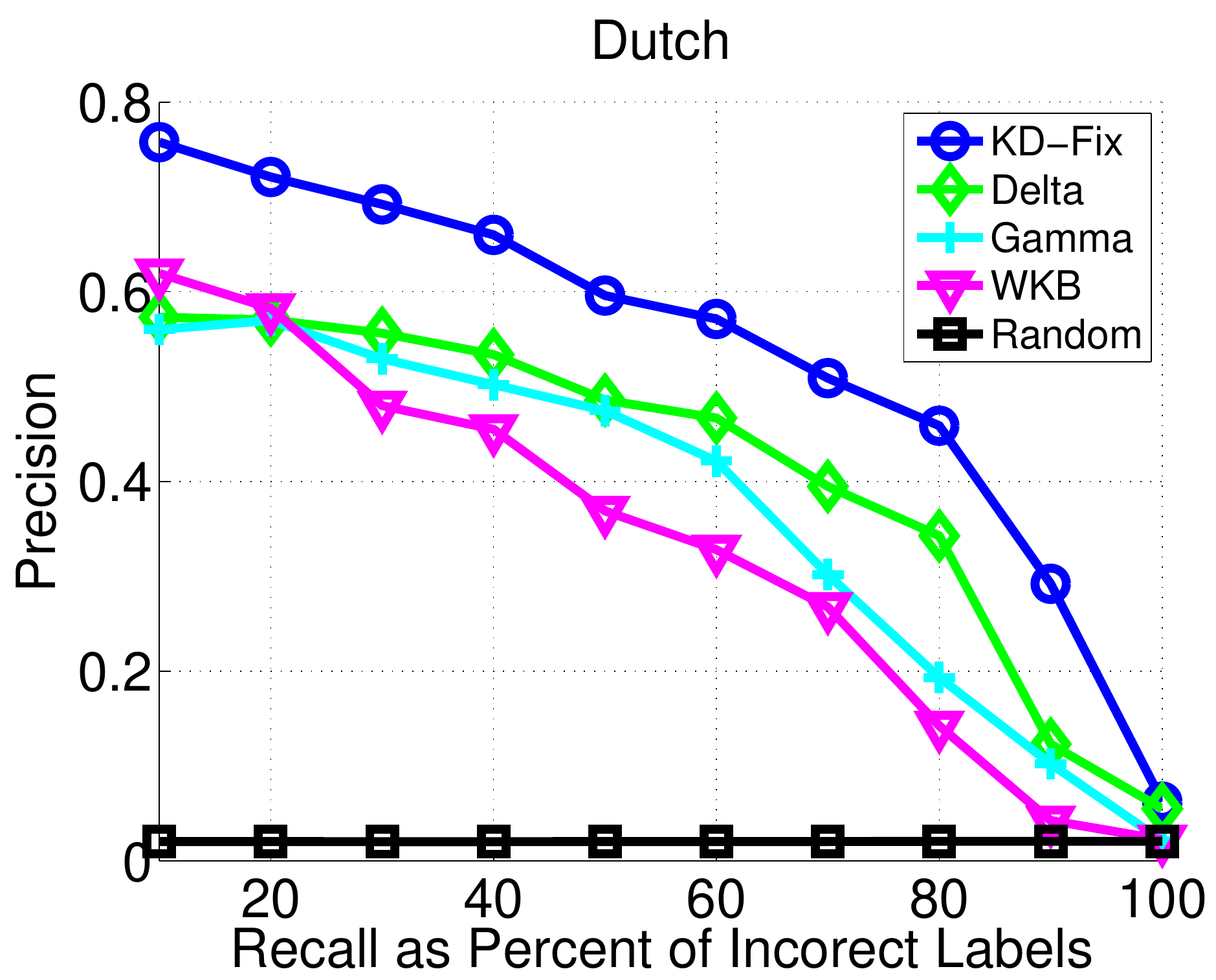}}&
\subfigure{\includegraphics[width=0.48\textwidth]{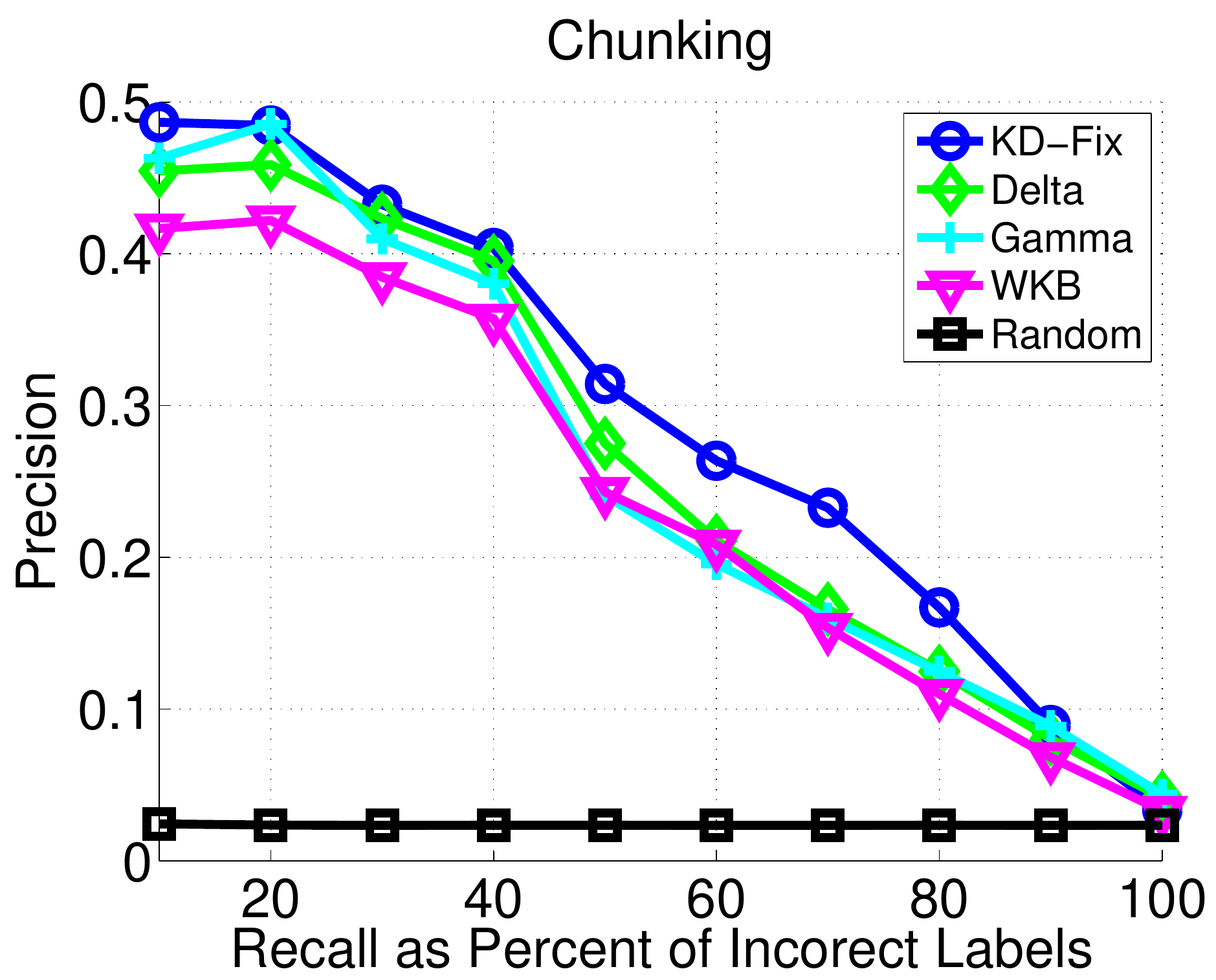}}
\end{tabular}
\caption{Precision in detection of incorrect labels as recall
  increases on three NER tasks and NP chunking in English.}
\label{fig:sequences_detection_precision_recall_CW}
\end{centering}
\end{figure}

The average precision for ranking the words of the test-set according
the confidence in the prediction for the seven methods in sequence
labeling is summarized in the left panel of
\figref{fig:sequences_avg_precision_CW_PA} when training with CW. The
algorithms are ordered from left-to-right: {\tt {\tt KD-Fix}}, {\tt
  KD-PC}, {\tt Delta}, {\tt Gamma}, {\tt WKB}, {\tt KB} and random
ordering. The average precision is computed by averaging the
individual precision values computed at all ranks of erroneous words.

From the plot we observe that when having a random ordering the
average precision is about the frequency of erroneous word, and
clearly random ordering achieves the lowest (worst) average
precision. The next two best methods are these based on K-Best predictions , where the weighted approach {\tt WKB} outperforms the
non-weighted version {\tt KB}. Thus, using the actual score value
into consideration improves the ability to detect erroneous
words. Next in performance are {\tt Gamma} and {\tt Delta}, the
margin-based and marginal-probabilities methods that outperform the
K-best methods in all four datasets. {\tt Delta} is better than {\tt
  Gamma} in two of the dataset and equal in the other two.  The two
best performing methods are {\tt KD-Fix} and {\tt KD-PC}, where
the former is better in three out of four datasets.  The relative
success of {\tt KD-Fix} compared to {\tt KD-PC} is surprising,
as {\tt KD-Fix} does not take into consideration the actual
uncertainty in the parameters learned by CW, and in fact replaces it
with a {\em fixed} value across all features.

Except for {\tt KD-PC} that takes the parameter confidence information
into consideration, all the other confidence estimation methods do not
assume a confidence-based learning approach. We thus repeated the
experiment using the passive-aggressive
algorithm (PA) rather than CW for training the model. The results appear in the right panel
of \figref{fig:sequences_avg_precision_CW_PA} and basically tell the
same story: {\tt KD-Fix} outperforms the margin-based and
marginal-probabilities methods {\tt Delta} and {\tt Gamma}, and the
K-Best Viterbi based methods, {\tt KB} and {\tt WKB} achieve lowest
performance with the weighted version better than the
non-weighted. Note that in general CW is slightly better than PA in
the prediction task (\secref{sec:perf_eval}) and thus retrieval of
erroneous words on the set labeled by the PA model is slightly an easier task,
this may explain some
of the bars in the right panel are higher than their corresponding bars in the left panel.

Average precision does not tell the whole story - it encapsulates the
detection of all the incorrect edges into a single number. More
refined analysis is described via precision-recall (PR) plots showing
the precision as more incorrect labels are detected. PR plots for
model trained with CW algorithm are shown in
\figref{fig:sequences_detection_precision_recall_CW}. The plots for
{\tt KD-PC} and {\tt KB} are omitted for clarity, {\tt KD-PC} curve
is very similar to {\tt KD-Fix} and {\tt KB} is worse than all the
rest. The plots present the incorrect-labels detection precision in
different recall values from $10\%$ to $100\%$.  We observe that the
advantage of the K-Draws methods over other methods is consistent
throughout the entire retrieval process.
Interestingly, for low recall values of around $10\%$, in NER in Dutch and Spanish, {\tt  WKB} performed better than {\tt Delta} and {\tt Gamma}, yet its performance quickly degraded for higher recall values.

To illustrate the effectiveness of the incorrect labels detection
process \tabref{table:seq_incorrect_label_detection_CW} presents the
number of incorrect labels detected vs. number of labels inspected for
English NER dataset. The test set for this task includes $50K$ words
and the classifier made mistake on only $1,650$ words, that is,
accuracy of $96.7\%$. We show the number of incorrect labels detected
after inspecting $500$, $2,500$ and $5,000$ labels which are $1, 5$
and $10\%$ of all labels. When using random inspection, the number of
incorrect labels detected is, as expected, $1\%, 5\%$ and $10\%$ of all
mislabeled words. Yet when inspecting the labels according to the
ranking induced by {\tt KD-Fix} method, $20\%, 70\%$ and $86\%$ of all
mislabeled words were detected for the same effort.

\begin{table}
\begin{centering}
\begin{tabular}{|c|c|c|c|}
\hline
Words inspected &{\tt KD-Fix} & {\tt Delta} & Random \tabularnewline
(\% of total)   & & & \tabularnewline
\hline
\hline
500 (1\%)       & \textbf{336 (20\%)} & 278 (17\%) & 15 (0.9\%) \tabularnewline
\hline
2,500 (5\%)     & \textbf{1,148 (70\%)} & 1,003 (61\%) & 83 (5\%) \tabularnewline
\hline
5,000 (10\%)    & \textbf{1,415 (86\%)} & 1,310 (79\%)& 164 (9.9\%)\tabularnewline
\hline
\end{tabular}
\caption{Number of incorrect labels detected, and the corresponding percentage of \emph{all mistakes}, after inspecting $500-5,000$ labels which are $1-10\%$ of all labels, using random ranking and ranking induced by {\tt KD-Fix} and {\tt Delta} methods.}
\label{table:seq_incorrect_label_detection_CW}
\end{centering}
\end{table}

\subsubsection{Dependency Parsing}
\begin{figure}[!t!]
\begin{center}
\begin{tabular}{cc}
\subfigure{\includegraphics[width=0.48\textwidth]{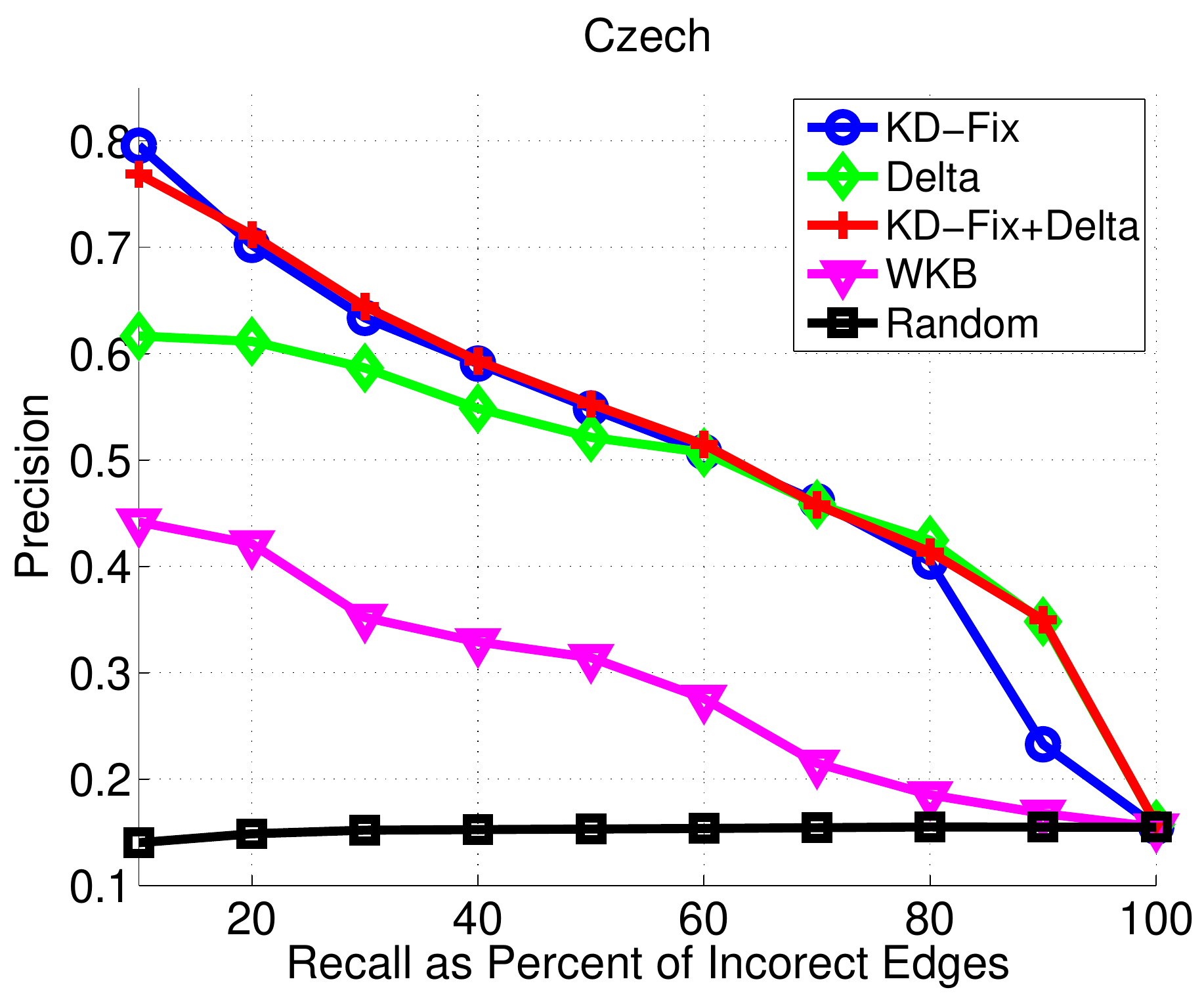}}&
\subfigure{\includegraphics[width=0.48\textwidth]{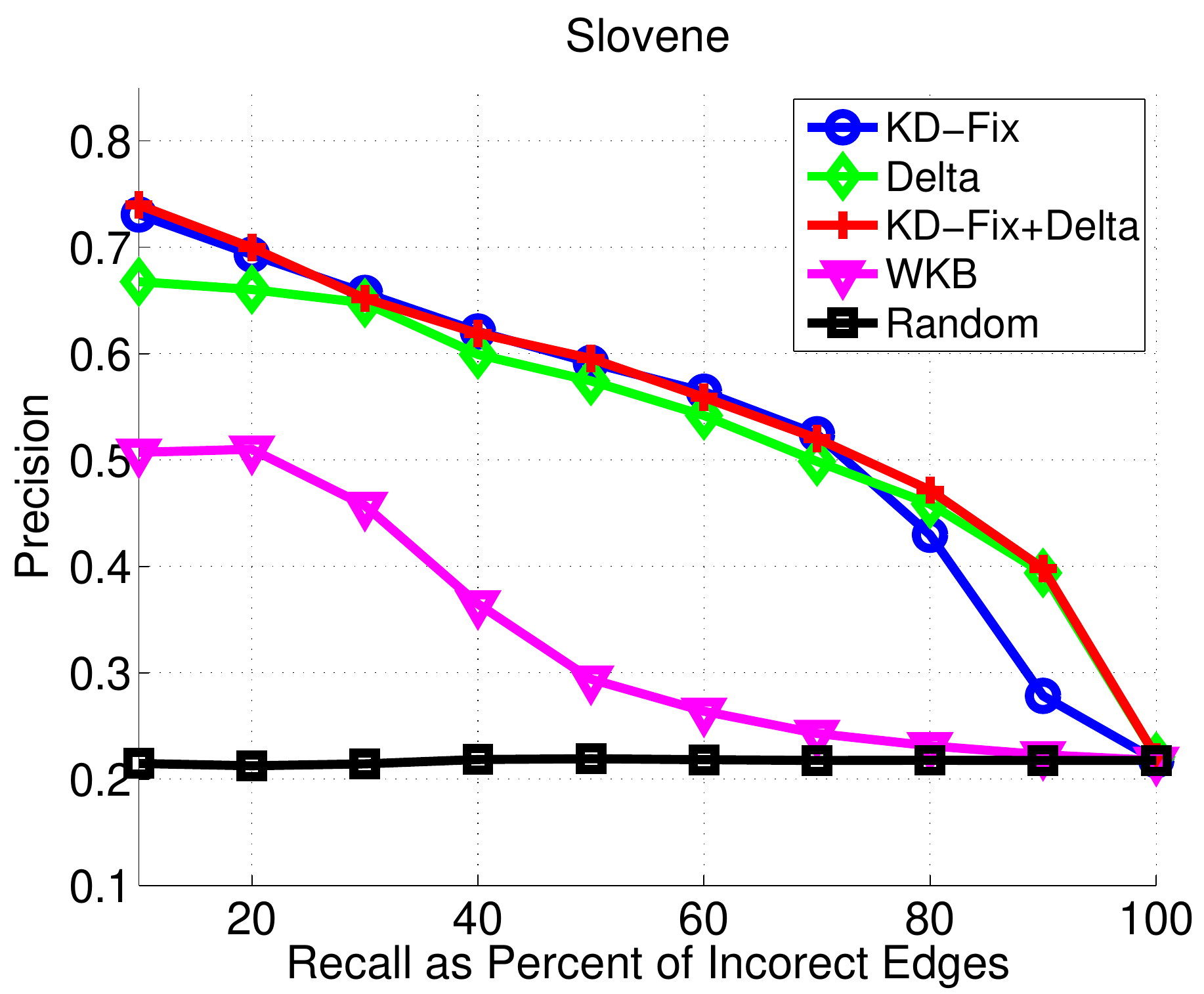}}\\
\subfigure{\includegraphics[width=0.48\textwidth]{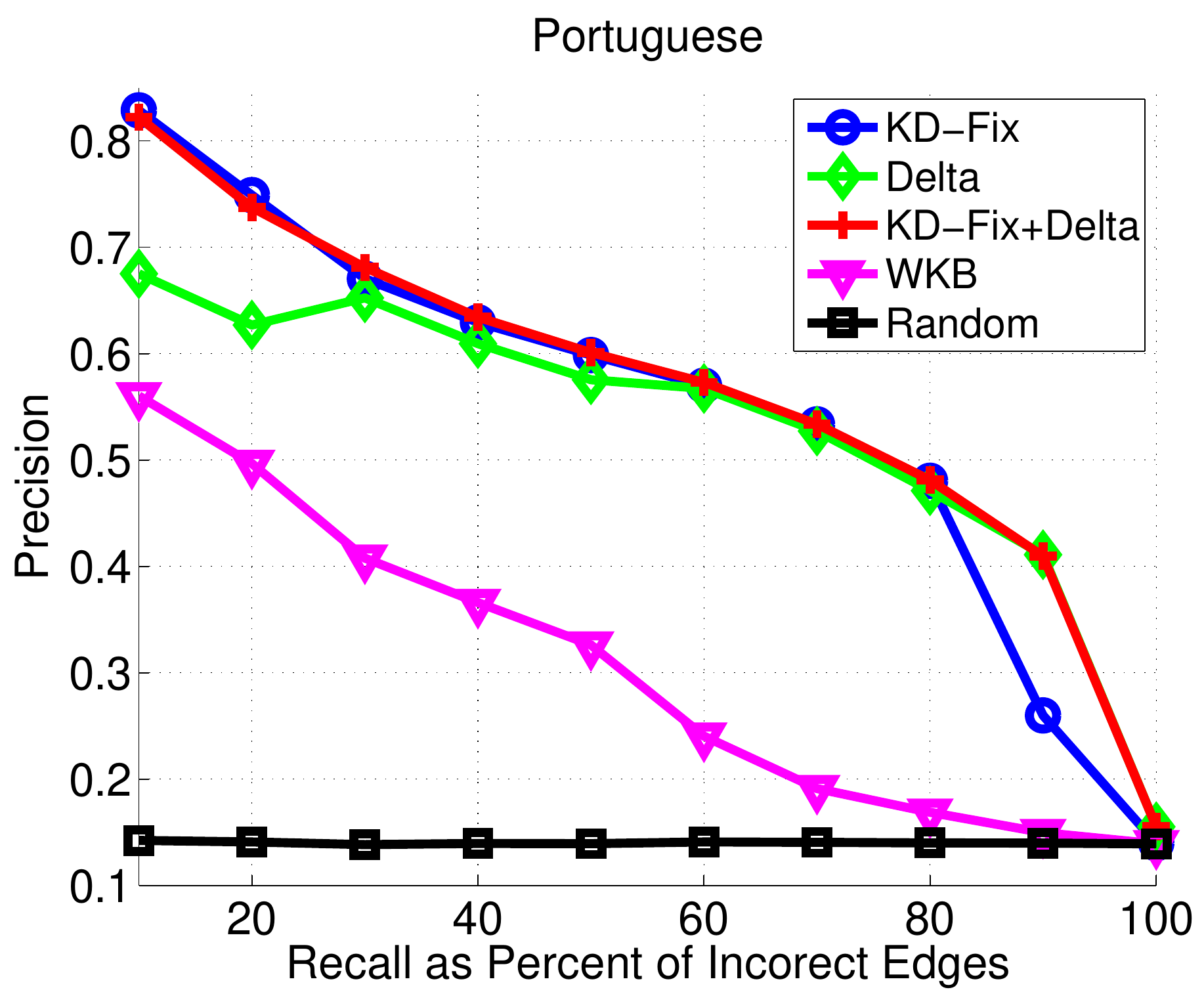}}&
\subfigure{\includegraphics[width=0.48\textwidth]{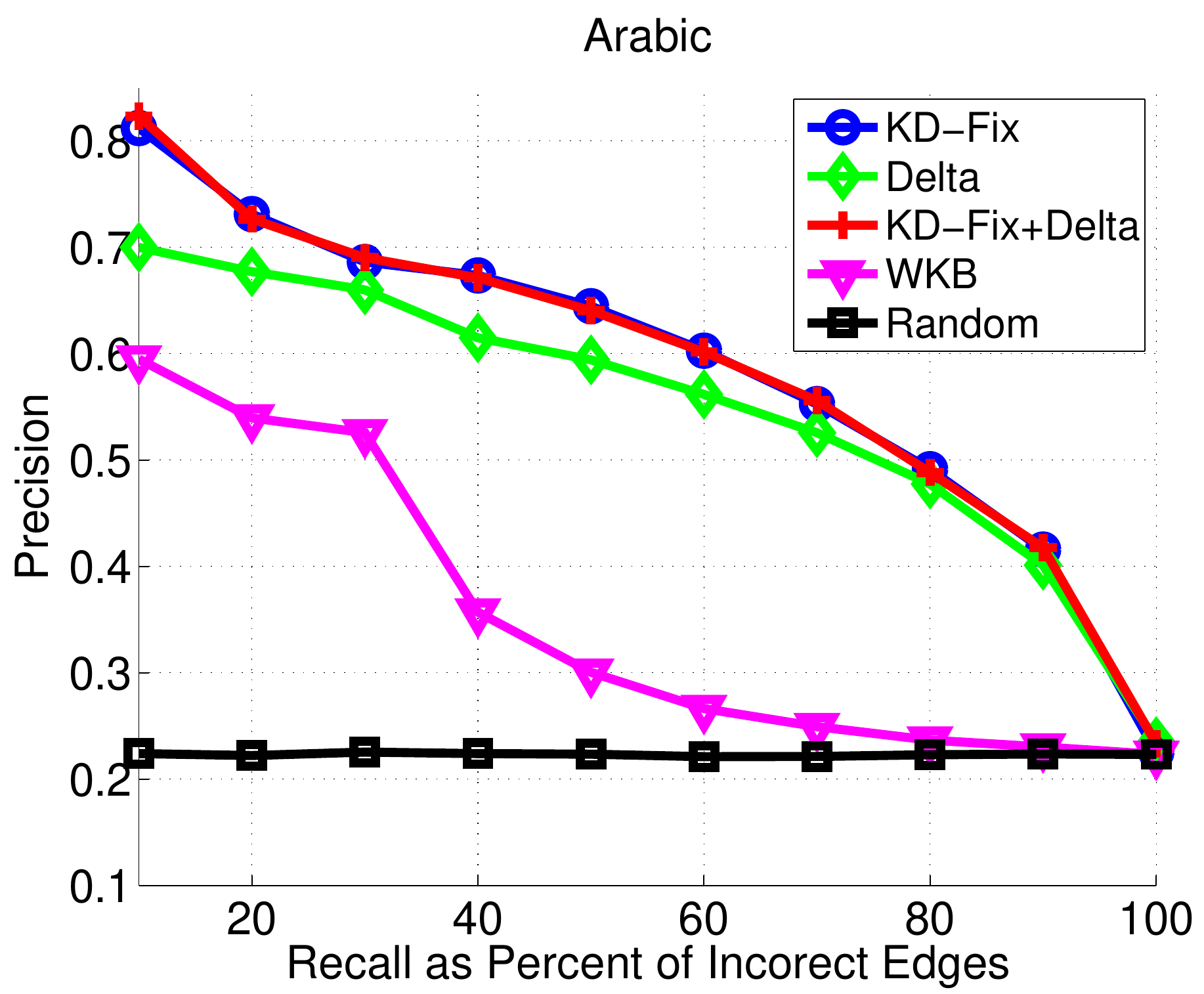}}
\end{tabular}
\caption{{Precision in detection of incorrect edges as recall
    increases on four dependency parsing datasets. {\tt KD-PC} curves
    are very similar to {\tt KD-Fix} and omitted for clarity.
}}
\label{fig:parsing_precision_recall_CW}
\end{center}
\end{figure}

We applied the same methodology for evaluating the confidence
estimation methods in the task of dependency parsing.  Here, all
predicted edges were ranked according to their confidence score
ordered from low to high, and ideally, erroneous edges by the parser
are ranked at the top.
A summary of the average precision, computed at all ranks of erroneous edges,
evaluated for all confidence estimation methods and all the $14$
datasets is summarized in \tabref{table:Parsing_AvgPrecAll_CW}.

The average precision achieved by random ordering is lower than all
the methods and is about equal to the error rate for each
dataset. Next are the K-Best methods, where the weighted version
performs better than the non-weighted version.  The margin-based {\tt
  Delta} method improves significantly over the the K-Best
methods. Finally, {\tt KD-fix} and {\tt KD-PC} methods achieve the best
performance with {\tt KD-Fix} a little better than {\tt KD-PC}, and
both K-Draws methods outperform {\tt Delta} method on average and in
$12$ of $14$ datasets. These results are consistent with the results
observed for sequence labeling.

The Precision-Recall (PR) plots for several datasets are shown in
\figref{fig:parsing_precision_recall_CW} and provide deeper insight of
the mistakes detection precision as more incorrect-edges are
detected. ({\tt KD-PC} plots are very similar to {\tt KD-Fix} and
omitted for clarity.) We observe that in most datasets {\tt KD-Fix}
performs significantly better than {\tt Delta} in the early detection
stage (first $10-20\%$ of the incorrect edges), while {\tt Delta}
performs better in late detection stages - the last $10-20\%$ of the
incorrect edges. The second and third rows of
\tabref{table:Parsing_AvgPrec_CW_PA} summarizes the precision obtained
by all methods after detecting only $10\%$ incorrect edges and after
detecting $90\%$ of the incorrect edges, averaged over all the
datasets. (The first row of \tabref{table:Parsing_AvgPrec_CW_PA}
summarized the average-precision averaged over all $14$ languages, and
is copied from the last row of \tabref{table:Parsing_AvgPrecAll_CW}
for easier reference.)  For example, in three datasets (Czech, Slovene
and Portuguese) of \figref{fig:parsing_precision_recall_CW}, we
observe an advantage of {\tt KD-Fix} for low recall and an advantage
of {\tt Delta} in high recall. This observation is consistent with
most other dataset not shown in these plots.  Yet, in few languages,
Arabic for example, {\tt KD-Fix} outperforms or equal to {\tt Delta}
along the entire range of recall values. This phenomena was not
observed for sequence labeling tasks where we found the K-Draws method
to outperform the other methods throughout the entire retrieval
process.

This phenomena emerges from the different properties of the two
algorithms.  {\tt KD-Fix} assigns at most $K$ distinct confidence
values to each edge - the number of models that agreed on that
particular edge.  As is, there is no mechanism to break ties (in each
of the $K+1$ levels) and thus edges that are assigned with the same
confidence level are ordered {\em randomly} relative to each other.
Furthermore, in most datasets large fraction of the edges, $\sim
70-80\%$, are assigned to one of the top-three possible confidence
scores (i.e. $(K-2)/K, (K-1)/K, 1$). As a results, the precision
performance of {\tt KD-Fix} drops sharply for recall values of $80\%$
and above. This can be seen by the fast decrease of the line with
circle markers in three of plots in
\figref{fig:parsing_precision_recall_CW}, except Arabic. On the other
hand, we hypothesize that the low precision values obtained by {\tt
  Delta} at low recall values (diamond in
\figref{fig:parsing_precision_recall_CW}) is because {\tt Delta} takes
into account only two parses, the margin between the highest scoring
edge (the predicted edge) and the second best edge, ignoring
information about additional edges with score close to the highest
score. In contrast, {\tt KD-Fix} integrates scores
of 
$K$ parse tress. In other words, {\tt Delta} is more sensitive to
small perturbations of score values compared with {\tt KD-Fix}.

Based on this observation we propose combining both {\tt KD-Fix} and
{\tt Delta}. The new method sets the confidence score of an edge to be
a weighted mean 
of the score values of {\tt KD-Fix} and {\tt Delta}, with weights {\em
  a} and {\em 1-a}, respectively, for a value of $a \approx
1$. 
If the confidence value of two edges according to {\tt KD-Fix} is
different, the contribution of the score outputted by {\tt Delta} is
negligible, and the final score is very close to the score of only
{\tt KD-Fix}.  On the other hand, if the score of {\tt KD-Fix} is the
same, as happen for large recall values, then {\tt Delta} 
breaks arbitrary ties. In other words, when ordering all edges
according to the new method, we first order edges according to
confidence score of {\tt KD-Fix}, then a secondary order is employed
according to the confidence values of {\tt Delta} among edges assigned
same confidence score by {\tt KD-Fix}.  Not surprisingly, we name this
method {\em {\tt KD-Fix}+{\tt Delta}}.

This new method enjoys the good of the two methods. As the results
show in \tabref{table:Parsing_AvgPrec_CW_PA} it achieves the highest
average-precision averaged over the $14$ datasets. It improves
average-precision over {\tt KD-Fix} in $12$ of $14$ datasets and over {\tt
  Delta} in all $14$ datasets. From the second and third row of
\tabref{table:Parsing_AvgPrec_CW_PA}, we see that it has Precision
very close to {\tt KD-Fix} for recall of $10\%$ ($0.729$ vs. $0.724$), and
very close to {\tt Delta} for recall of $90\%$ ($0.351$ vs. $0.348$). Moving
to \figref{fig:parsing_precision_recall_CW}, we observe that the curve
associated with the new method (red ticks) is in general as high as
the curves associated with {\tt KD-Fix} for low values of recall, and
as high as the curves associated with {\tt Delta} for large values of
recall.

Finally, similar to sequence labeling task, in dependency parsing all
confidence estimation methods, except for {\tt KD-PC}, can be used
with a model that does not maintain parameters confidence information.
We repeated the experiment but now training a model with the
passive-aggressive algorithm, rather than CW. The results appear in
the fourth row of \tabref{table:Parsing_AvgPrec_CW_PA}.  The results
based on PA are consistent with the results based on CW, {\tt {\tt
    KD-Fix}} outperforms the margin-based and the K-Best trees
methods, and combining {\tt KD-Fix} and {\tt Delta} improves the
performance.

\begin{table}
\begin{centering}
\begin{tabular}{|c|c|c|c|c|c|c|c|}
\hline
&{\tt KD-Fix}&{\tt KD-PC}&{\tt Delta}&{\tt WKB}&{\tt KB}&{\tt KD-Fix}&Random
\tabularnewline
&&&&&&+ {\tt Delta}&
\tabularnewline
\hline
\hline
Arabic	& 0.621	& \textbf{0.623}	& 0.565	& 0.373	& 0.366	& 0.622	& 0.223\tabularnewline
\hline
Bulgarian	& \textbf{0.491}	& 0.466	& 0.463	& 0.257	& 0.238	& 0.494	& 0.102\tabularnewline
\hline
Chinese	& 0.465	& \textbf{0.466}	& 0.452	& 0.190	& 0.150	& 0.473	& 0.102\tabularnewline
\hline
Czech	& 0.539	& \textbf{0.548}	& 0.506	& 0.301	& 0.290	& 0.555	& 0.152\tabularnewline
\hline
Danish	& \textbf{0.502}	& 0.497	& 0.460	& 0.303	& 0.280	& 0.499	& 0.124\tabularnewline
\hline
Dutch	& 0.534	& \textbf{0.536}	& 0.527	& 0.370	& 0.358	& 0.544	& 0.167\tabularnewline
\hline
English	& \textbf{0.489}	& 0.459	& 0.477	& 0.291	& 0.278	& 0.496	& 0.112\tabularnewline
\hline
German	& 0.426	& 0.418	& \textbf{0.469}	& 0.254	& 0.232	& 0.472	& 0.114\tabularnewline
\hline
Japanese	& 0.512	& 0.525	& \textbf{0.535}	& 0.235	& 0.151	& 0.541	& 0.064\tabularnewline
\hline
Portuguese	& \textbf{0.586}	& 0.570	& 0.559	& 0.323	& 0.311	& 0.606	& 0.143\tabularnewline
\hline
Slovene	& 0.561	& \textbf{0.573}	& 0.555	& 0.345	& 0.332	& 0.581	& 0.218\tabularnewline
\hline
Spanish	& \textbf{0.637}	& 0.634	& 0.592	& 0.351	& 0.342	& 0.644	& 0.184\tabularnewline
\hline
Swedish	& 0.528	& \textbf{0.533}	& 0.496	& 0.289	& 0.273	& 0.527	& 0.137\tabularnewline
\hline
Turkish	& \textbf{0.603}	& 0.590	& 0.589	& 0.372	& 0.342	& 0.609	& 0.221\tabularnewline
\hline
\hline
Average	& \textbf{0.535}	& 0.531	& 0.518	& 0.304	& 0.282	& \textbf{0.547}	& 0.147\tabularnewline
\hline
\end{tabular}
\caption{Average precision in ranking all edges according to confidence values.}
\label{table:Parsing_AvgPrecAll_CW}
\end{centering}
\end{table}

\begin{table}
\begin{centering}
\begin{tabular}{|c|c|c|c|c|c|c|c|}
\hline
&{\tt KD-Fix}&{\tt KD-PC}&{\tt Delta}&{\tt WKB}&{\tt KB}&{\tt KD-Fix}&Random
\tabularnewline
&&&&&&+ {\tt Delta}&
\tabularnewline
\hline
\hline
Avg-Prec&\textbf{0.535}&0.531&0.518&0.304&0.282&\textbf{0.547}&0.147\tabularnewline
\hline
Prec @10\%&\textbf{0.729}&0.723&0.644&0.470&0.441&\textbf{0.724}&0.145\tabularnewline
\hline
Prec @90\%&0.270&0.279&\textbf{0.351}&0.157&0.151&\textbf{0.348}&0.147\tabularnewline
\hline
\hline
Avg-Prec (PA) &\textbf{0.539}& - &0.513&0.305&0.278&\textbf{0.548}&0.149\tabularnewline
\hline
\end{tabular}
\caption{Row 1: Average precision in ranking all edges according to confidence values, average over all 14 languages.
Rows 2-3: Precision in detection of incorrect edges when detected 10\% and 90\% of all the incorrect edges.
Row 4: Average precision in ranking all edges according to confidence values, average over all 14 languages, using PA training.}
\label{table:Parsing_AvgPrec_CW_PA}
\end{centering}
\end{table}

\subsection{Absolute Confidence}
\label{sec:abs_con}

\begin{figure}[!t!]
\begin{centering}
\begin{tabular}{cc}
\subfigure[NER English]{\includegraphics[width=0.48\textwidth]{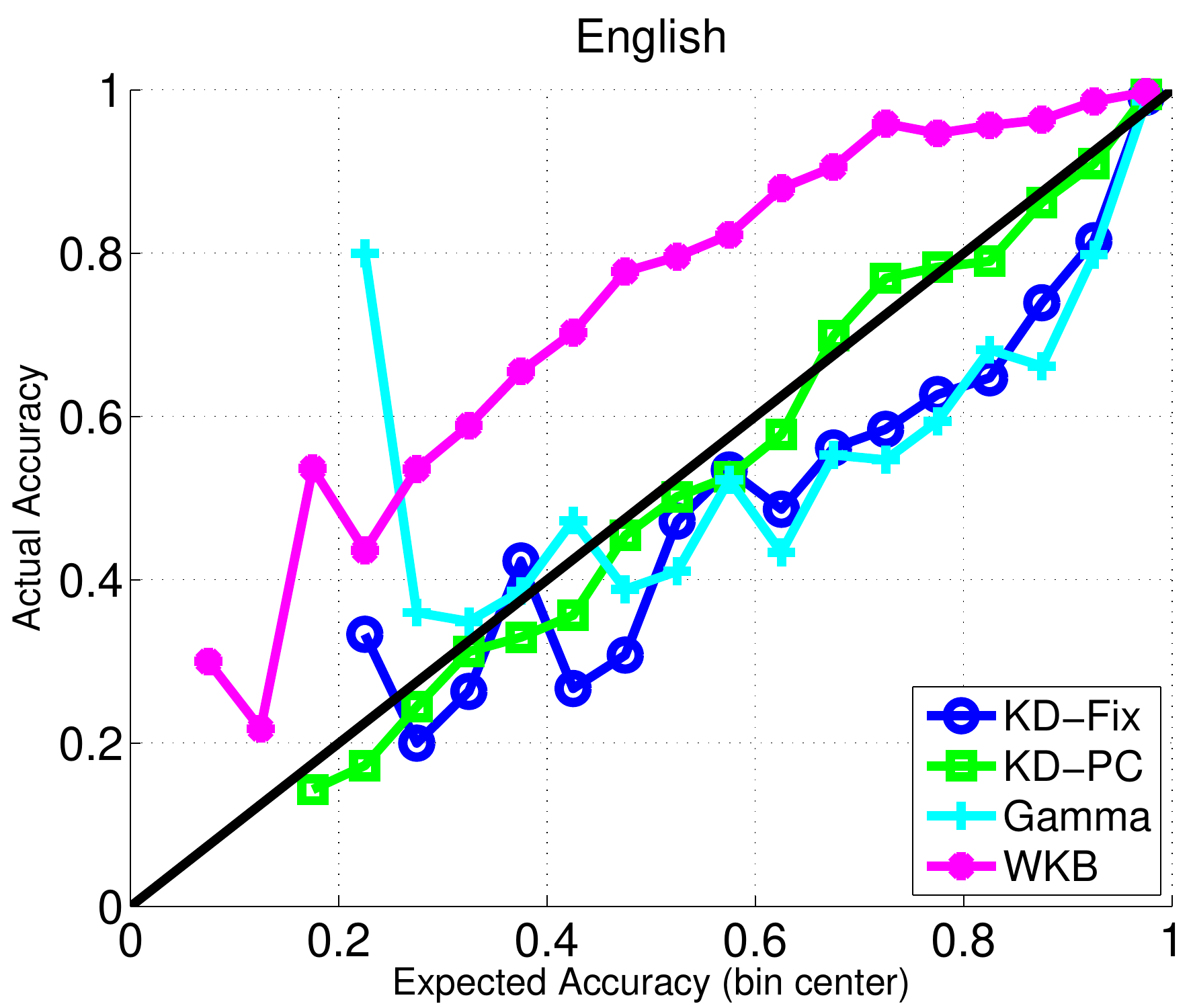}}&
\subfigure[NER Dutch]{\includegraphics[width=0.48\textwidth]{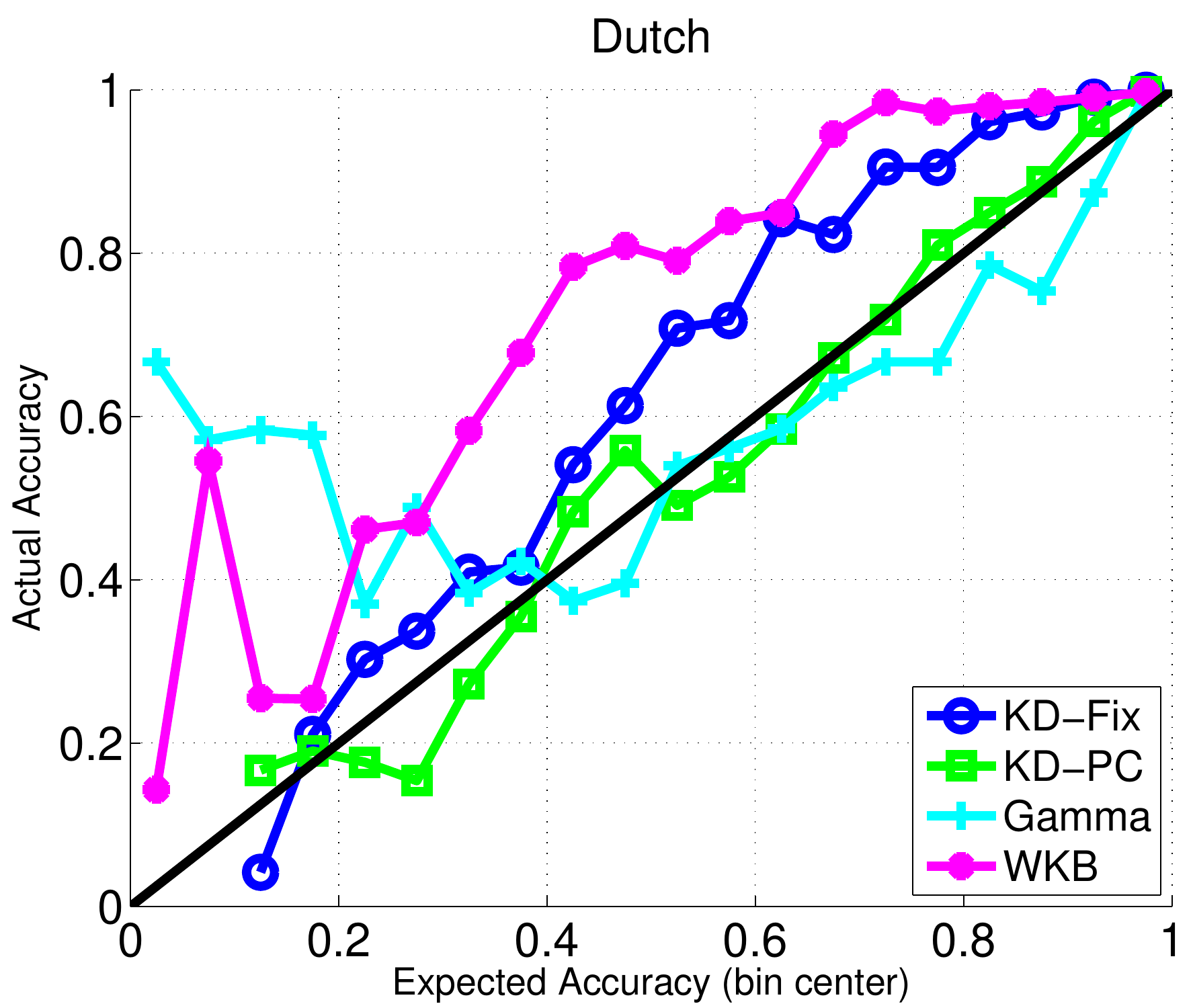}}\\
\subfigure[NER Spanish]{\includegraphics[width=0.48\textwidth]{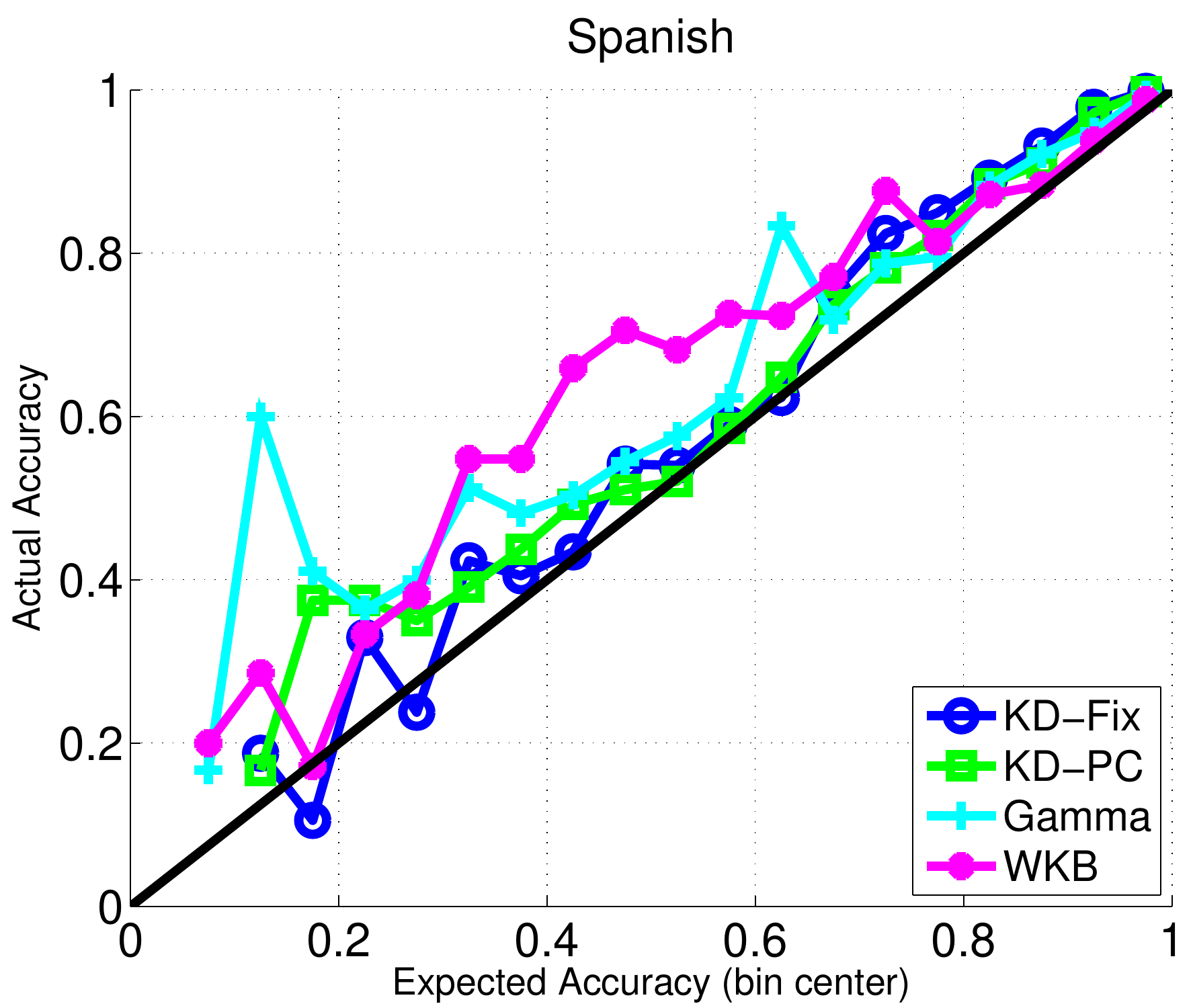}}&
\subfigure[NP
Chunking]{\includegraphics[width=0.48\textwidth]{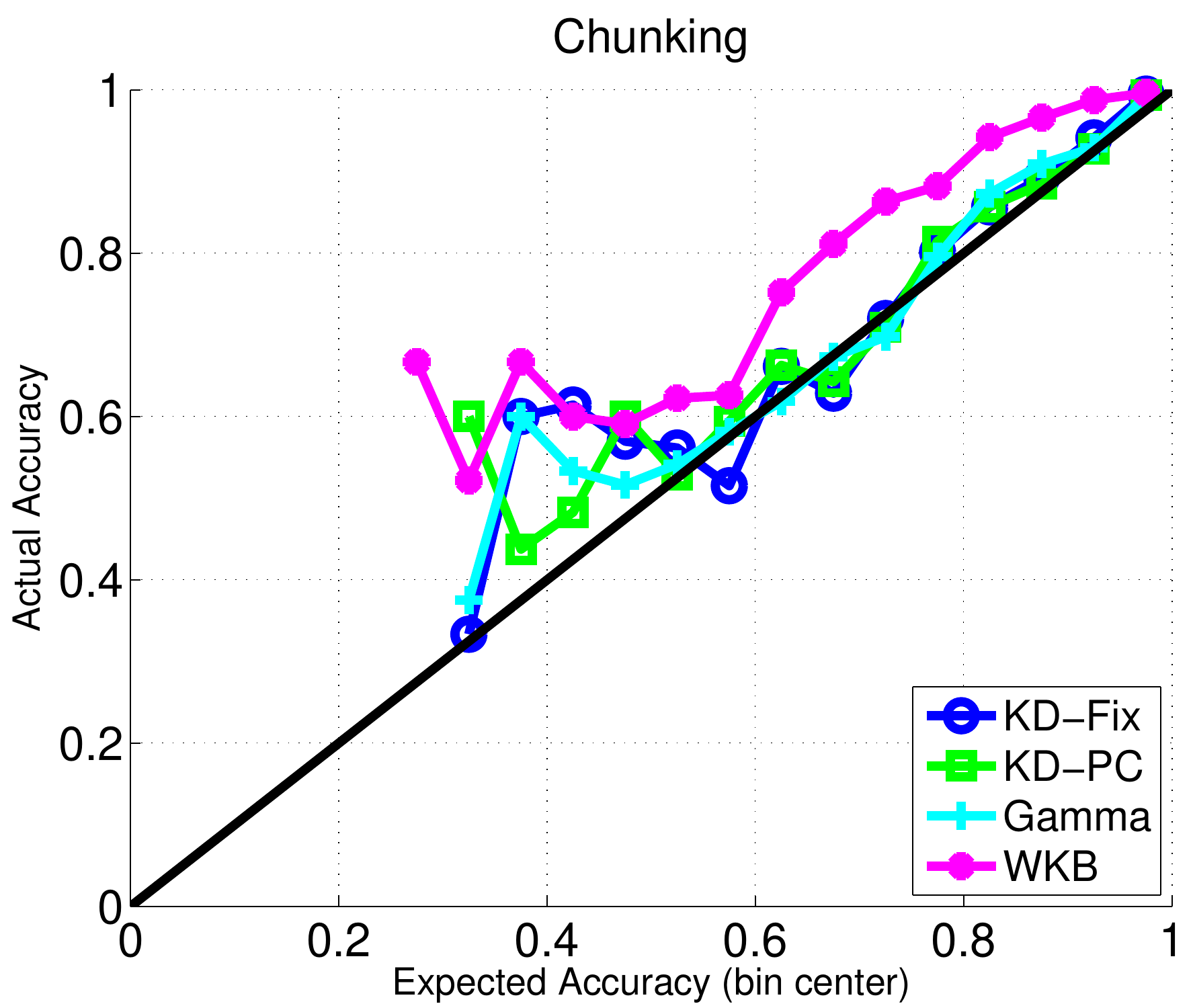}}
\end{tabular}
\caption{Predicted accuracy vs. actual accuracy in each bin. Best
  performance is obtained by methods close to the line $y=x$ (black
  line) for four tasks sequential labeling tasks. Four methods are
  compared: weighted $K$-Viterbi ({\tt WKB}), $K$-draws PC ({\tt
    KD-PC}) and $K$-draws fixed covariance ({\tt {\tt KD-Fix}}) and
  {\tt Gamma}.}
\label{fig:sequences_confidence_bins_CW}
\end{centering}
\end{figure}

\begin{figure}[!t!]
\begin{centering}
\begin{tabular}{cc}
\subfigure[NER English]{\includegraphics[width=0.48\textwidth]{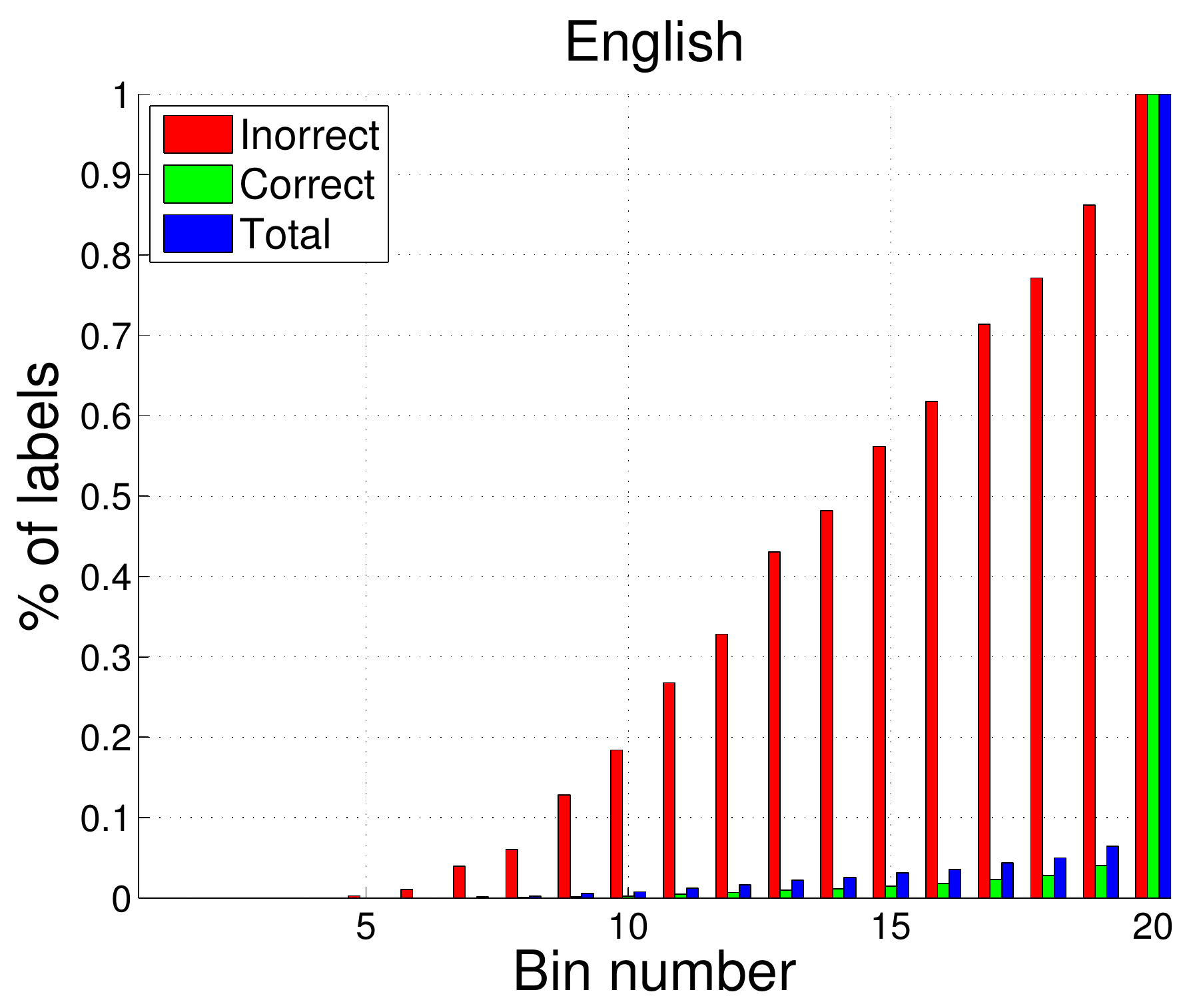}}&
\subfigure[NER Dutch]{\includegraphics[width=0.48\textwidth]{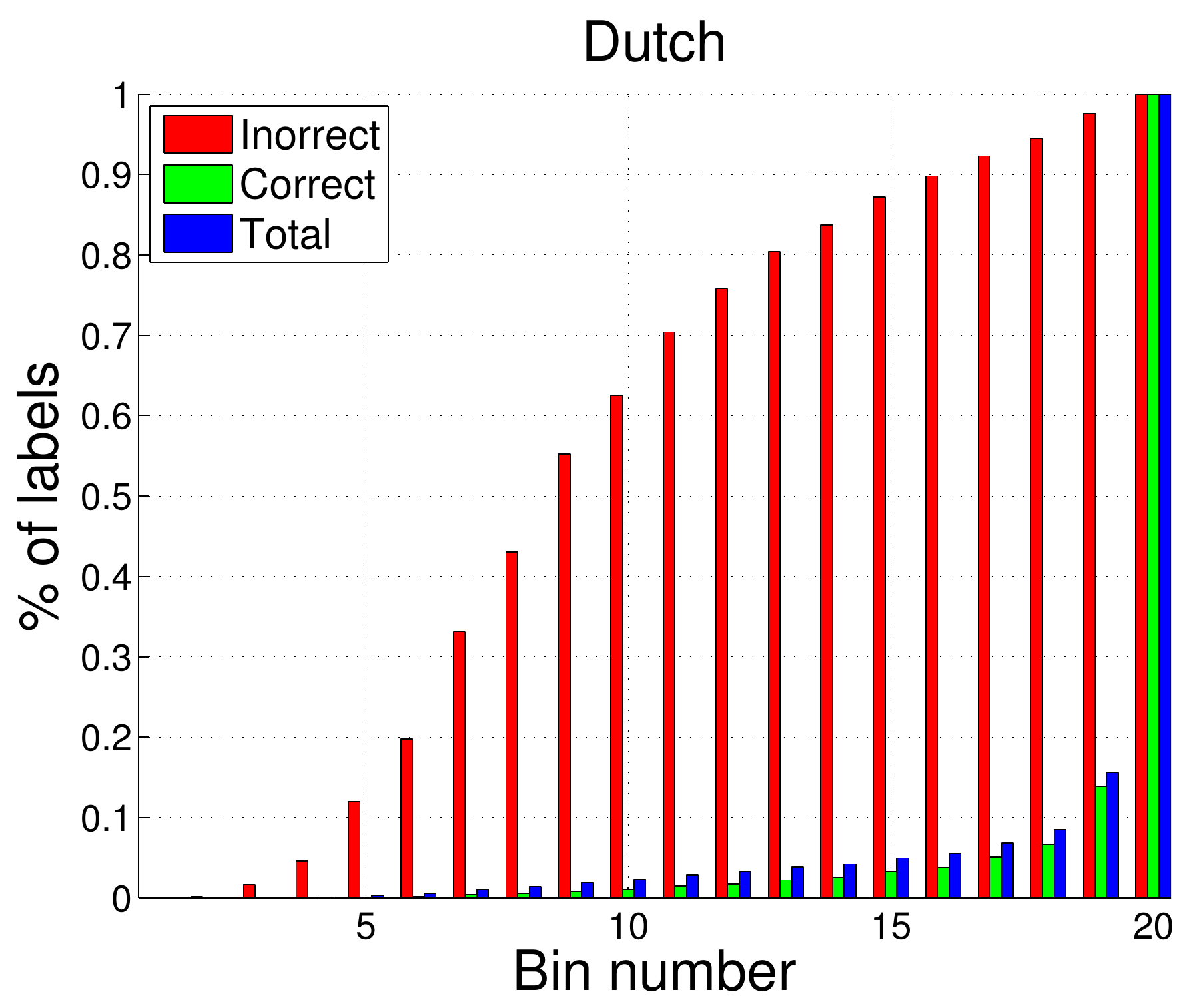}}\\
\subfigure[NER Spanish]{\includegraphics[width=0.48\textwidth]{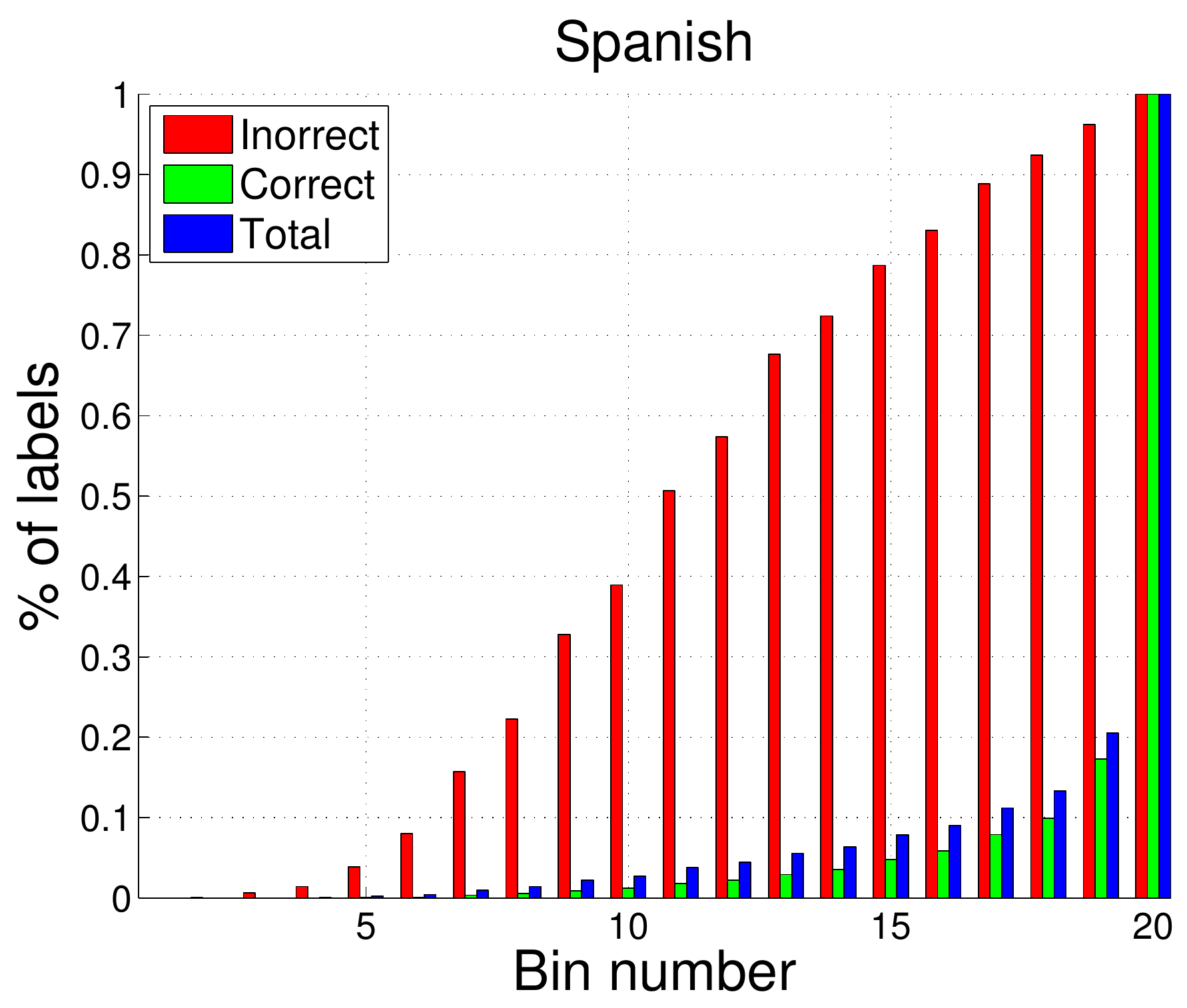}}&
\subfigure[NP Chunking]{\includegraphics[width=0.48\textwidth]{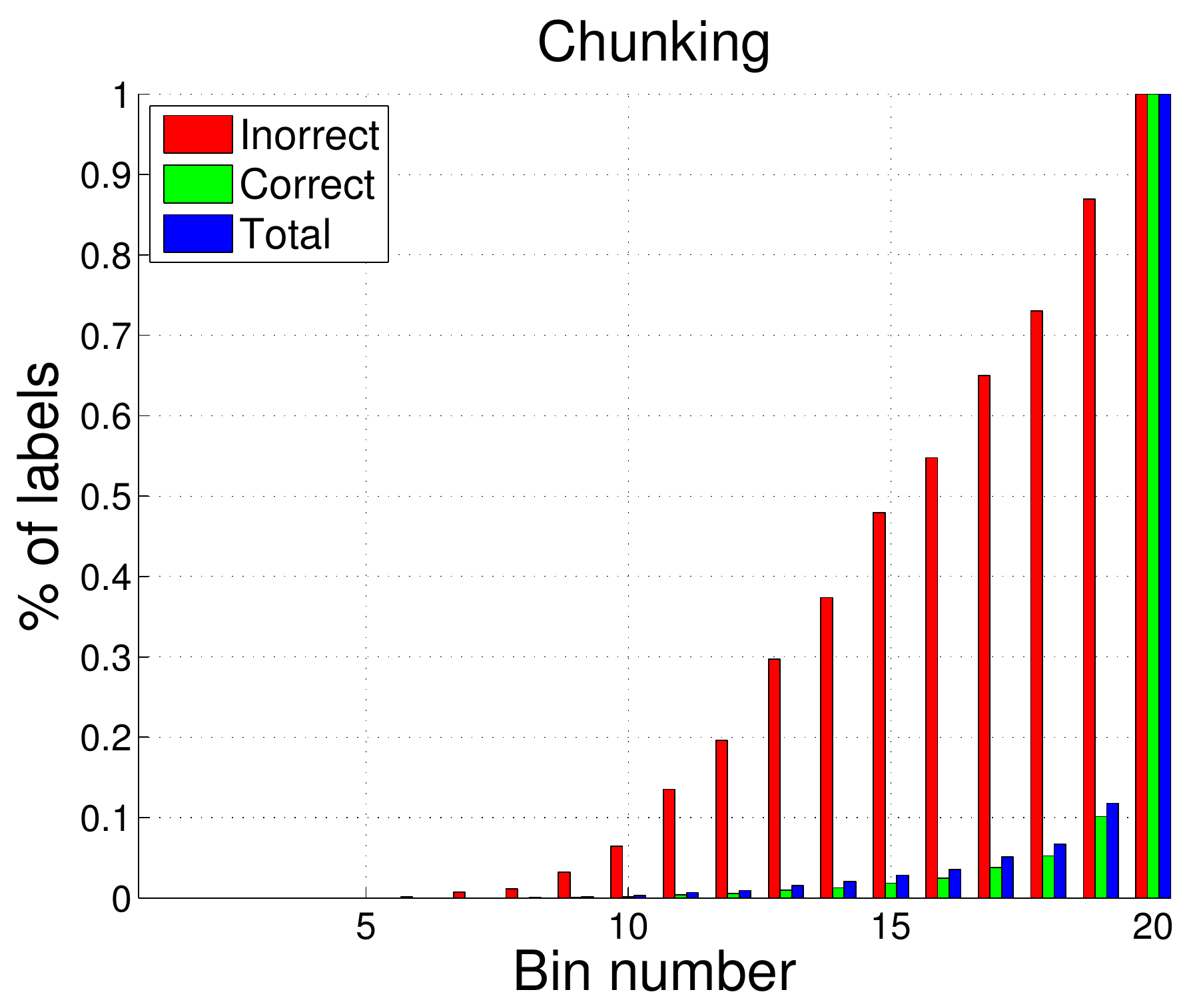}}
\end{tabular}
\caption{The cumulative distribution of the labels in the bins
  according to confidence scores assigned by {\tt KD-Fix} method on
  the four sequence labeling datasets. Blue bars represent the
  distribution of all the label in the bins as percentage of total
  number of labels, green and red bars represent the distribution of
  correct/incorrect labels as percentage of the
  \emph{correct/incorrect} labels.}
\label{fig:sequences_confidence_bins_CW_population}
\end{centering}
\end{figure}

\begin{figure}[!t!]
\begin{centering}
\subfigure[Chinese]{\includegraphics[width=0.32\textwidth]{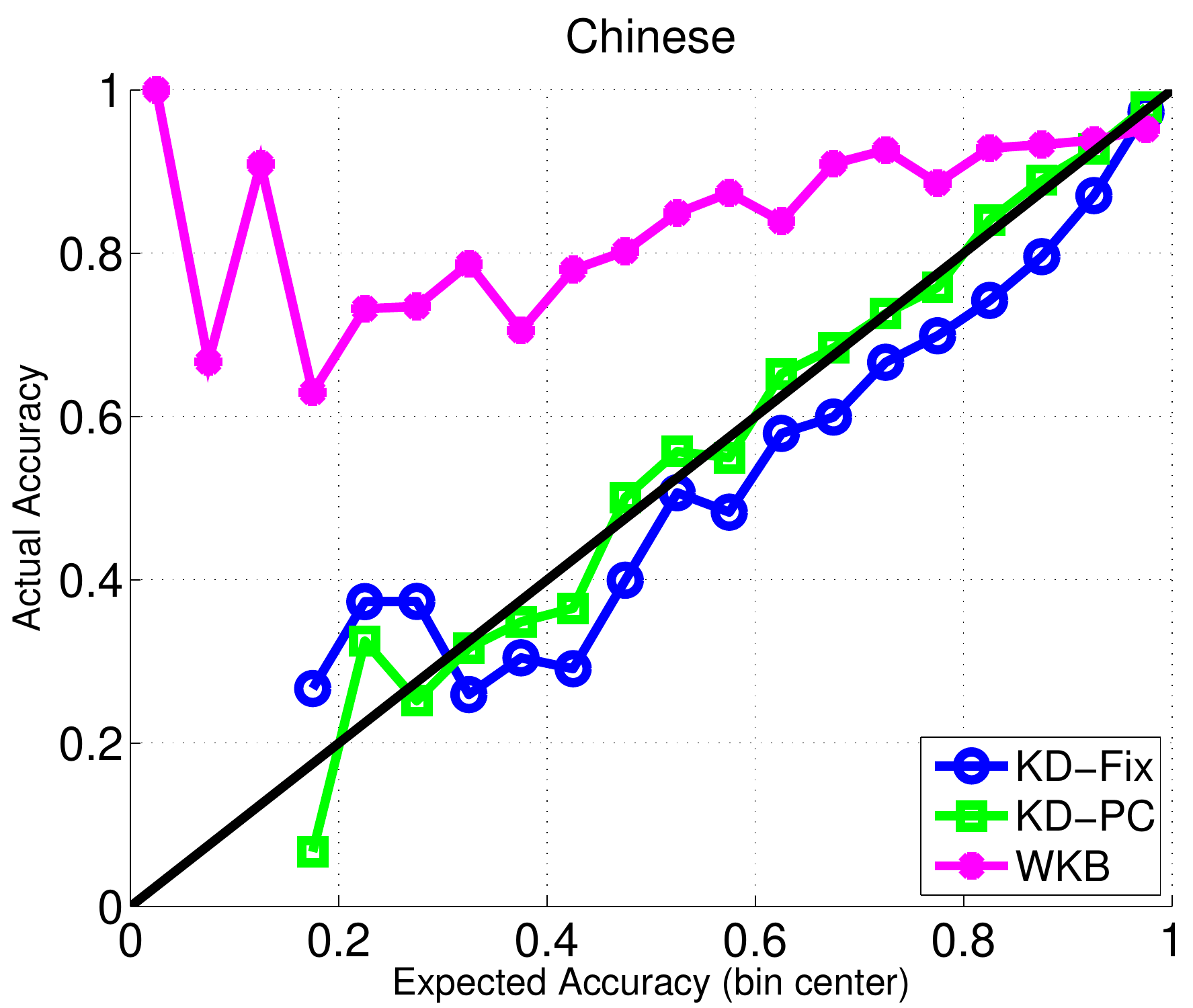}}
\subfigure[Swedish]{\includegraphics[width=0.32\textwidth]{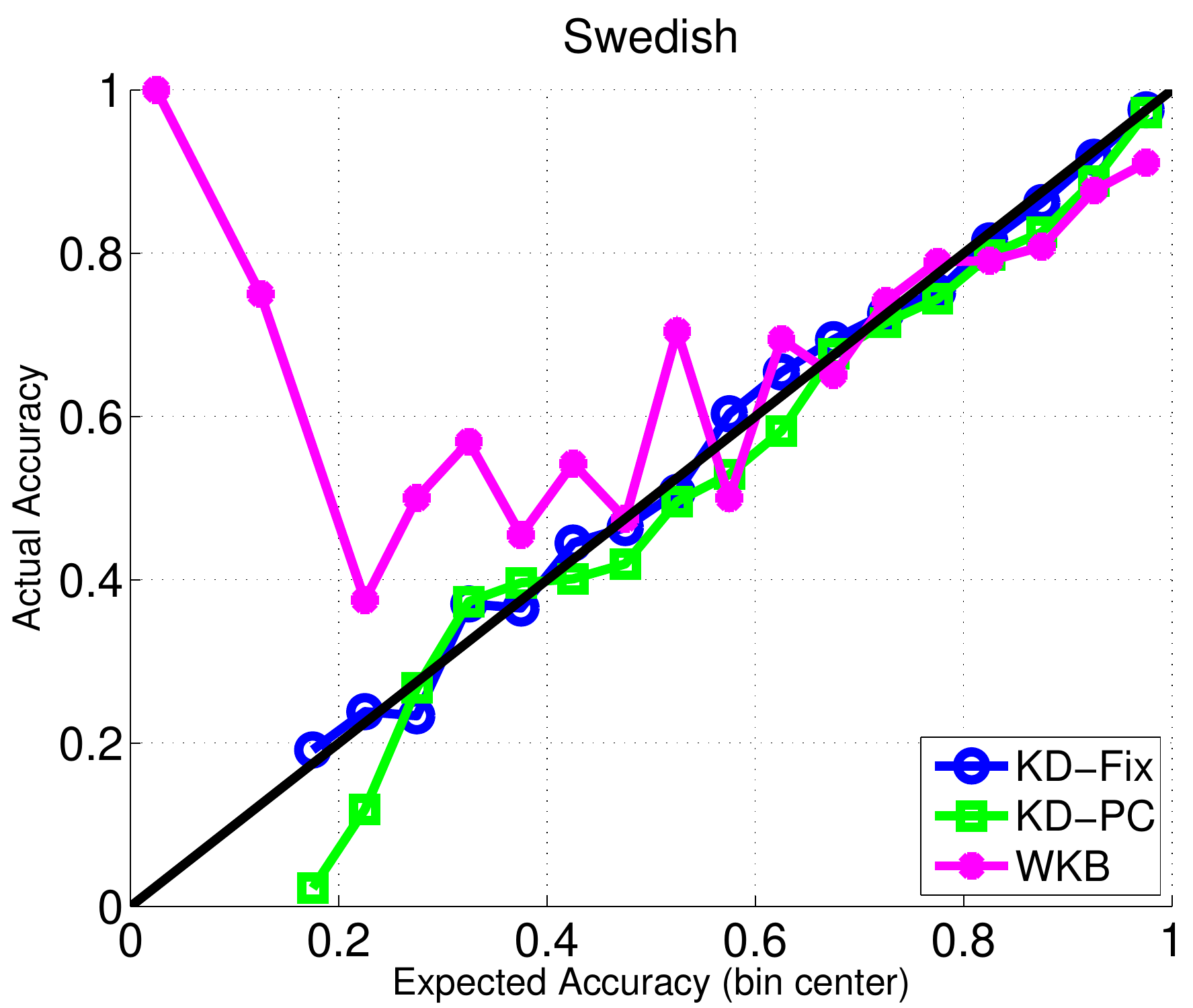}}
\subfigure[Turkish]{\includegraphics[width=0.32\textwidth]{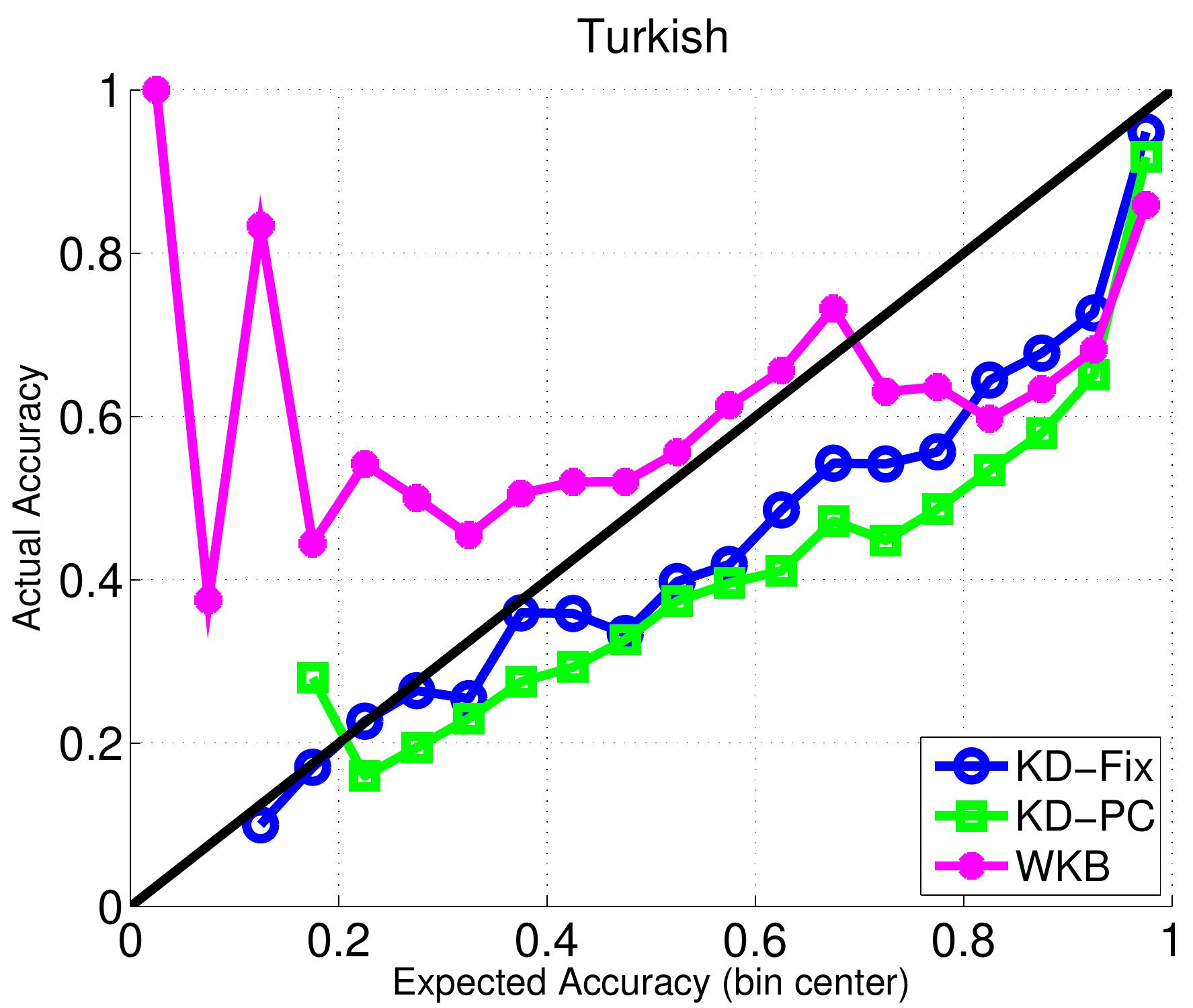}}
\subfigure[Danish]{\includegraphics[width=0.32\textwidth]{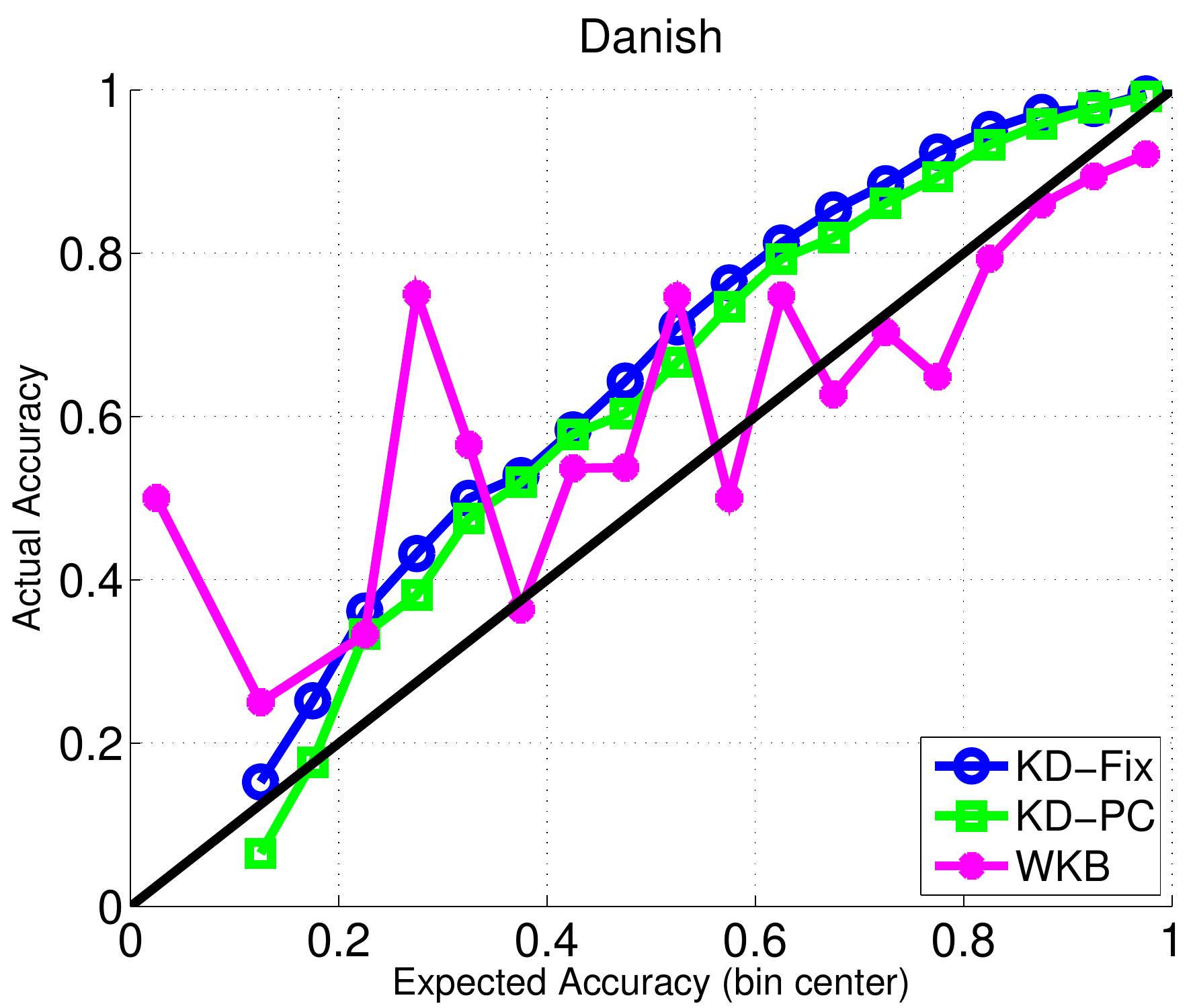}}
\subfigure[Arabic]{\includegraphics[width=0.32\textwidth]{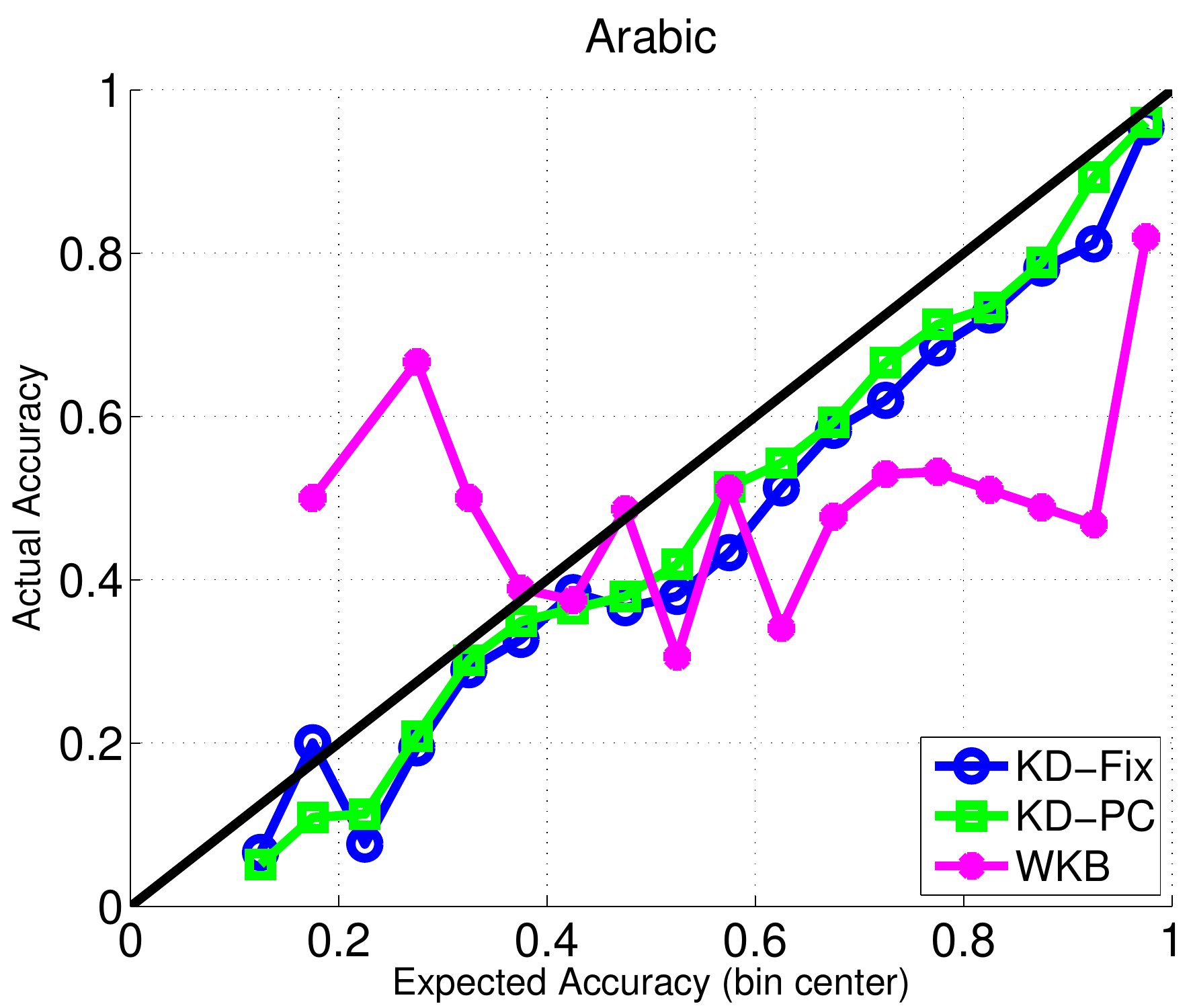}}
\subfigure[Bulgarian]{\includegraphics[width=0.32\textwidth]{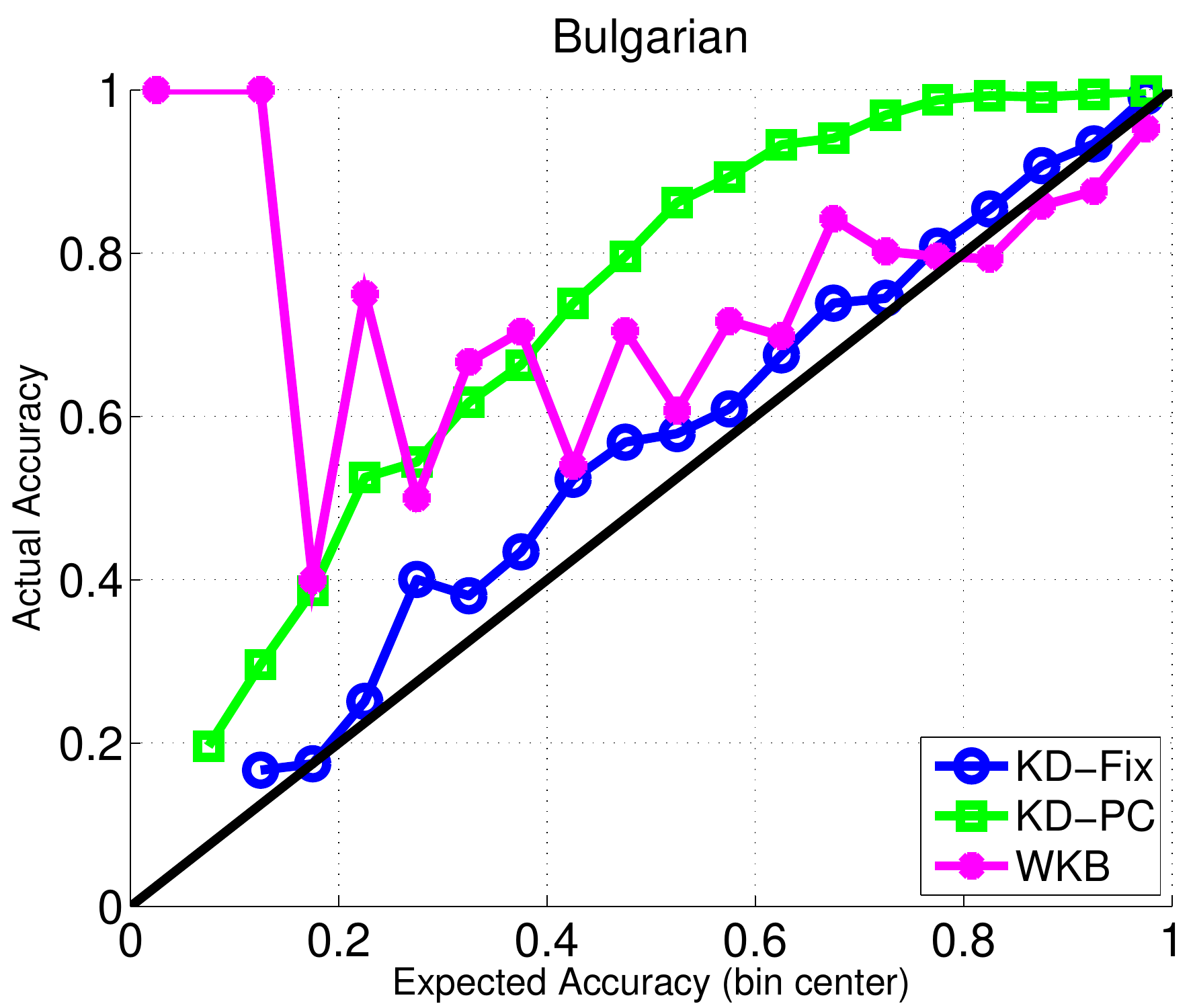}}
\subfigure[Czech]{\includegraphics[width=0.32\textwidth]{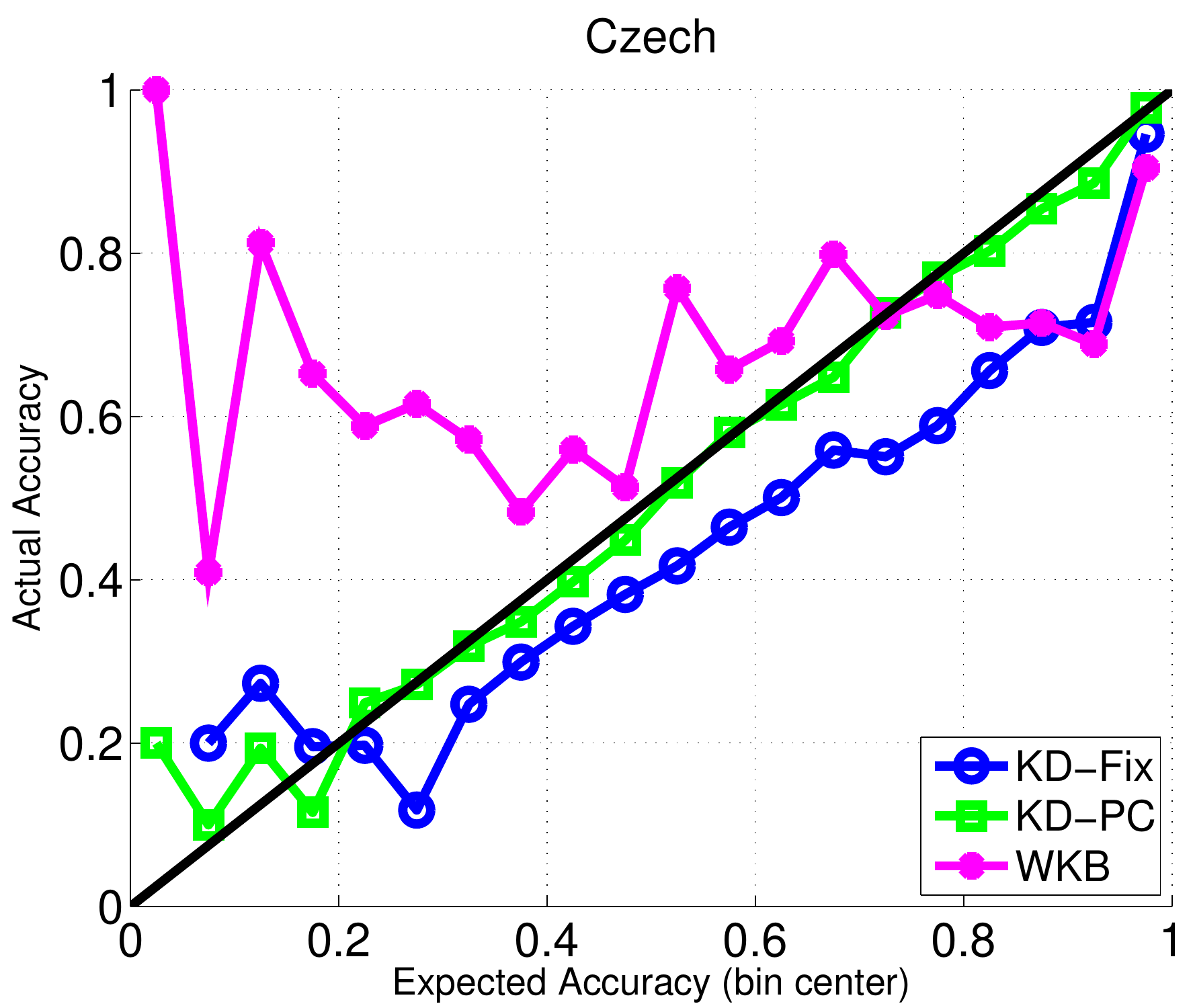}}
\subfigure[Dutch]{\includegraphics[width=0.32\textwidth]{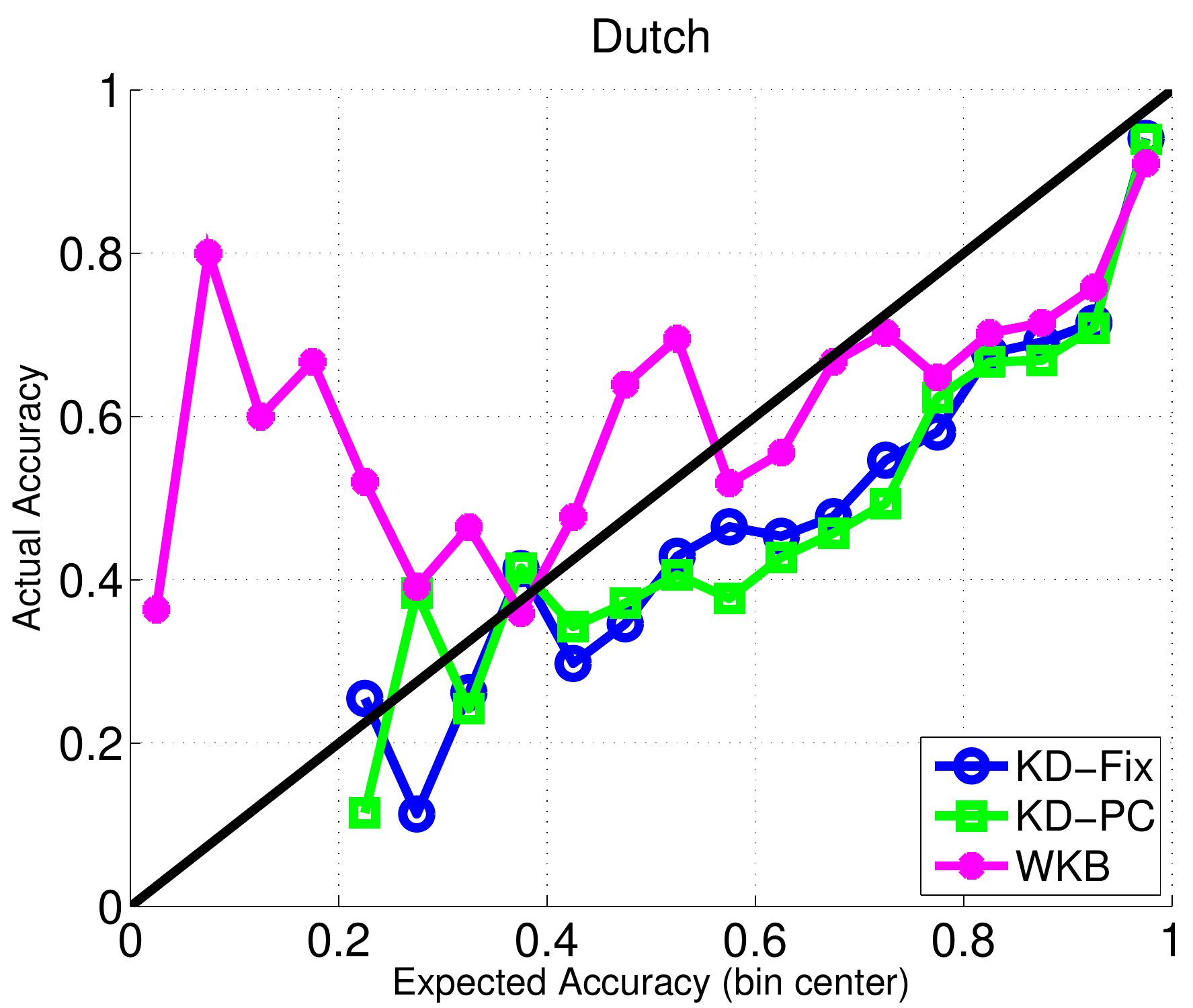}}
\subfigure[English]{\includegraphics[width=0.32\textwidth]{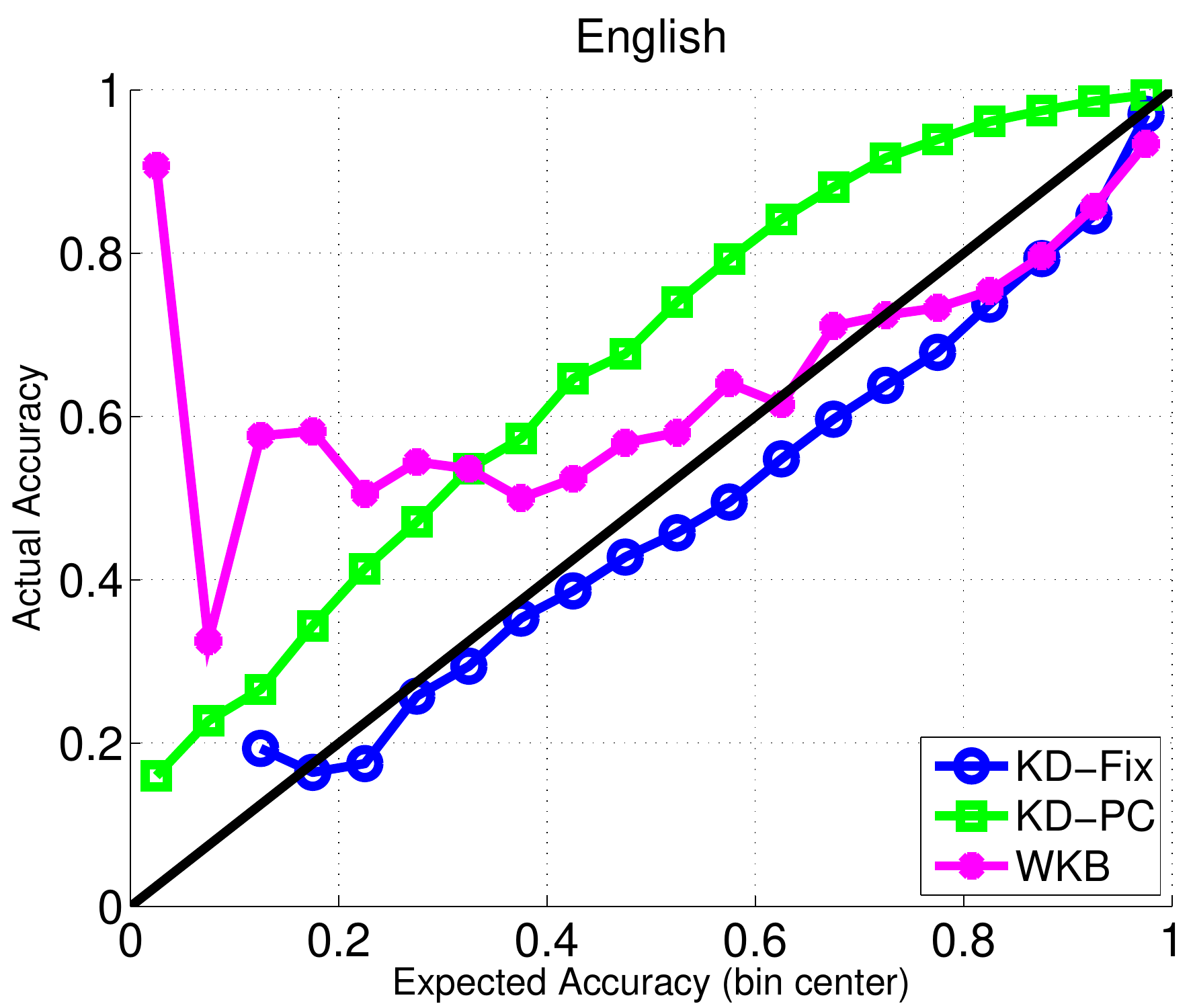}}
 \caption{{Evaluation of {\tt KD-Fix}, {\tt KD-PC} and {\tt WK-Best}
     by comparing predicted accuracy vs. actual accuracy in each bin
     on several of the dependency parsing datasets. Best performance is
     obtained for curves close to the line {\em y=x} (black line).}}
\label{fig:parsing_confidence_bins_CW}
\end{centering}
\end{figure}

\begin{table}[!t!]
\begin{center}
{
\begin{tabular}{|l|c|c|c|c|c|}
\hline
&{\tt KD-Fix} & {\tt KD-PC} & {\tt Gamma} & {\tt WKB} & Random \tabularnewline
\hline
\hline
NER English & 0.036	 & 0.021	 & 0.053	 & 0.108	 & 0.548 \tabularnewline
\hline
NER Dutch & 0.049	 & 0.024	 & 0.046	 & 0.104	 & 0.559 \tabularnewline
\hline
NER Spanish & 0.034	 & 0.030	 & 0.049	 & 0.046	 & 0.543 \tabularnewline
\hline
NP Chunking & 0.022	 & 0.020	 & 0.019	 & 0.056	 & 0.557 \tabularnewline
\hline
\hline
Average & 0.035	 & \textbf{0.024}	 & 0.042	 & 0.078	 & 0.552 \tabularnewline
\hline
\end{tabular}
\label{table:seq_RMSE_CW}
\caption{Root mean square error (RMSE) of the absolute confidence
  value by the confidence estimation methods for model trained with CW
  for sequence labeling tasks.}
}
\end{center}
\end{table}

\begin{table}[!t!]
\begin{center}
{
\begin{tabular}{|l|c|c|c|c|}
\hline
&{\tt KD-Fix} & {\tt KD-PC} & {\tt WKB} & Random \tabularnewline
\hline
\hline
Arabic	& 0.076	& 0.057	& 0.175	& 0.402 \tabularnewline
\hline
Bulgarian	& 0.037	& 0.223	& 0.093	& 0.492 \tabularnewline
\hline
Chinese	& 0.044	& 0.024	& 0.223	& 0.497 \tabularnewline
\hline
Czech	& 0.087	& 0.024	& 0.109	& 0.451 \tabularnewline
\hline
Danish	& 0.119	& 0.095	& 0.068	& 0.476 \tabularnewline
\hline
Dutch	& 0.098	& 0.105	& 0.094	& 0.441 \tabularnewline
\hline
English	& 0.041	& 0.144	& 0.059	& 0.483 \tabularnewline
\hline
German	& 0.090	& 0.094	& 0.102	& 0.481 \tabularnewline
\hline
Japanese	& 0.064	& 0.039	& 0.122	& 0.524 \tabularnewline
\hline
Portuguese	& 0.099	& 0.107	& 0.085	& 0.465 \tabularnewline
\hline
Slovene	& 0.156	& 0.137	& 0.151	& 0.404 \tabularnewline
\hline
Spanish	& 0.083	& 0.028	& 0.123	& 0.429 \tabularnewline
\hline
Swedish	& 0.023	& 0.031	& 0.114	& 0.465 \tabularnewline
\hline
Turkish	& 0.105	& 0.142	& 0.127	& 0.404 \tabularnewline
\hline
\hline
Average	& \textbf{0.080}	& 0.089	& 0.117	& 0.458 \tabularnewline
\hline
\end{tabular}
\label{table:parsing_RMSE_CW}
\caption{Root mean square error (RMSE) of the absolute confidence
  value by the confidence estimation methods for model trained with CW
  for dependency parsing.}
}
\end{center}
\end{table}

A second aspect of confidence prediction is the individual confidence
values outputted by the various methods, rather than only comparing
pairs of values.  As before, suitable confidence estimation methods
were applied on the entire set of predicted labels
\footnote{The output of the {\tt Delta} method can not be interpreted on
  their own, only relatively. In the context of binary classification
  there are few methods to generate absolute confidence from relative
  one~\cite{Platt99} using a sigmoid function. Yet, it could not be
  tuned properly in our setting since we tuned all methods according
  to their performance in {\em relative} confidence scenario
  (average-precision for ranking), where all parameters of Platt's
  method perform equally good.
}.
For every dataset and every
algorithm we grouped the words according to the value of their
confidence. Specifically, we used twenty (20) bins dividing uniformly
the confidence range into intervals of size $0.05$.
For each bin, we computed the fraction of words predicted correctly
from the words assigned to that bin. Ultimately, the value of the
computed frequency should be about the center value of the interval of
the bin. Formally, bin indexed $j$ contains words with confidence
value in the range $[(j-1)/20,j/20)$ for $j=1 \dots 20$. Let $b_j$ be
the center value of bin $j$, that is $b_j = j/20 - 1/40$. The
frequency of correct words in bin $j$, denoted by $c_j$ is the
fraction of words with confidence
\(
\nu\in[(j-1)/20,j/20)
\)
that their assigned label is correct. Ultimately, these two values
should be the same, $b_j=c_j$, meaning that the confidence information
is a good estimator of the frequency of correct predictions. Methods
for which $c_j > b_j$ are too pessimistic, predicting too high
frequency of erroneous predictions, while methods for which $c_j <
b_j$ are too optimistic, predicting too low frequency of erroneous
words.

The results for sequence labeling are summarized in
\figref{fig:sequences_confidence_bins_CW} and for dependency parsing in \figref{fig:parsing_confidence_bins_CW}. Each panel in each figure summarizes the results for a single
task (or language in parsing), where the value of the center-of-bin
$b_j$ is plotted vs. the frequency of correct prediction $c_j$,
connecting the points associated with a single algorithm. Best
performance is obtained when the resulting line is close to the line
$y=x$.  Four algorithms are shown: {\tt KD-Fix}, {\tt KD-PC}, {\tt
  Gamma} and {\tt WK-Best}. The results of the {\tt K-Best} method were
inferior to all other methods and omitted for clarity.

For sequence labeling we observe from the plots that {\tt WKB} is too pessimistic as its
corresponding line is above the line $y=x$. {\tt Gamma} method
tracks the line $x=y$ pretty closely in NP-Chunking and NER-Spanish datasets but it is too optimistic in the other two with its corresponding line is below the line $y=x$.
The {\tt {\tt KD-Fix}} method is too pessimistic on NER-Dutch and too optimistic on
NER-English. The best method is {\tt KD-PC} which tracks the line $x=y$ pretty closely in all four datasets.

For dependency parsing, in most languages {\tt KD-Fix} and {\tt KD-PC}
methods perform very similarly. The distribution of this qualitative
behavior of the KD methods among the $14$ datasets is: too optimistic
in $2$ datasets, too pessimistic in $7$ 
and close to the line $y=x$ in $5$ datasets. The confidence scores
produced by the {\tt WKB} method are in general worse than {\tt
  KD-Fix} and are too pessimistic with the line above $y=x$. In quite
a few datasets we observe that {\tt WKB} is too optimistic in some
confidence range and too pessimistic in another range. In most plots
both in \figref{fig:sequences_confidence_bins_CW} and in
\figref{fig:parsing_confidence_bins_CW} the curves on the left (low
bin-values) are far from the line $y=x$ and with more fluctuation
compared with the right area curves. This is because the left (low confidence) bins are
less populated and thus the estimates are noisier.

\figref{fig:sequences_confidence_bins_CW_population} shows the
distribution of the words in the bins according to confidence scores
assigned by {\tt KD-Fix} method on the four sequence labeling
datasets. Blue bars represent the distribution of all the labels in
the bins as percentage of total number of labels, green and red bars
represent the distribution of correct/incorrect labels as percentage
of the \emph{correct/incorrect} labels. We see that $\sim\!\!80-90\%$
of the labels populate the highest confidence bin, another
$\sim\!\!5\%$ populate the second highest bin and the rest of the bins
are lightly populated. Among the correct labels even higher percentage
is concentrated at the highest confidence bin, while the incorrect
labels are distributed more evenly over many of the lower confidence
bins.

\begin{table}[!t!]
  \begin{center}
    {
      \begin{tabular}{|l|c|c|c|c|c|}
        \hline
        &{\tt KD-Fix} & {\tt KD-PC} & {\tt Gamma} & {\tt WKB} & Random \tabularnewline
        \hline
        \hline
        Sequences CW & 0.035 & \textbf{0.024} & 0.042 & 0.078 & 0.552 \tabularnewline
        \hline
        Sequences PA & \textbf{0.031} & - & 0.039 & 0.080 & 0.543 \tabularnewline
        \hline
        \hline
        Parsing CW & \textbf{0.080} & 0.089 & - & 0.117 & 0.458 \tabularnewline
        \hline
        Parsing PA & \textbf{0.068} & - & - & 0.122 & 0.456 \tabularnewline
        \hline
      \end{tabular}
     \caption{Average root mean square error (RMSE) of the absolute confidence value by the confidence estimation methods for models trained with CW and with PA. For sequences the results are averaged over all four datasets and for parsing over all $14$ languages.}
}
      \label{table:seq_parsing_RMSE_CW_PA}
\end{center}
\end{table}

The predicted vs. actual accuracy plots do not reflect the fact that different
bins are not populated uniformly . We thus compute for each
method the root mean-square error (RMSE) in predicting the bin center
value given by
 \[
 RMSE = \sqrt{\frac{\sum_j n_j (b_j-c_j)^2}{\sum_j n_j}} ~,
\]
where $n_j$ is the number of words in the j$th$ bin.  The computed
RMSE values are presented
in 
Table~11 
both sequence labeling and
parsing averaged over datasets.
We observed a similar trend to the one
appeared in the bins plots.  For sequences, when using CW for
training 
(top row in Table~11) 
, {\tt WKB} is the
least-performing method (after Random), then {\tt Gamma} and {\tt
  KD-Fix}, where each was better than the other in two datasets, and
on average {\tt KD-Fix} achieved lower RMSE. {\tt KD-PC} performed
best where it achieved lowest RMSE in three of four datasets and on
average.  Similar results, obtained when training with PA, appear in
the second row of the table. {\tt KD-Fix} achieved lowest RMSE in
three of four datasets and on average, and {\tt WKB} performs worst.

For parsing Table~11 
 presents the RMSE results
averaged over all $14$ languages, for parser trained with CW (third
row) and with PA (fourth row). For CW, both K-Draws methods perform
better than {\tt WK-Best}. {\tt KD-Fix} yield lower RMSE than {\tt
  KD-PC} in seven of the languages and higher in the other seven, and
on average {\tt KD-Fix} performed better than {\tt KD-PC}
(Table~10 
presents results for all
languages). This is in oppose to the results observed for sequences
where {\tt KD-PC} performed better than {\tt KD-Fix} in all four
datasets. When using PA for training, we also see that {\tt KD-Fix}
performed better than {\tt WK-Best} method.

To conclude, both {\tt KD-Fix} and {\tt KD-PC} outperform all other
methods in both settings and most datasets. The former is slightly
better than the later, except in absolute evaluation for sequence
labeling.

\subsection{Effect of $K$ value on K-Draws methods performance}
\label{sec:effect_k}
\begin{figure}[!t!]
\begin{centering}
\begin{tabular}{cc}
\subfigure{\includegraphics[width=0.48\textwidth]{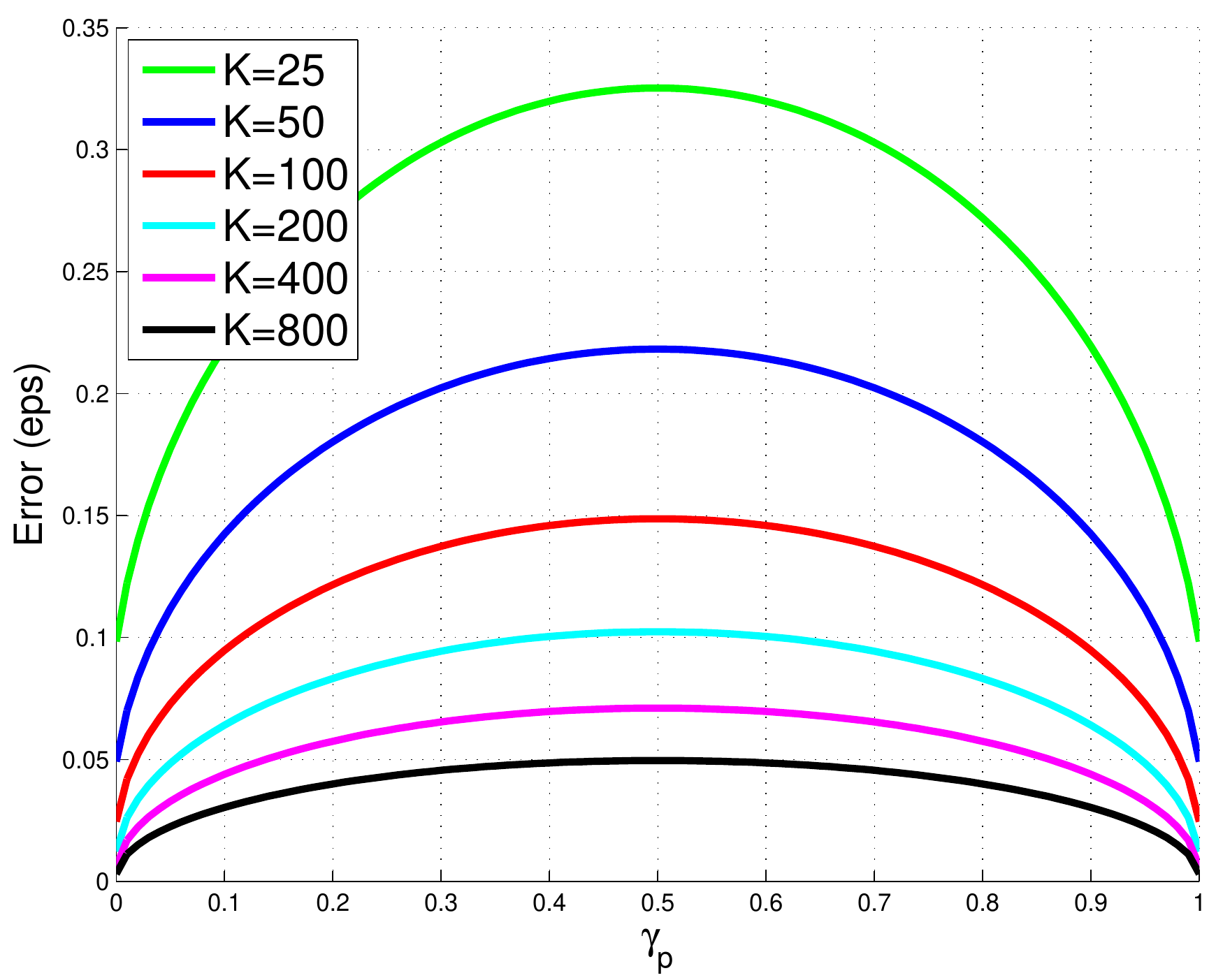}}&
\subfigure{\includegraphics[width=0.48\textwidth]{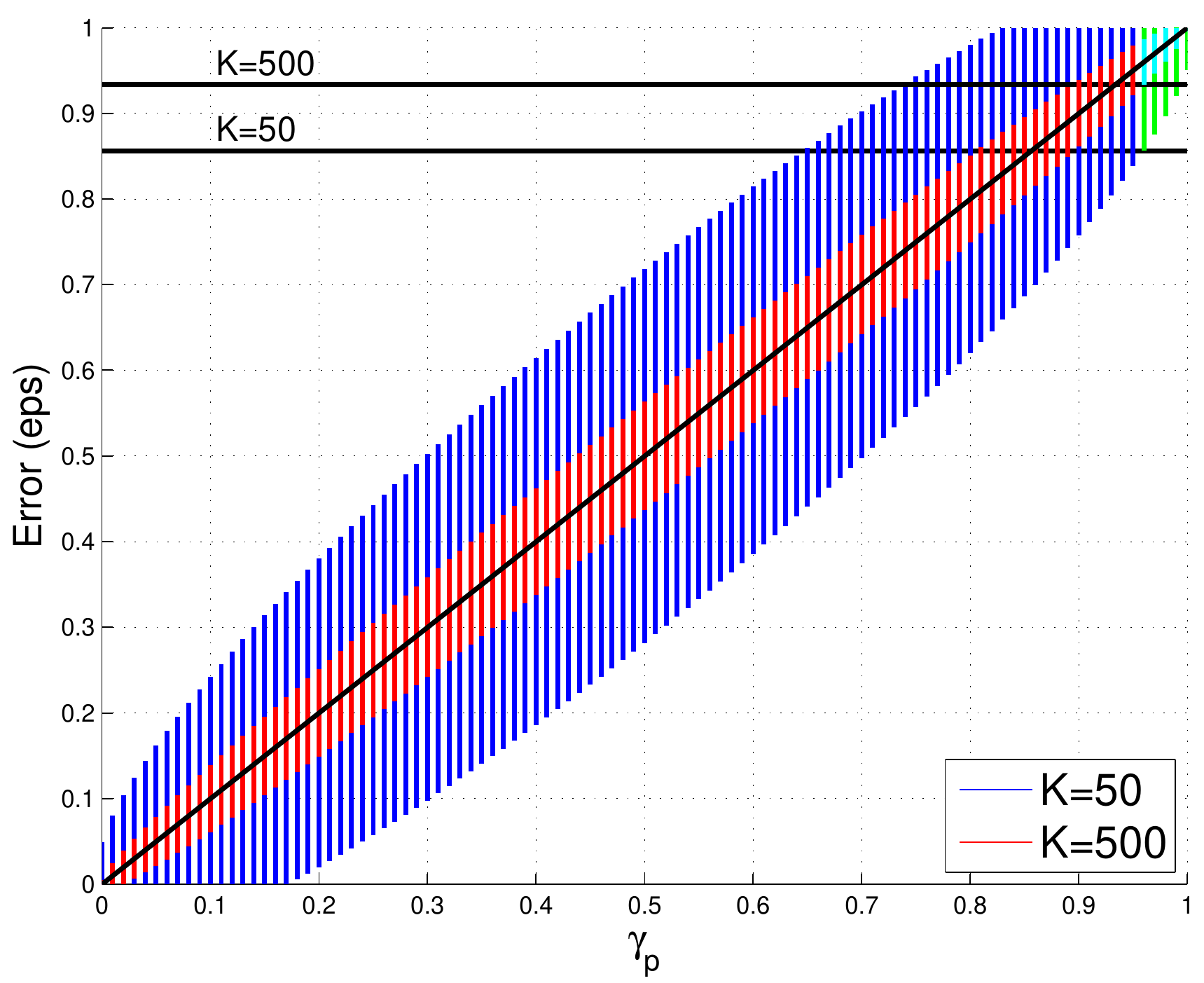}}
\end{tabular}
\caption{Left: The absolute error bounds for $\gamma_p$ estimation
  using {\tt KD-Draws} method with different number of samples $K$ for
  $\gamma_p \in [0,1]$ at confidence level $95\%$. Right: The interval
  of possible values for $\nu_p$ using $K=50$ and $K=500$ at confidence level
  $95\%$.}
\label{fig:error_bound_by_K}
\end{centering}
\end{figure}

\begin{figure}[!t!]
\begin{centering}
\begin{tabular}{cc}
\subfigure[Sequence
Labeling]{\includegraphics[width=0.48\textwidth]{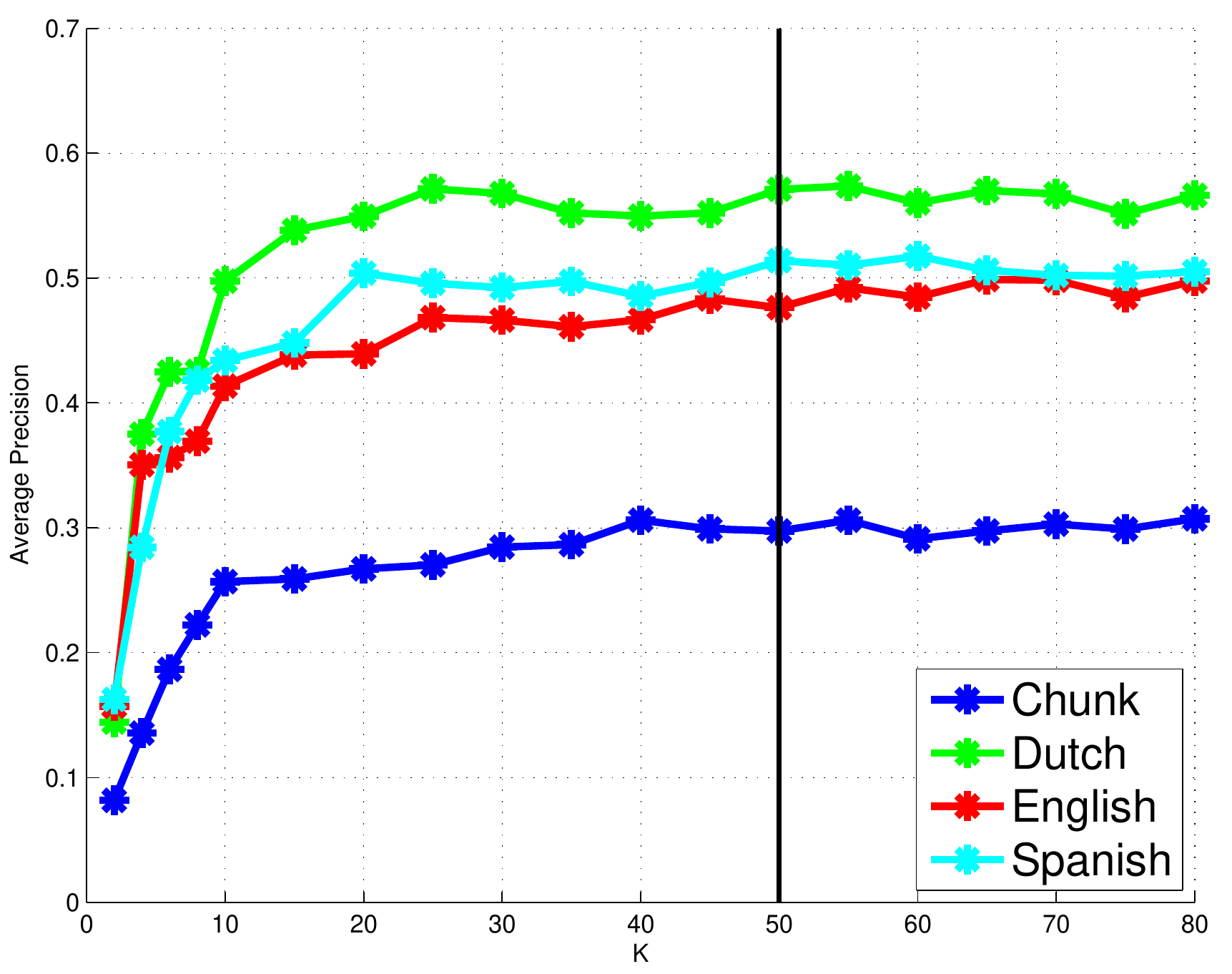}}&
\subfigure[Dependency
Parsing]{\includegraphics[width=0.48\textwidth]{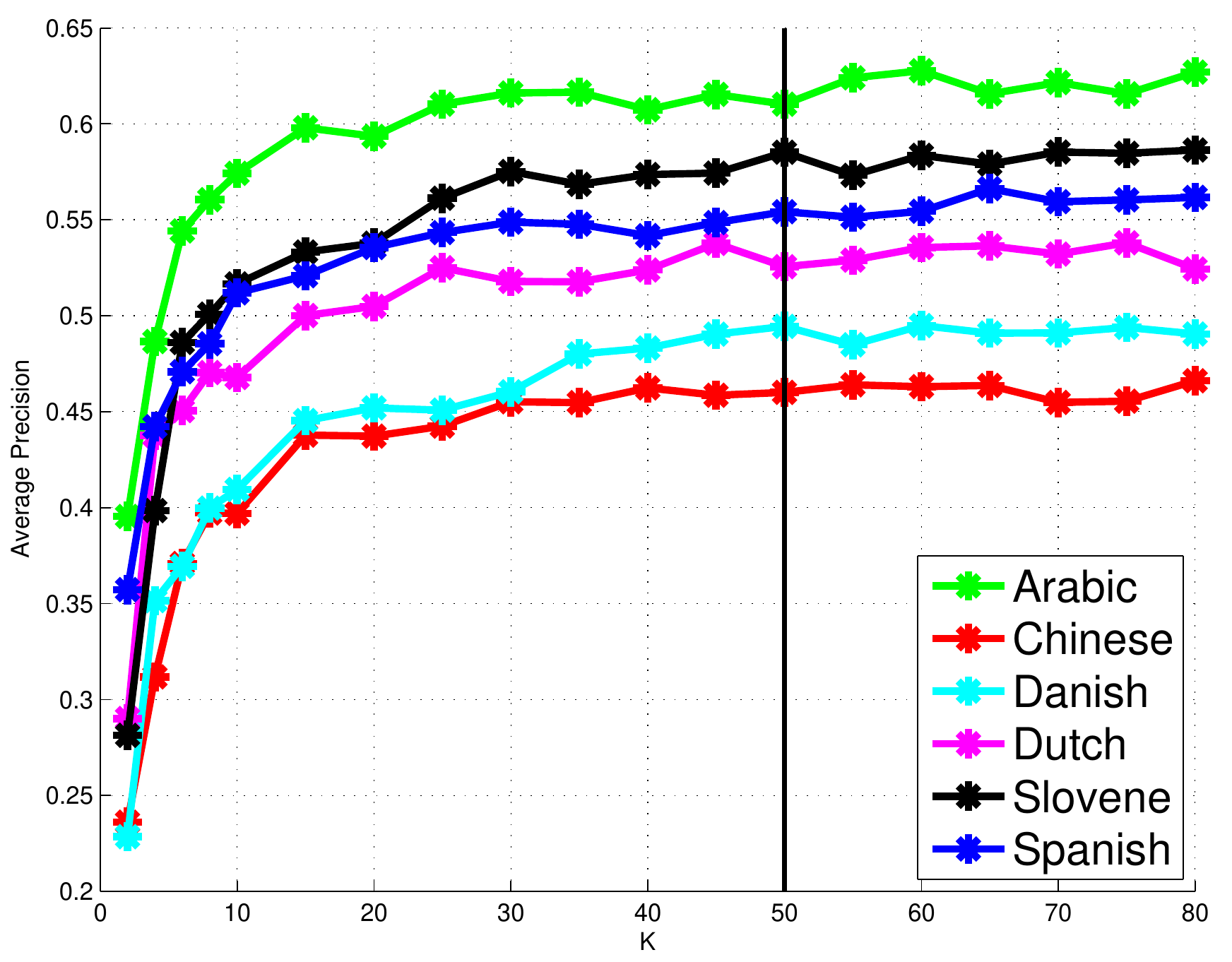}}
\end{tabular}
\caption{Average precision in ranking all predictions according to
  confidence scores assigned by {\tt KD-Fix} method with K value in
  the range of 2 to 80 for sequence labeling (left) and dependency
  parsing (right). The black vertical line indicates the value of
  $K=50$ used in our experiments.}
\label{fig:seq_parsing_multi_K_KDFix_AvgPrec}
\end{centering}
\end{figure}

One of the two parameters that need to be tuned for the {\tt K-Draws}
methods is $K$,
the number of alternative labeling. As discussed above, for each word
in each sentence these methods eventually output a single value
between zero and one, interpreted as the probability of the prediction
being correct. Conceptually, for every word $p$ we associate a coin
with probability $\gamma_p$. Our goal is to estimate these
probabilities up to a satisfying level, denoted by $\eps$. These
methods are used to estimate the coin's bias by sampling from it, as
defined in \eqref{k_draw_eq}, \( \nu_p = {\left\vert\left\{i~:~
      \hat{y}_p = z^{(i)}_p \right\}\right\vert}/{K} ~.  \) Applying
the inequality of ~\namecite{Chernoff52} and of ~\namecite{Hoeffding-pisbrv-63} we have,
\begin{equation}
\pr{\vert \gamma_p - \nu_p \vert \geq \eps} \leq 2 \exp \paren{-2 K
  \eps^2} ~.
\label{chernoff}
\end{equation}
Denote by $N$ the total number of words in the test set, using
the union bound we have,
\[
\pr{\exists \textrm{ a word } p \textrm{ s.t. } \vert \gamma_p - \nu_p
  \vert \geq \eps} \leq \sum_p \pr{\vert \gamma_p - \nu_p \vert \geq
  \eps} \leq 2N \exp \paren{-2 K
  \eps^2} ~.
\]

Let $0<\delta<1$ be some confidence level. Then upper
bounding the right-hand-side of the last inequality with $\delta$ and
solving for $K$ we get,
\[
K \geq \frac{\log\paren{2N/\delta}}{2\eps^2} ~.
\]
Note that if we set $\eps$ to be the length of a confidence bin, then
by setting the value of $K$ to be above the lower bound, we bound the
probability that a word will not fall in the correct bin, or one of
its two closet neighbors.

Concretely, we used $20$ bins, thus we set
$\eps=1/20=0.05$. Additionally, we set $\delta=0.05$, and $N=500,000$
(see Table~2 and Table~3)
we get, $K\approx 3,363$.

This estimate is pessimistic in many aspects. First, it is based on
the assumption that the estimates $\nu_p$ of $\gamma_p$ are
independent, yet this is clearly not the case since features that tie
the prediction for few words are used. In the extreme case, where the
prediction of all words in a sentence are completely tied (that is,
knowing the labeling of one word, induces a deterministic labeling of
all other words), then the number of all words $N$ should be replaced
with the number of all sentences a typical value of about $25,000$,
which yields a value of $K \approx 2,764$.

The discussion above is pessimistic in another sense. It
assumes 
we need that all estimates $\nu_p$ will be close to their correct
values $\gamma_p$. However, we are often not interested in estimating
the bias values $\gamma_p$ per-se, but estimating confidence for a
practical purpose.  For example, in the task of incorrect prediction
detection, a good separation between the labels with high and low
$\gamma_p$ values is sufficient and error on the absolute value is
acceptable.  Therefore we can take the actual value of $\gamma_p$ into
consideration. Using the inequality of \namecite{Bernshtein46} (see
also \cite{Uspensky37}) we have,
\begin{equation}
\pr{\vert \gamma_p - \nu_p \vert \geq \eps} \leq 2 \exp \paren{-\frac{ K
  \eps^2}{2\paren{\gamma_p(1-\gamma_p) + \eps/3}}} ~.
\end{equation}
For $\gamma_p$ values close to $1$ the error is small which allows
effective detection of high confidence labels.  Comparing the
right-hand-side of the last equation to $\delta/N$ we solve for the
error interval and get a $\gamma-$dependent error interval
\[
\eps(\gamma_p) = \frac{\frac{2}{3}L+\sqrt{(\frac{2}{3}L)^2+8KL\gamma_p(1-\gamma_p)}}{2K}.
\]
where $L=\log{\frac{2N}{\delta}}$.
In other words, for each value $\gamma_p$ we get a different
error interval $\eps=\eps(\gamma_p)$ such that with probability
greater than $1-\delta$ all estimates $\nu_p$ falls within the
interval $[\gamma_p - \eps(\gamma_p) , \gamma_p + \eps(\gamma_p)]$.

The left plot of \figref{fig:error_bound_by_K} presents error
intervals for all $\gamma_p$ estimated using {\tt KD-Draws} method
with different number of samples $K$ with probability greater than
$95\%$. For values of $\gamma_p$ far from $0.5$ the intervals
$\eps(\gamma_p)$ are low, compared with values of $\gamma_p$ close to
$0.5$. The right plot presents for each value of $\gamma_p$ the
interval where its estimate $\nu_p$ may fall with high-probability
using $K=50$ (Blue) and $K=500$ (Red) for confidence level $95\%$.
The two horizontal lines mark (with high-probability) the lowest
possible estimate $\nu_p$ for $\gamma_p=0.95$, the lower and upper
lines for $K=50$ and $500$ respectively. We see that for $K=50$ words
with $\gamma_p>0.65$ may have their corresponding estimate $\nu_p$
greater than words with $\gamma_p = 0.95$, while when using $K=500$
only words with $\gamma_p>0.9$ may have the corresponding estimate
$\nu_p$ greater than labels with $\gamma_p = 0.95$.  This means, for
example, that if we are interested in words that their prediction is
with high-confident, then when using $K=500$ and picking all words
with $\nu_p>0.95$ with high probability we will pick only words with
$\gamma_p>0.9$.  As expected, greater values of $K$ yield better error
estimates so the choice of $K$ should be guided by the application of
the confidence scores and the sensitivity to the estimate accuracy.
Additionally, we observed in
\figref{fig:sequences_confidence_bins_CW_population} that most (more
than $\%90$) words have estimates $\nu_p>0.95$. So for most words the
estimates is pretty close to the correct values, which will not affect
much the average precision or f-measure.

Although theoretically about $K=500$ is expected to yield good results, still this value is large for reasonable purposes.
We evaluated the affect of $K$ for various values of $K$ and
empirically found that for our practical purposes relatively small
values of $K$ are sufficient and increasing $K$ beyond these values
does not improve performance.  Indeed, as mentioned above, for the
{\tt K-Draws} methods we set the value of $K$ by picking a best value
of $K$ on a {\em development set} evaluating average precision in the
task of incorrect prediction detection. Eventually we set a single
value of $K=50$ for all datasets.  Indeed, this value is one order of
magnitude smaller than the most optimistic estimate above. One
explanation is that the bounds are worst-case in the sense that we
assumed that the estimate $\nu_p$ will be smaller than $\gamma_p$ for
all words with high-confidence and larger than $\gamma_p$ for all
words with low-confidence. Yet, there is no reason that this will
happen in practice for all such words, only for a small fraction of
them, and thus smaller values of $K$ yield in practice accurate
estimates.

\figref{fig:seq_parsing_multi_K_KDFix_AvgPrec} presents average
precision results achieved on the test sets using the {\tt KD-Fix}
method with different values of $K$ from $2$ to $80$ for all sequence
labeling tasks and several dependency parsing tasks (the rest of the
languages follow the same trend and omitted from the plot for
clarity). We observe that even with $K=2$, only two samples per
sentence, the average precision results are better than random ranking
in all tasks. As $K$ is increased to $10$ the performance is greatly
improved. For $K=10$, despite the very large theoretical estimation
error, the results are better than the {\tt K-best} methods for all
tasks and for the NER tasks even better than {\tt Gamma} and {\tt
  Delta}. As $K$ is further increased the results continue to improve,
yet at a more moderate rate, until reaching maximal results at
$K\approx30$ for most tasks and results remain steady for larger
values of $K$ up to 80. These curves can be used to tradeoff
performance (average precision) with time, as the time complexity of
this method is linear in $K$.

\section{Applications}
Additionally of confidence estimation being useful by itself in some
contexts, it is also useful for solving other problems.  The next two
sections present two example applications built on top of the
algorithms presented so far.  The first application is using the
confidence information to trade-off precision and recall. In
\secref{sec:prec_recall_trade} we describe a modification of an NER
system, allowing it to label less words as named-entities, but ask
that the labeling will be with high precision. The second application,
presented in \secref{sec:active}, is performing active leaning in the
context of sequence labeling, using the confidence information as a
tool to choose which sentences should be labeled by the annotator.

\subsection{Precision-Recall Tradeoff}
\label{sec:prec_recall_trade}

\begin{figure}[!t!]
\begin{centering}
{\includegraphics[width=0.7\textwidth]{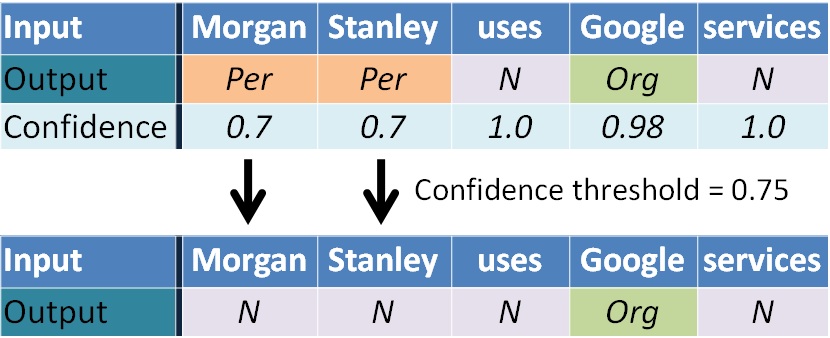}}
\caption{Example of NER system that discards tags with low confidence scores in order to improve precision possibly by sacrificing recall. }
\label{fig:precision_recall_tradeoff_example}
\end{centering}
\end{figure}

\begin{figure}[!h!] \begin{centering} \begin{tabular}{cc}
      {\includegraphics[width=0.43\textwidth]{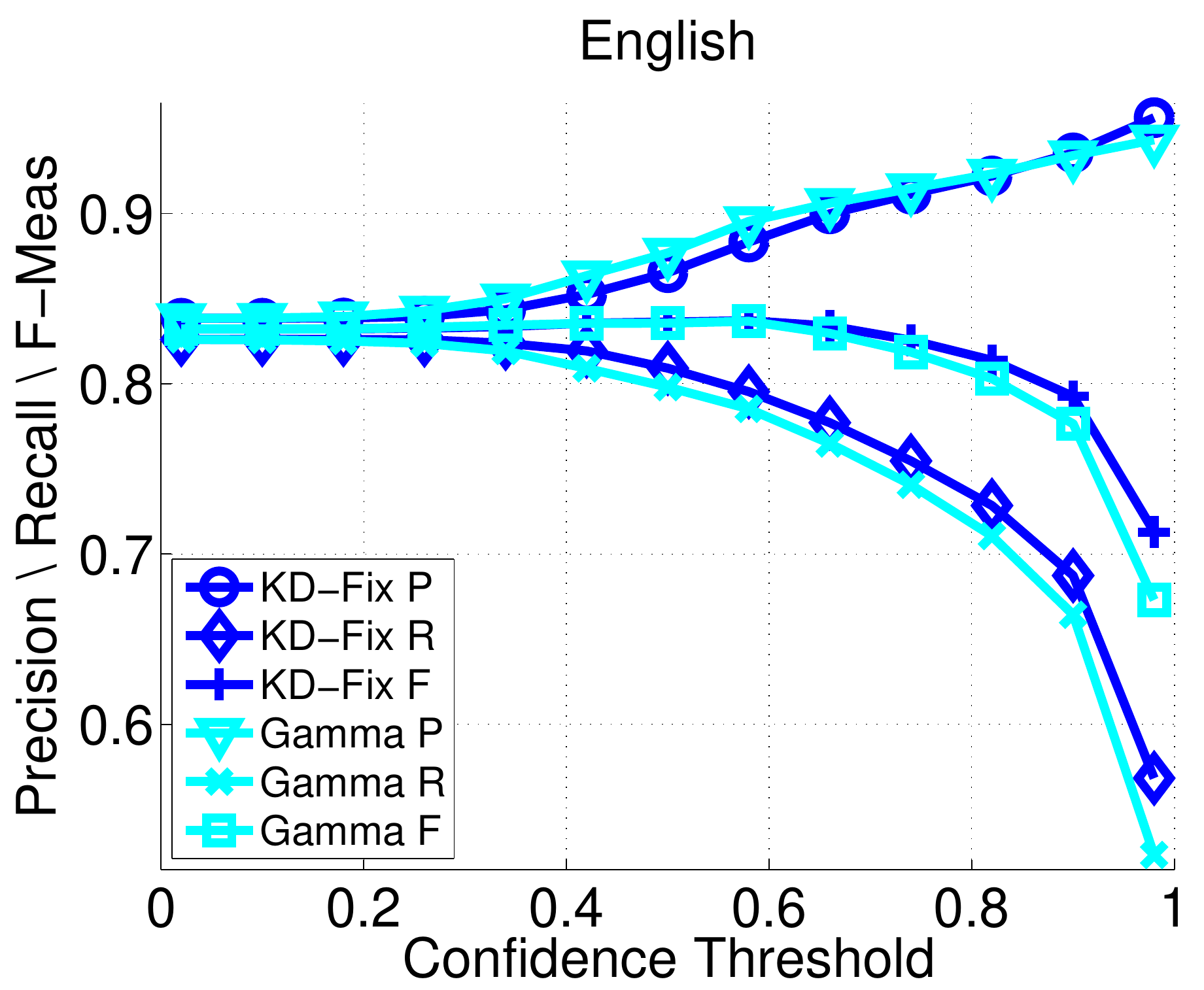}}&
      {\includegraphics[width=0.43\textwidth]{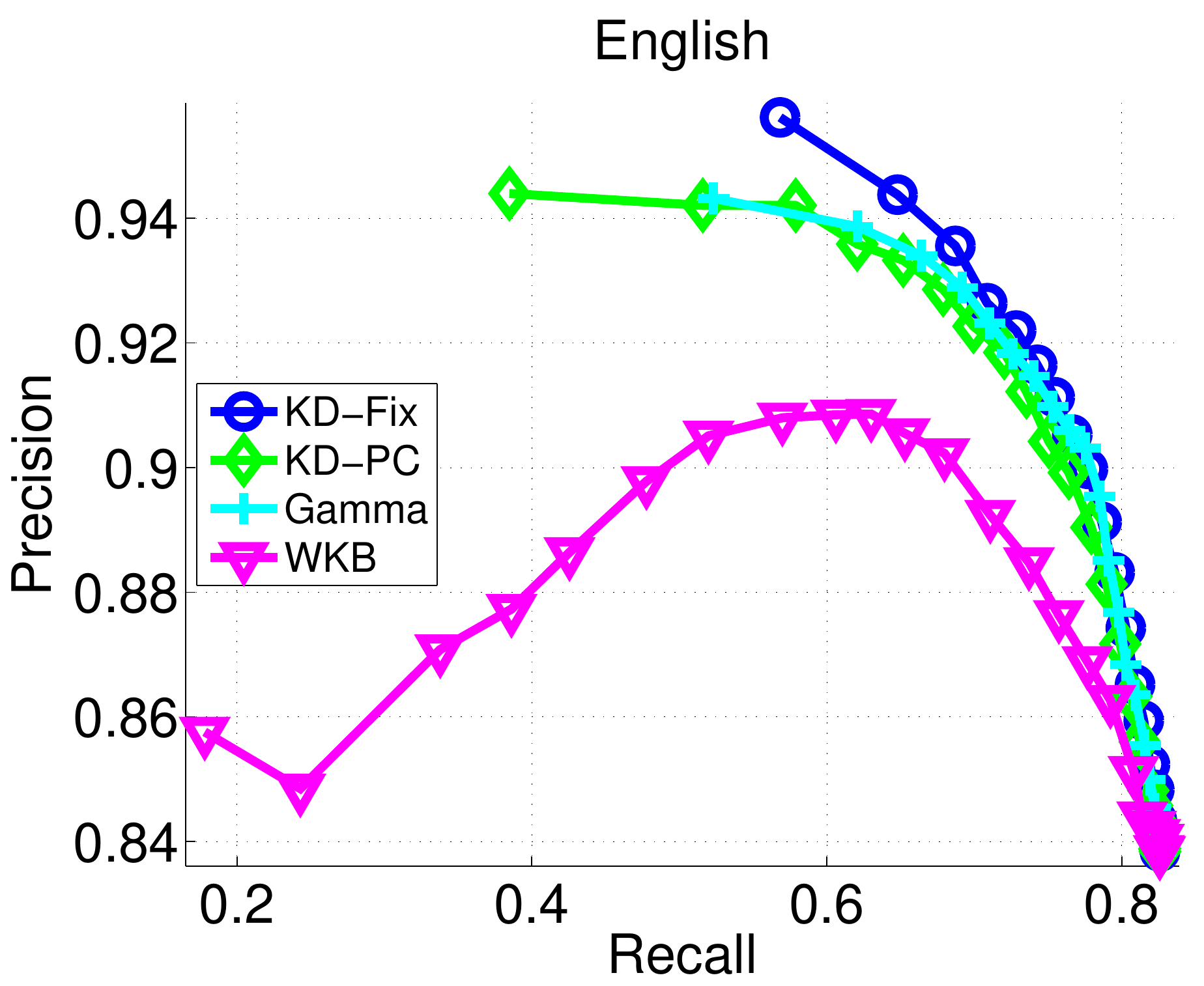}}\\
      {\includegraphics[width=0.43\textwidth]{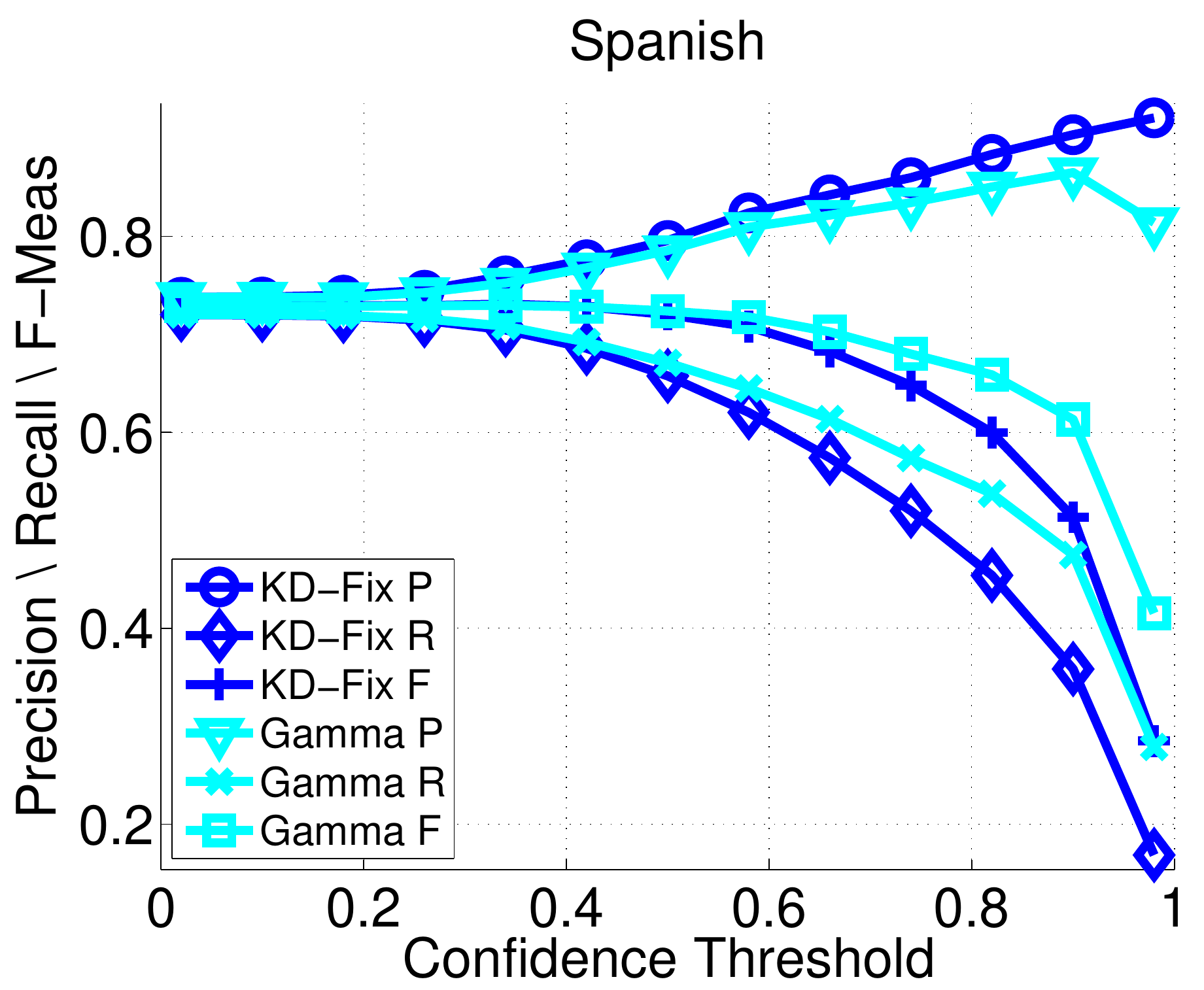}}&
      {\includegraphics[width=0.43\textwidth]{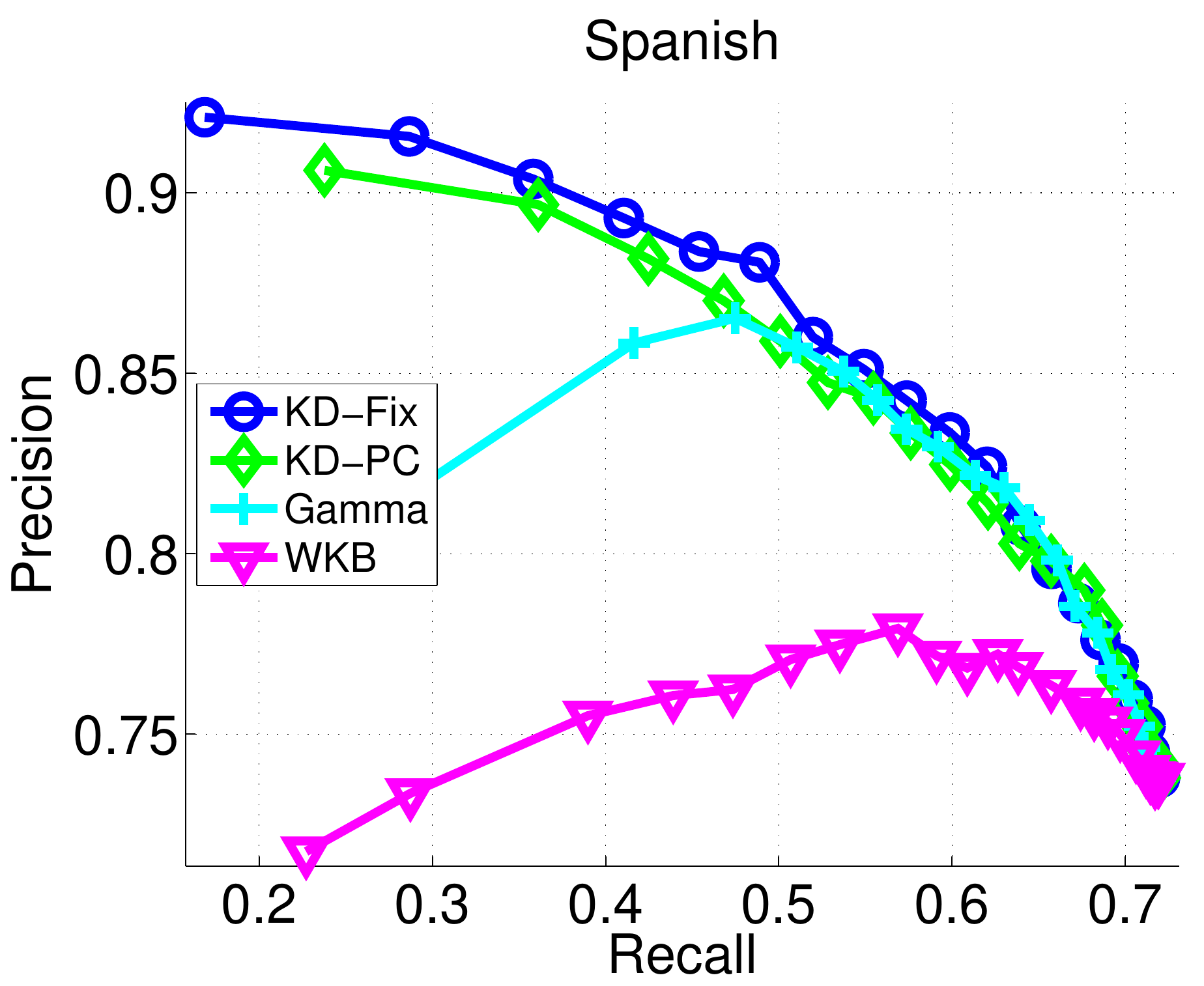}}\\
      {\includegraphics[width=0.43\textwidth]{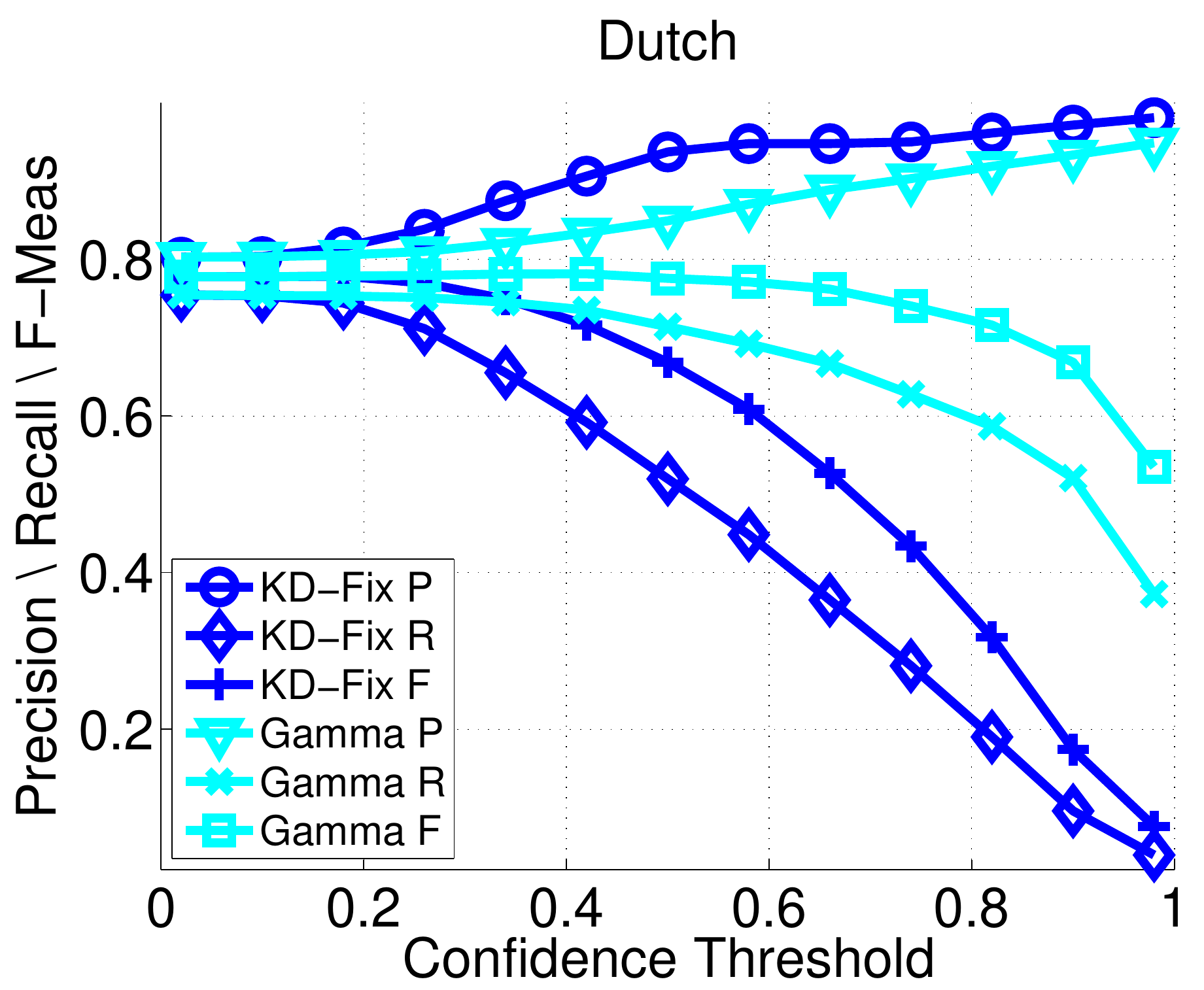}}&
      {\includegraphics[width=0.43\textwidth]{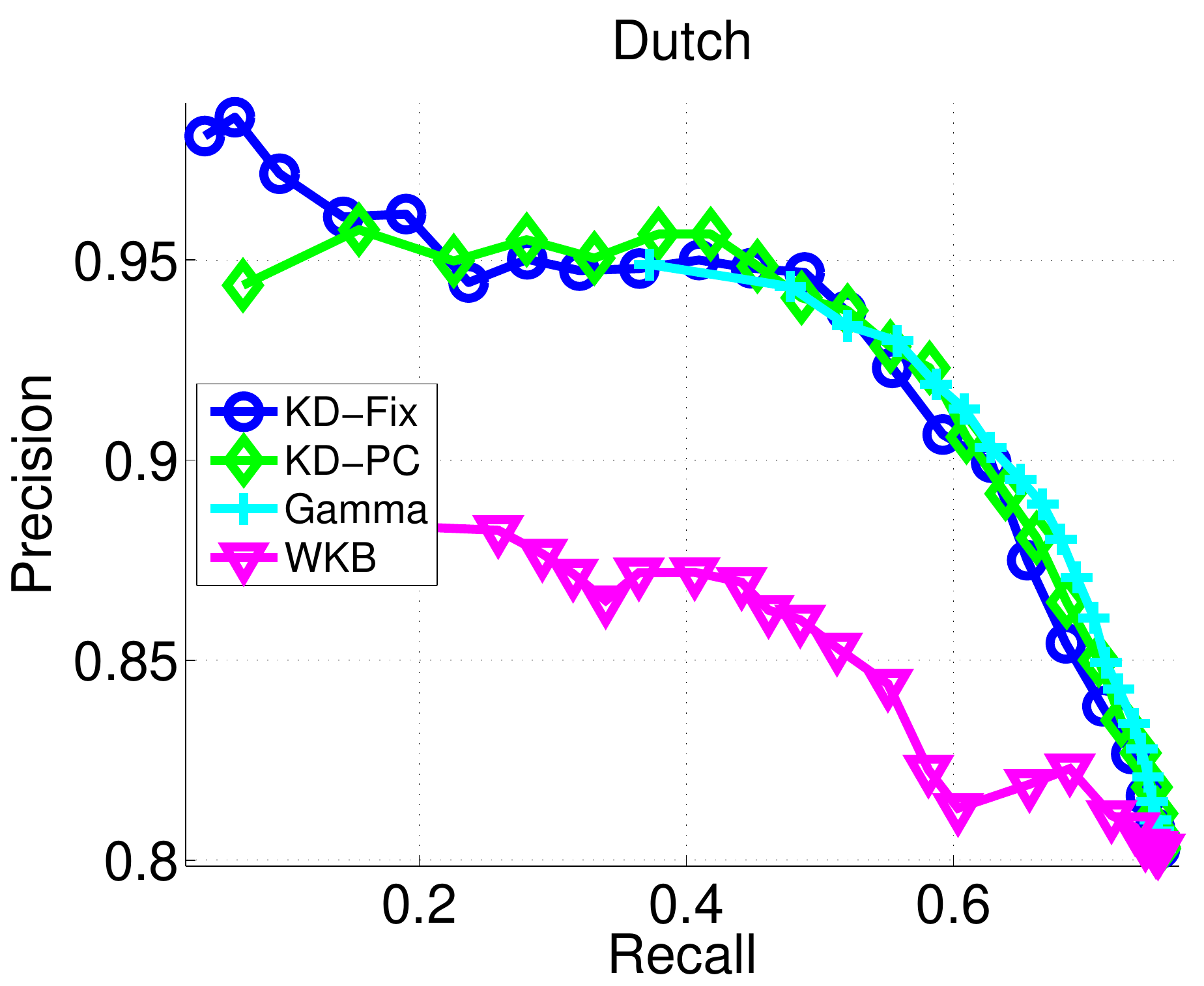}}
    \end{tabular} \caption{Trading recall for higher precision in NER. Left column
      present the precision, recall and f-measure as
      the confidence threshold is increased from 0 to 1 reflecting
      stronger bias for precision. The three lines of the same color
      are always (top to bottom) Precision, F-measure and Recall.
      Right column present precision-recall scores comparison
      for performing the tradeoff based on confidence scores provided
      by the different methods. }
\label{fig:sequences_precision_recall_tradeoff_CW}
    \end{centering}
\end{figure}

Labeling all words with a specific tag yields maximal recall for that
tag, as {\em all} words occurrence of that tag are labeled correctly,
yet with the price of low precision, as many words are tagged
wrongly. On the other hand, not labeling even one word (or labeling
all word with the special tag of {\em no tag}) yields maximal
precision for all tags, as there is not even a single word that is
labeled mistakenly by some tag, yet the precision is zero as none of
the words that should be tagged are indeed tagged. Clearly, one can
move from the second extreme to the first by labeling or tagging more
and more words with some tag.  In various information extraction
systems it is desirable to have the ability to control the tradeoff
between high precision or high recall. In some scenarios a user may
prefer that the system may label fewer words with a tag (low recall)
yet to have the labeled words be tagged correctly (high precision),
and in another scenarios the opposite is preferred, that is, labeling
more words with some tag, yet at the price of low precision.

We propose the ability to perform such a system based on the
confidence information outputted by various algorithms. We demonstrate
the system using the task of named entity recognition. First, a NER
algorithm produces a labeling for a given sentence, and then a
confidence algorithm is used to output additional confidence score for
each label. The system uses a tradeoff parameter $t\in[0,1]$ to shift between
the two extremes of high-recall and high-precision. The label of all
words labeled with some tag, such as {\em Per} or {\em Loc} that their confidence is
below the threshold $t$ is replaced by the {\em N} label indicating that no tag
is associated with that word. The larger the value of $t$
is, the tag of more words that were labeled as named entities will be modified to a no-tag and the recall
would be reduced. The lower the value of $t$ is, the lower number of
words for which their tag is modified, and the recall is higher. In
the optimal case, with a perfect confidence estimation algorithm, only the tag of
words labeled with incorrect tag would been changed, yielding higher precision with no decrease of recall.
\figref{fig:precision_recall_tradeoff_example} illustrates such NER system. The words \emph{Morgan Stanley} are first labeled as \emph{Person} with low confidence score of $0.7$, the word \emph{Google} is labeled as \emph{Organization} with high confidence score of $0.98$ and the rest of the words are labeled with confidence score $1.0$ as {\em N} indicating they are not part of named entity. In this example the tradeoff threshold $t$ is set to $0.75$, so the labels of \emph{Morgan Stanley} that have confidence score below the threshold are replaced by {\em N} while the label of \emph{Google} is not replaced as it has confidence score above the threshold. In this example, indeed \emph{Morgan Stanley} was incorrectly labeled as \emph{Person} instead of \emph{Organization} so the replacement operation improves the system's precision.

The left column of \figref{fig:sequences_precision_recall_tradeoff_CW}
shows the NER precision, recall and f-measure as the confidence
threshold value $t$ increases from $0$ to $1$ which causes
increasingly more tags such as {\em Per} to be replaced to a non-tag
{\em N}. Each plot presents two sets of curves when using the
confidence scores output by the {\tt KD-Fix} algorithm and {\tt
  Gamma} algorithm. The performance of {\tt KD-PC} plot is similar to
that of {\tt KD-Fix} and of {\tt WKB} is worse. Thus, both methods
omitted for clarity. {\tt Delta} is omitted since its output is not in
the desired range of absolute confidence prediction.
The top, middle and bottom curves for each of the two algorithms are
precision, f-measure and recall respectively.

For English we observe a similar trend for both algorithms. For
confidence threshold lower than $\approx0.3$ no labels are replaced
and the precision and recall scores remains constant. For larger
confidence threshold values the precision score increases as we
expected, yet the recall score drops. This indicates that both correct
and incorrect NE tags are being replaced with a non-tag {\em N}. For
confidence threshold values between $\approx0.3-0.7$ the f-measure
score remains about constant while precision score increases from
$0.83$ to $0.90$ and recall score drops from $0.82$ to $0.77$. This is
a successful trade-off between recall and precision for the optimal
f-measure value. For larger confidence threshold values $\approx0.7-1$
the precision score continues to increase up to $0.95$, yet at the
price of significant drop in the recall score to $0.56$ and the
f-measure score drops as well.

For NER in Spanish, we observe that for confidence thresholds up to $0.6$, both methods allow similar balanced precision recall tradeoff such that the f-measure score is steady.
Yet, for threshold values greater than $0.6$ the tradeoff based on
{\tt KD-Fix} confidence scores improves precision at cost of decreased
recall more aggressively compared to tradeoff based on {\tt Gamma}.
The maximal precision score achieved using {\tt KD-Fix} is $0.92$
compared to only $0.87$ with {\tt Gamma}. In Spanish NER we also
observe that when using {\tt Gamma} for confidence, the precision
score starts to drop along with recall scores for threshold values
greater than $0.9$. This indicates that more correct NE tags are
replaced with a no-tag {\em N} compared with incorrect NE tags, which
clearly is undesirable. A similar behavior of more aggressive tradeoff
between precision and recall for {\tt KD-Fix} than for {\tt Gamma} is
observed for Dutch.
Here, the maximal precision score achieved using {\tt KD-Fix} is
$0.99$ compared to only $0.95$ with {\tt Gamma}. Determining which
behavior is preferable, aggressive or passive tradeoff, is application
specific, yet for any given precision score having higher recall score
is clearly preferable.
For NP-chunking we observed similar trend (plots omitted) yet as precision starts from very high score of $0.95$ the increase was less significant compared to the NER cases.

An alternative view of the same tradeoff is shown using the standard
precision-recall curves in the plots in the right column of
\figref{fig:sequences_precision_recall_tradeoff_CW}. Each of the
plots shows the precision vs. recall for four confidence estimation
methods algorithms: {\tt KD-Fix}, {\tt KD-PC}, {\tt Gamma} and {\tt
  WKB}, one plot for each dataset. The bottom-right point of each
curve reflects the precision and recall before using our algorithms to
filter or reduce tags. (In fact, all curves coincide at this point as the same model is used to make label or tag the test data.)

As the confidence threshold value increases precision score improves
and recall score decreases, yet each method goes via different route
of trading-off precision and recall. An optimal confidence estimation
method distinguishes between correct and incorrect NE tags labels, and
thus improves precision while maintaining a constant recall values. Such optimal behavior is reflected by a vertical curve at the right area of the plots.

For all datasets {\tt WKB} confidence score yields poor
performance where the precision gain is small compared to the loss
in recall. Furthermore, at some point precision drops as well.
Additionally, {\tt KD-Fix} and to some extent {\tt KD-PC} allow the
system to achieve higher-values of precision (curve ends with high
value). {\tt Gamma} achieves similar or slightly better precision
values for high-recall, yet is not able to improve precision compared
with {\tt KD-Fix} (NER English and NER Dutch), or even the precision
drops together with recall (NER Spanish).

We experimented also in performing tradeoff in the opposite direction:
trading precision for higher recall. We used similar approach as
described above. For words labeled with a no-tag {\em N} and with
confidence score lower than some threshold $t$, we replaced the {\em
  N} with some tag, which had the second highest score value according
to the prediction model. By construction, a tag of some NE (such as
{\em Per}) will by chosen. This tradeoff task is harder than
trading-off recall and precision as described above, as even if a word
that is incorrectly labeled with a no-tag {\em N} is identified, still the
algorithm is required to choose what tag should it be labeled with. We observed that our confidence estimation methods detect
effectively incorrect no-tag labels {\em N}. Using {\tt KD-Fix} method
the average confidence score of all the words labeled incorrectly with
a no-tag {\em N} is $0.6$, $0.4$ and $0.3$ respectively in NER
English, Spanish and Dutch. These values are low compared to the high
average confidence scores of $0.99$, $0.98$ and $0.92$ respectively, of all words assigned correctly with a
no-tag {\em N} (\figref{fig:sequences_confidence_bins_CW_population} illustrates the low overlapping in the distribution of confidence scores between the correct and incorrect labels).
Yet choosing the correct tag for a word is a hard task. Eventually recall values for
all three languages improved at most by $0.02$ while precision score
dropped by $\approx0.2$.

\begin{figure}[!h!] \begin{centering} \begin{tabular}{cc}
      {\includegraphics[width=0.43\textwidth]{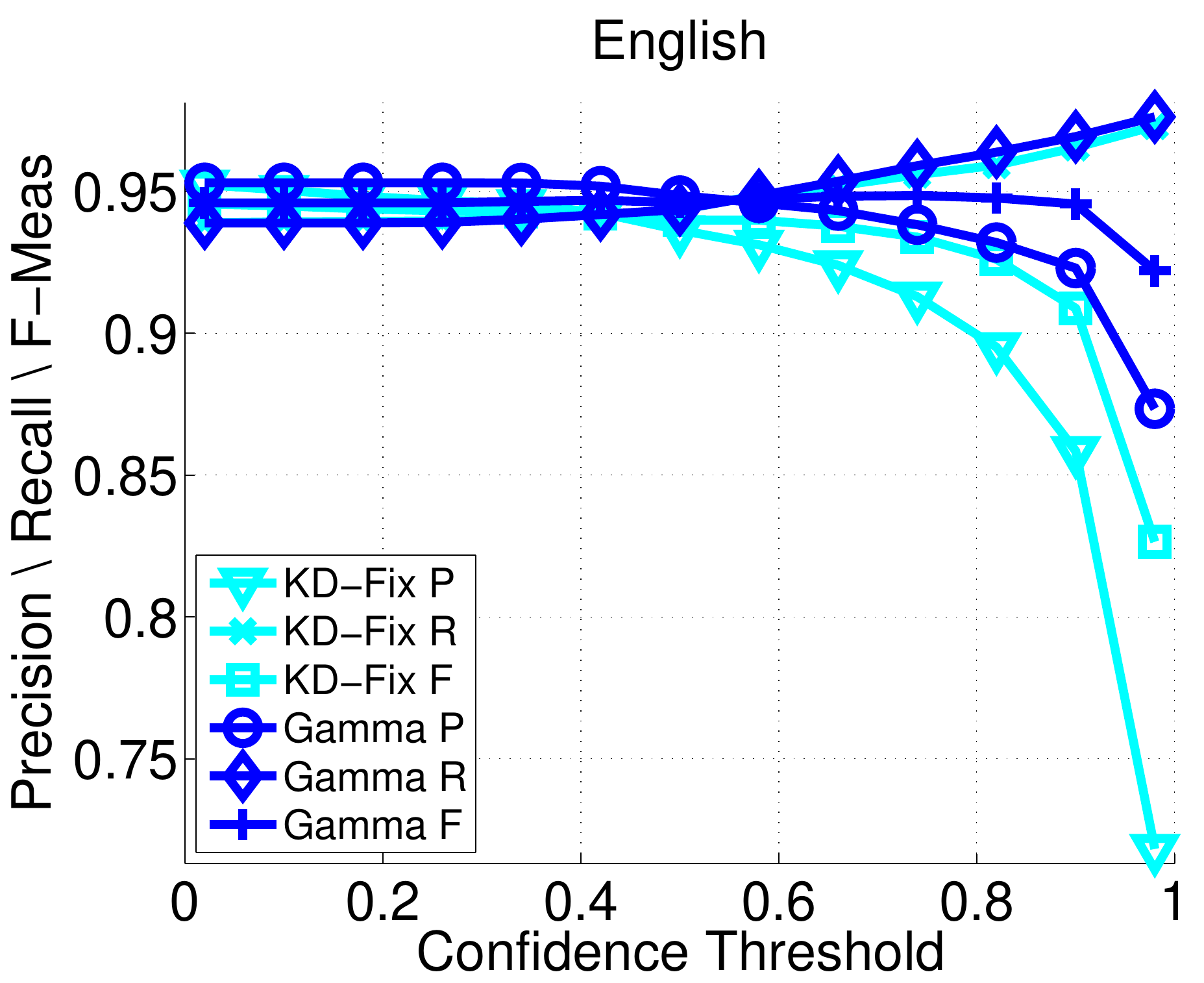}}&
      {\includegraphics[width=0.43\textwidth]{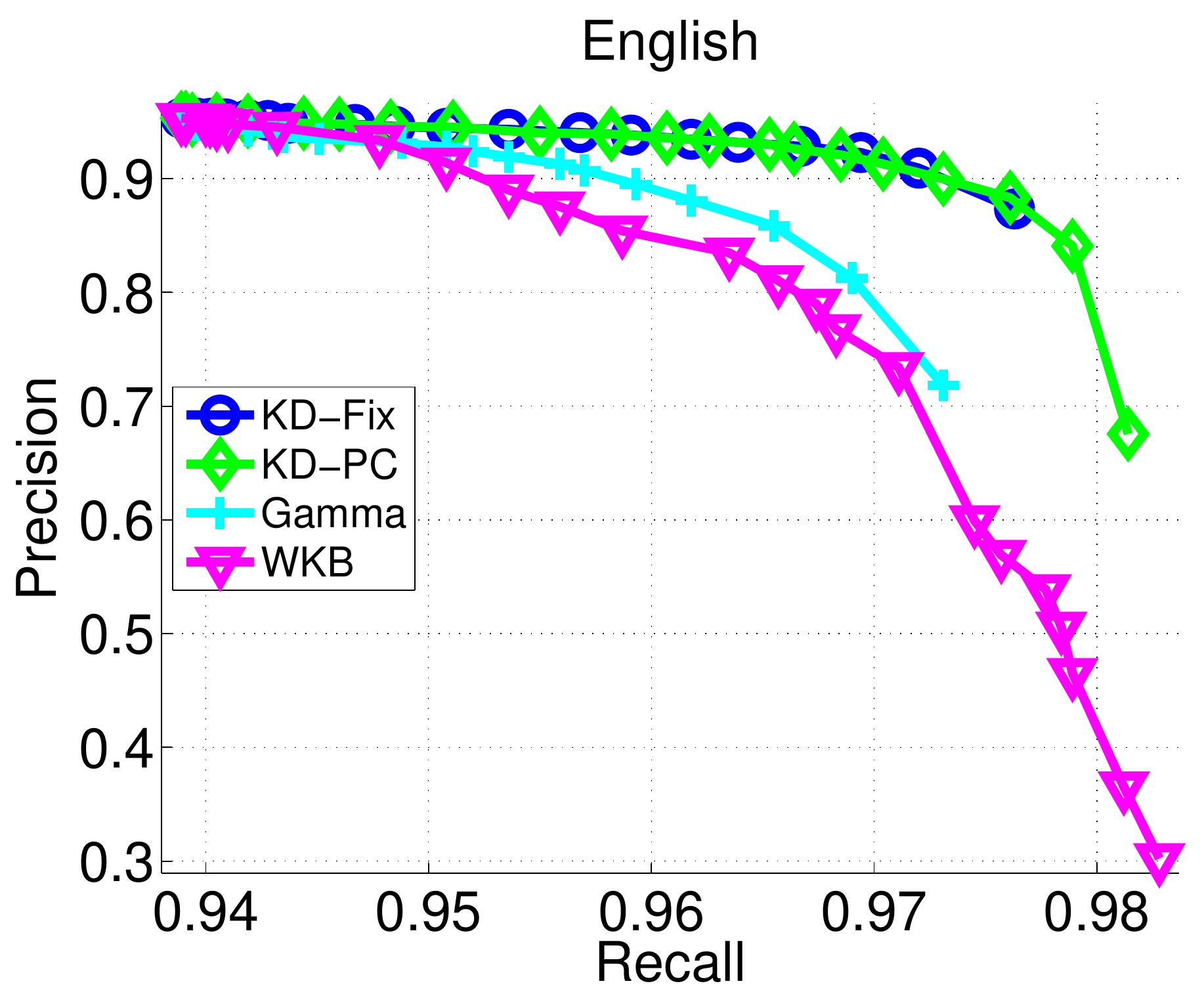}}\\
      {\includegraphics[width=0.43\textwidth]{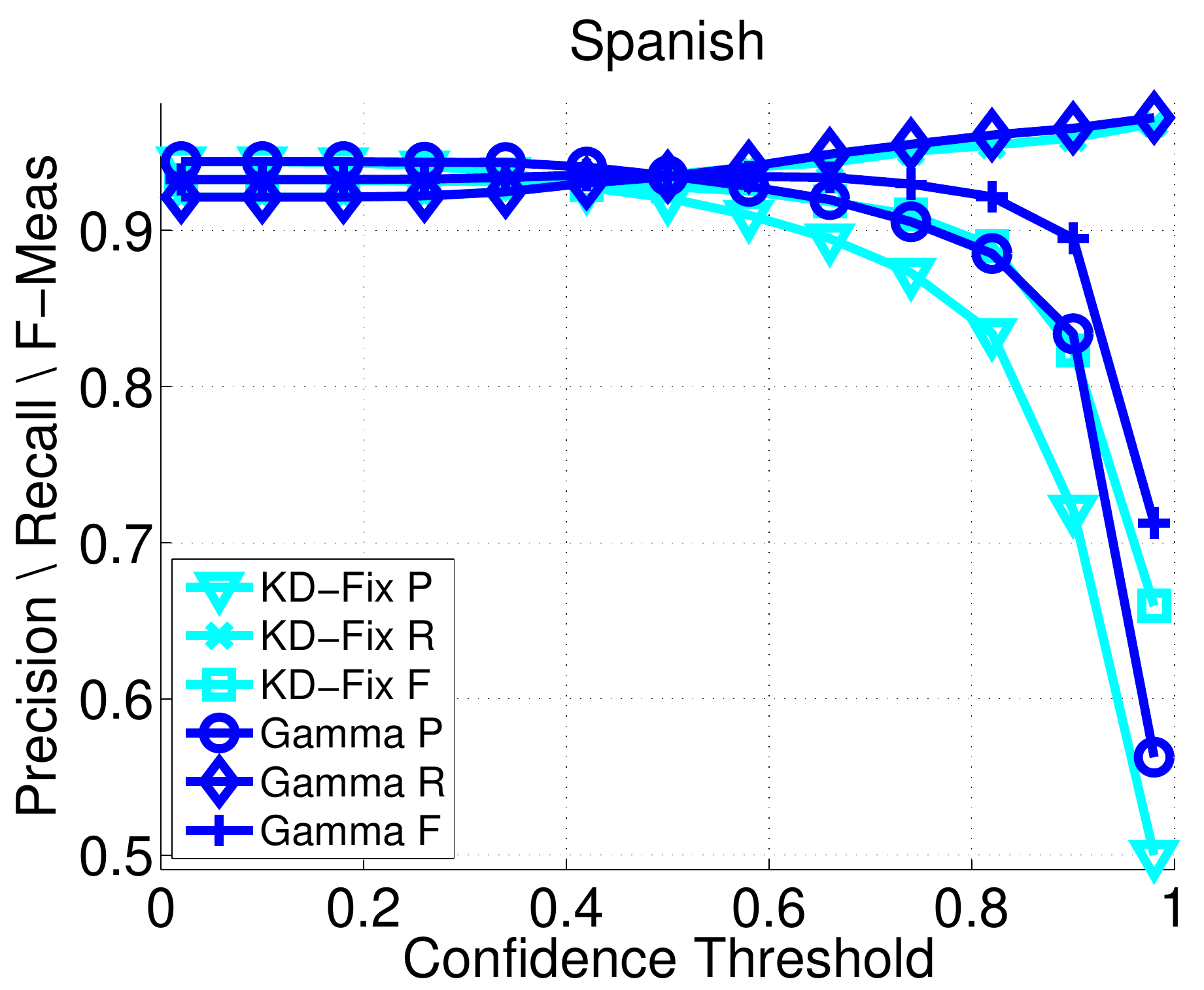}}&
      {\includegraphics[width=0.43\textwidth]{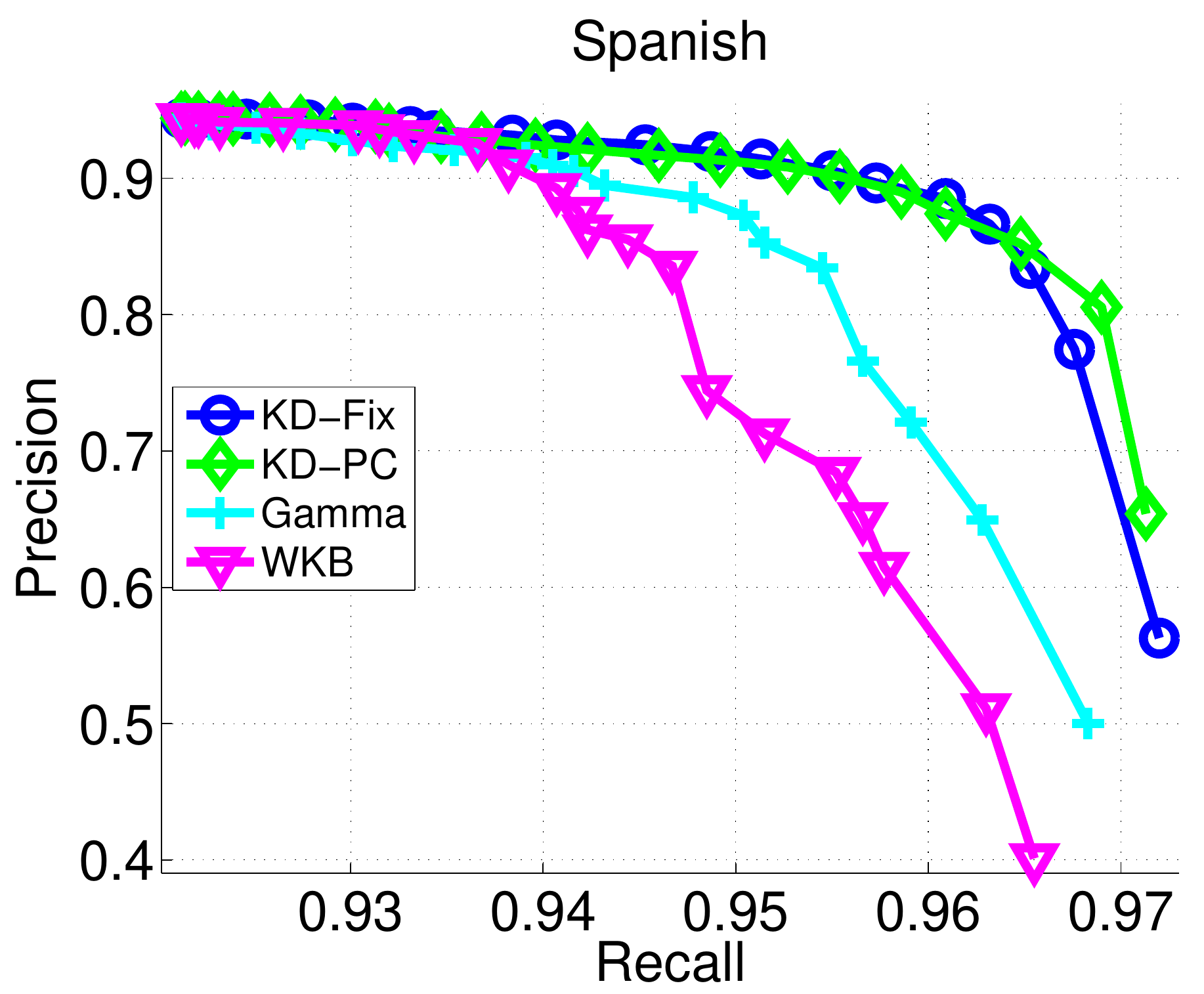}}\\
      {\includegraphics[width=0.43\textwidth]{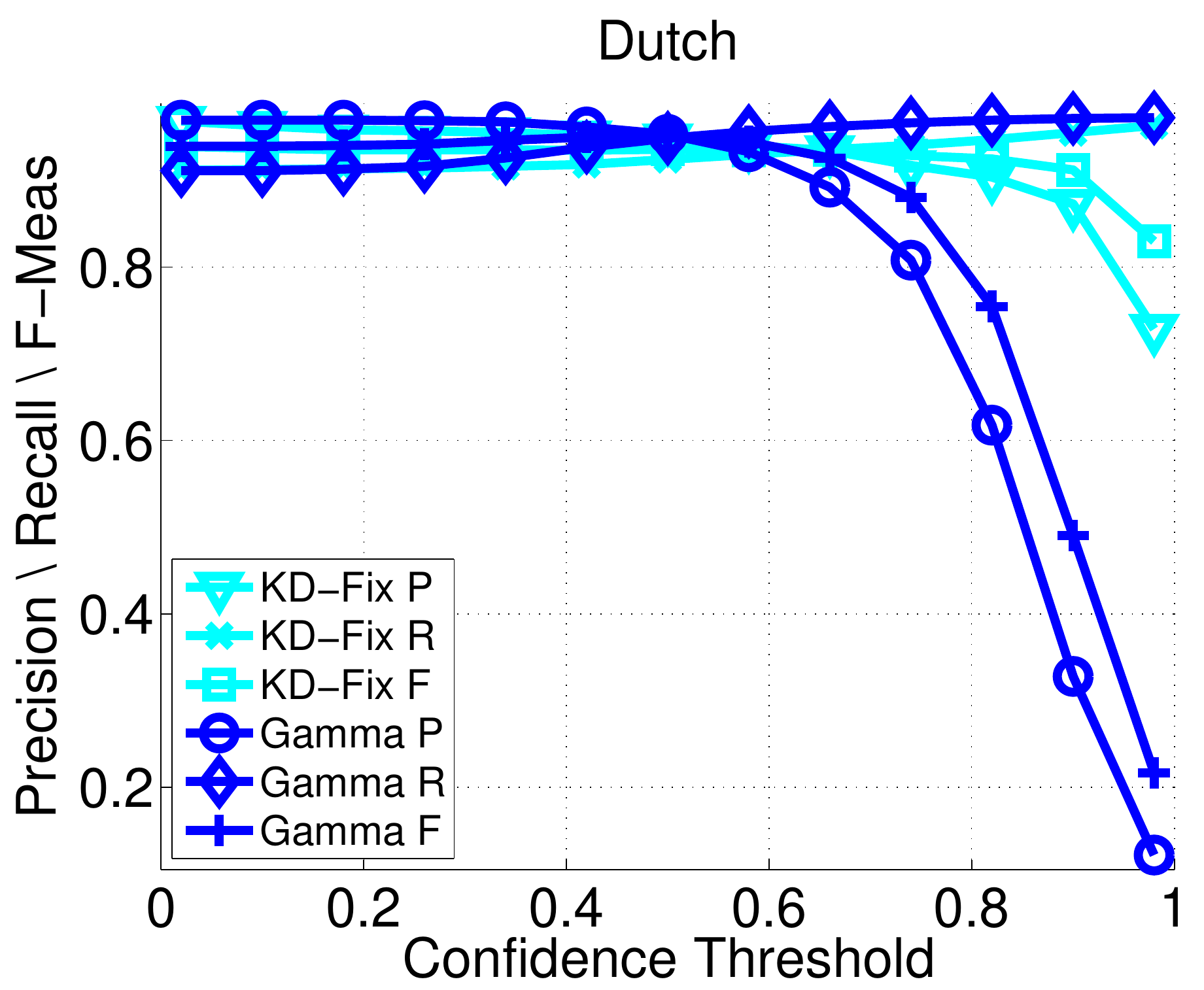}}&
      {\includegraphics[width=0.43\textwidth]{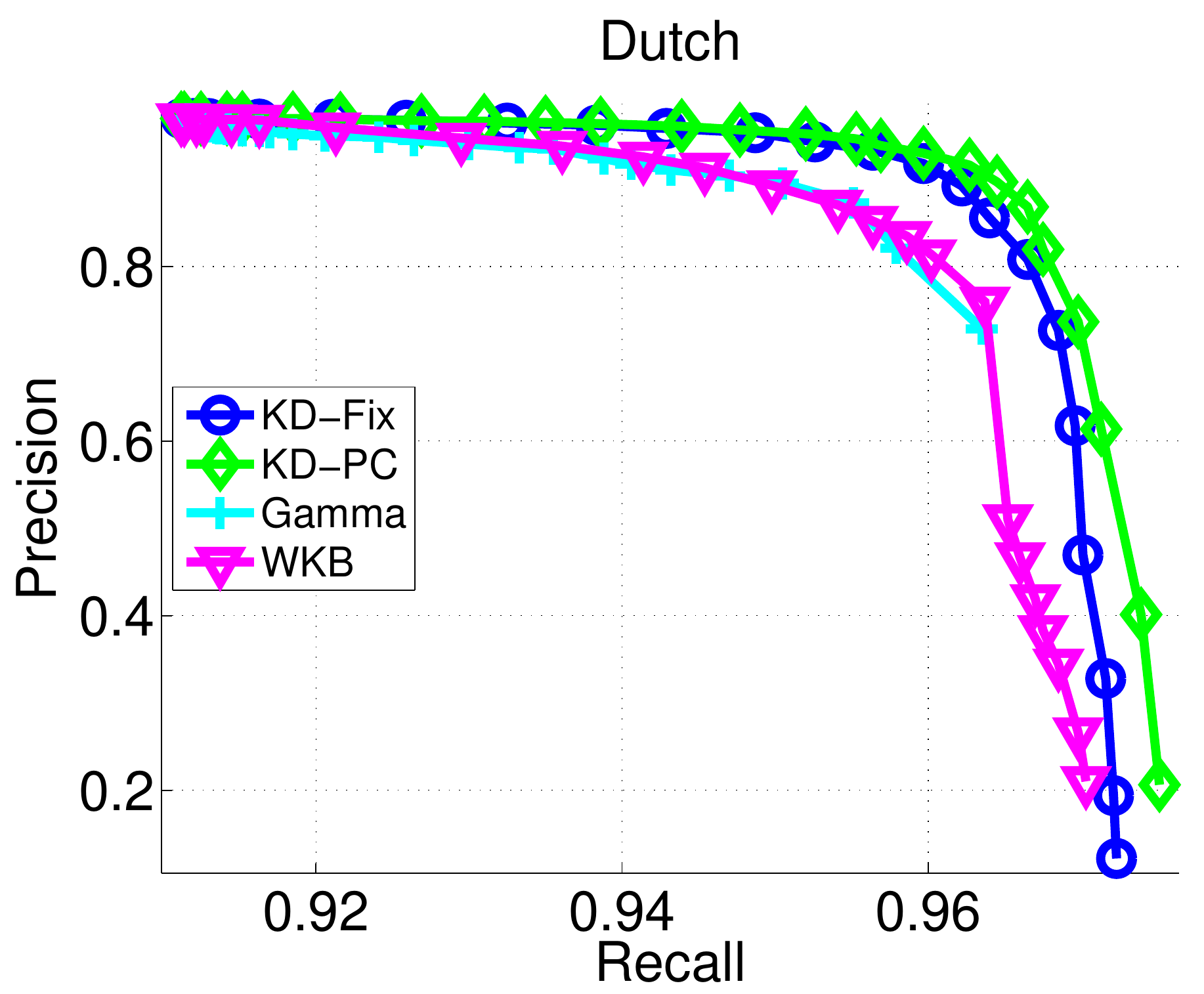}}
    \end{tabular} \caption{Trading precision for higher recall in NER. Left column
      present the precision, recall and f-measure as
      the confidence threshold is increased from 0 to 1 reflecting
      stronger bias for recall. Right column present precision-recall scores comparison
      for performing the tradeoff based on confidence scores provided
      by the different methods. }
\label{fig:sequences_recall_for_precision_tradeoff_CW}
    \end{centering}
\end{figure}

We therefore made the task easier by merging the four NE categories
({\em Person, Location, Organization, Misc}) to a single NE
category. A word that is labeled with low confidence as no-tag {\em N}
can now be tagged as NE without committing to a specific category. The
results are presented in
\figref{fig:sequences_recall_for_precision_tradeoff_CW}. We see that
using a single NE category improves the base precision and recall
scores in all three datasets, the f-measure scores for English,
Spanish and Dutch improve from $0.83$, $0.72$, $0.77$ to $0.94$,
$0.93$, $0.94$ respectively (these f-measure scores are close to the
score achieved for NP chunking task which also has just a single tag
category).

For English when using {\tt KD-Fix} confidence scores to tradeoff we
observe that for confidence threshold lower than $\approx0.3$ no
labels are replaced. Then for confidence threshold values between
$\approx0.3-0.9$ the f-measure score remains about constant (even
slightly improves) while recall score increases from $0.94$ to $0.97$
and precision score drops from $0.95$ to $0.92$. This is a balanced
trade-off between recall and precision for the optimal f-measure
value. For confidence threshold values greater than $0.9$ the recall
score increases a little more almost to $0.98$ but the precision score
drops to $0.87$ and the f-measure score drops as well.  Using {\tt
  Gamma} confidence score for this dataset allows similar improvement
in recall, but the precision scores drops significantly even for low
confidence threshold values and falls to $0.72$. Similar quantitative behavior was observed both for NER in Spanish and Dutch.

The right column of
\figref{fig:sequences_recall_for_precision_tradeoff_CW} present the
precision-recall curves using all the confidence estimation methods
algorithms: {\tt KD-Fix}, {\tt KD-PC}, {\tt Gamma} and {\tt WKB}, one
plot for each dataset. The top-left point of each curve reflects the
precision and recall before labeling any low confidence no-tags words
as NE. As the confidence threshold value increases recall score
improves and precision score decreases.

For all datasets {\tt KD-Fix} and {\tt KD-PC} perform better than the
other methods by maintaining higher precision score for the same
recall scores, and in Spanish and Dutch datasets they also allow the
system to achieve higher recall values. In English {\tt WKB} allowed
achieving the highest recall value yet at cost of very low precision.

Finally, we note that such an experiment is
not relevant in the context of parsing, as each word is assigned to
another word (with an edge), and the asymmetry in NER labels of having
tags and no-tags does not exist.

\subsection{Active Learning}
\label{sec:active}

\begin{figure}[!h!]
\begin{centering}
\begin{tabular}{cc}
{\includegraphics[width=0.44\textwidth]{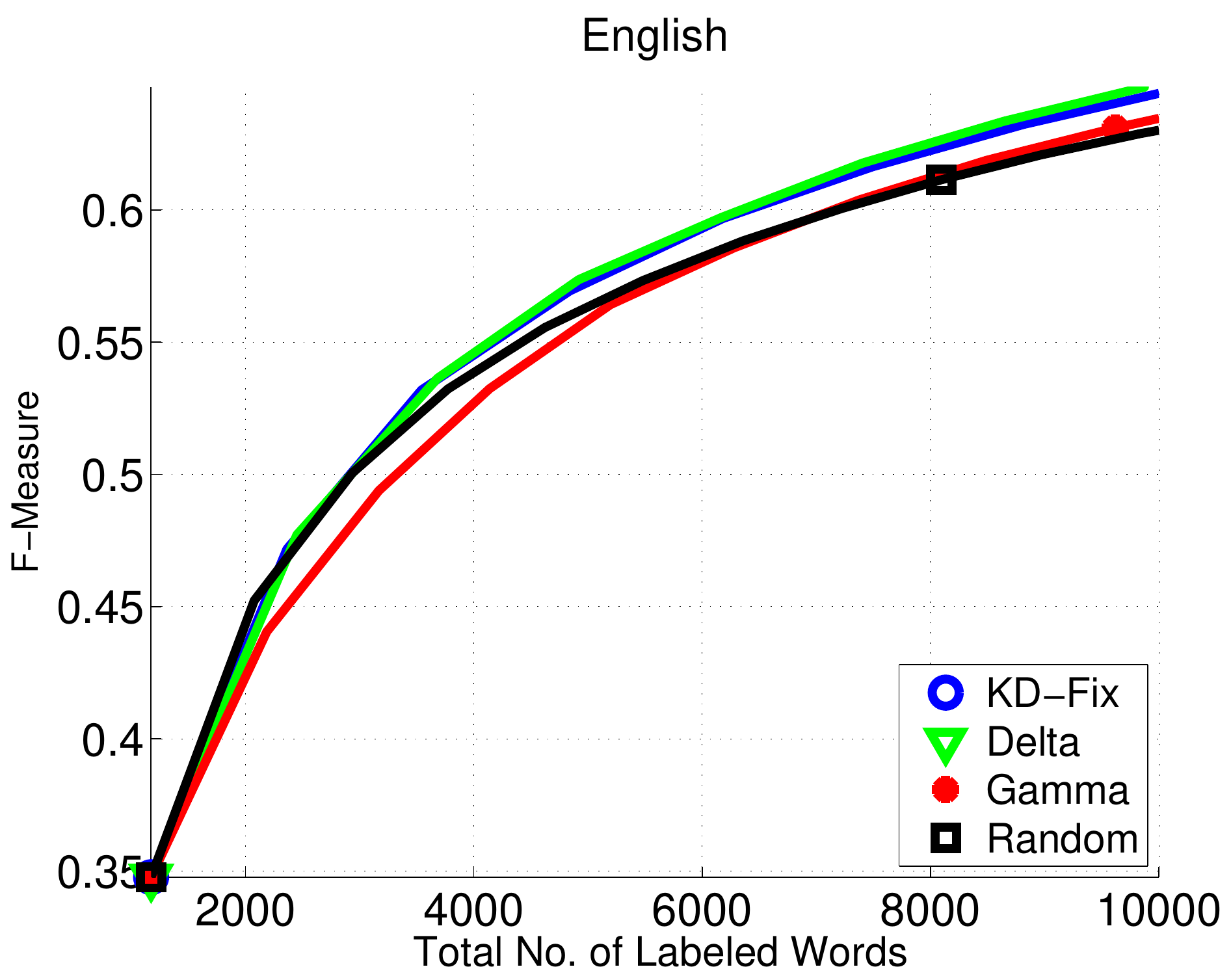}}&
{\includegraphics[width=0.44\textwidth]{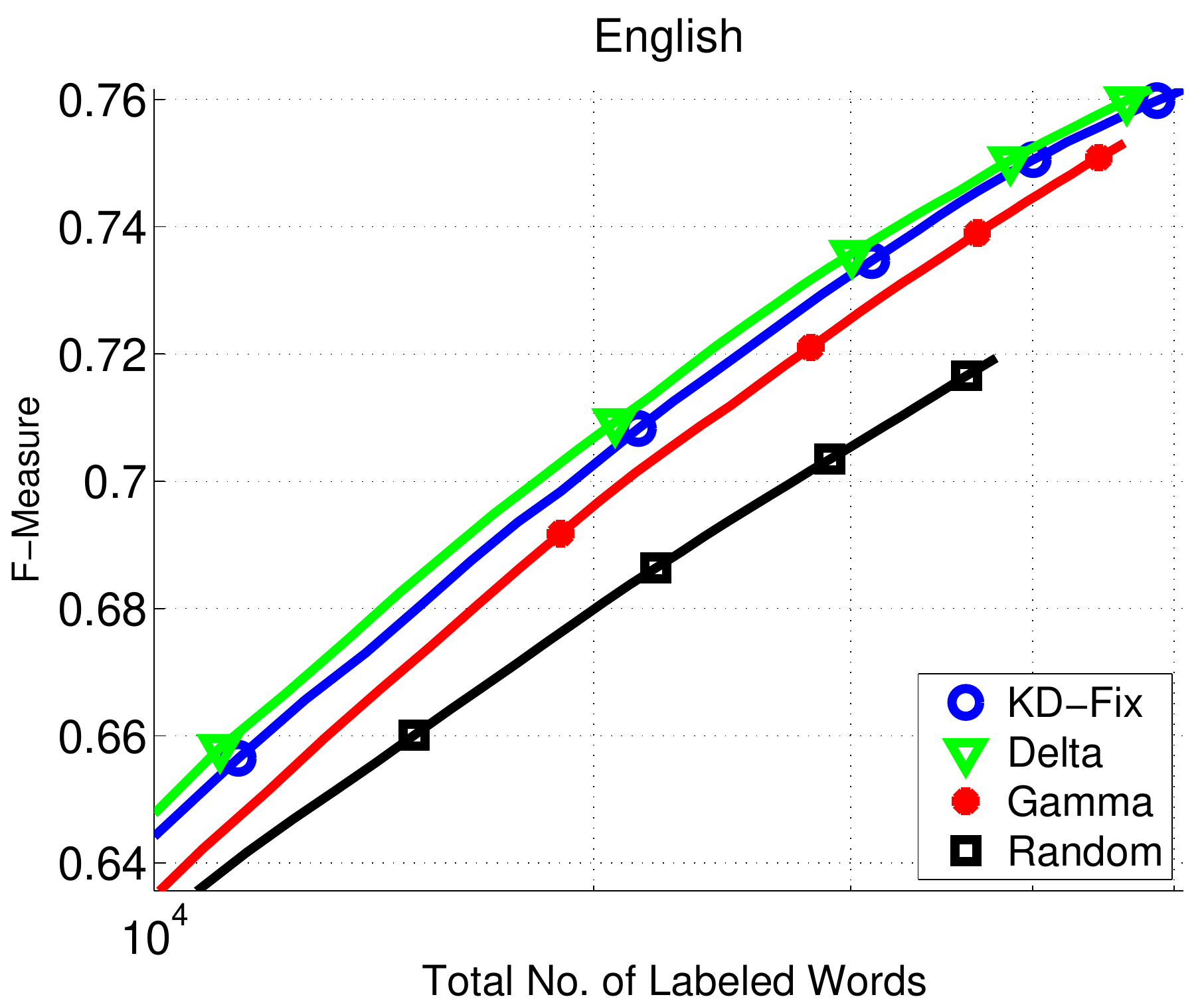}}\\
{\includegraphics[width=0.44\textwidth]{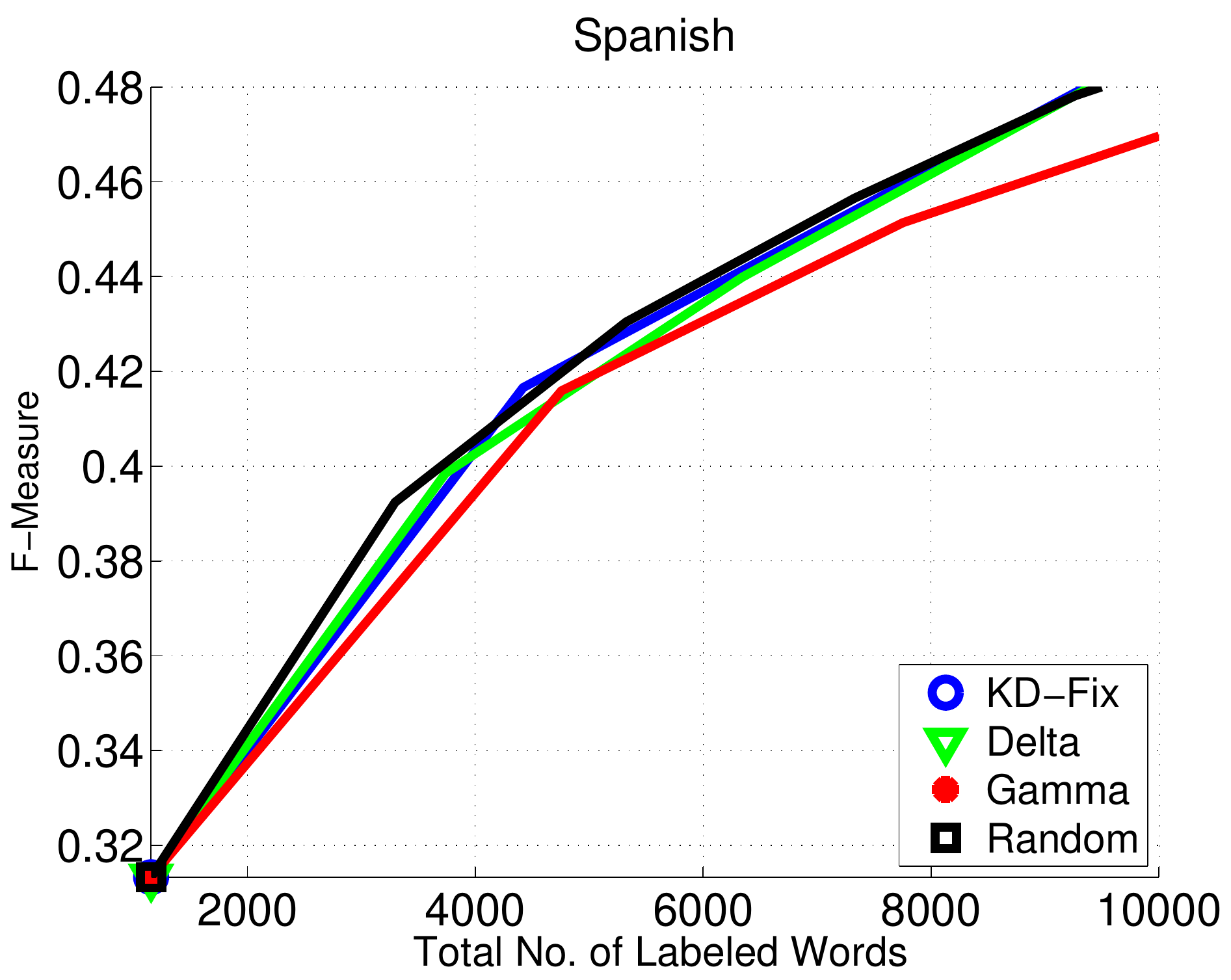}}&
{\includegraphics[width=0.44\textwidth]{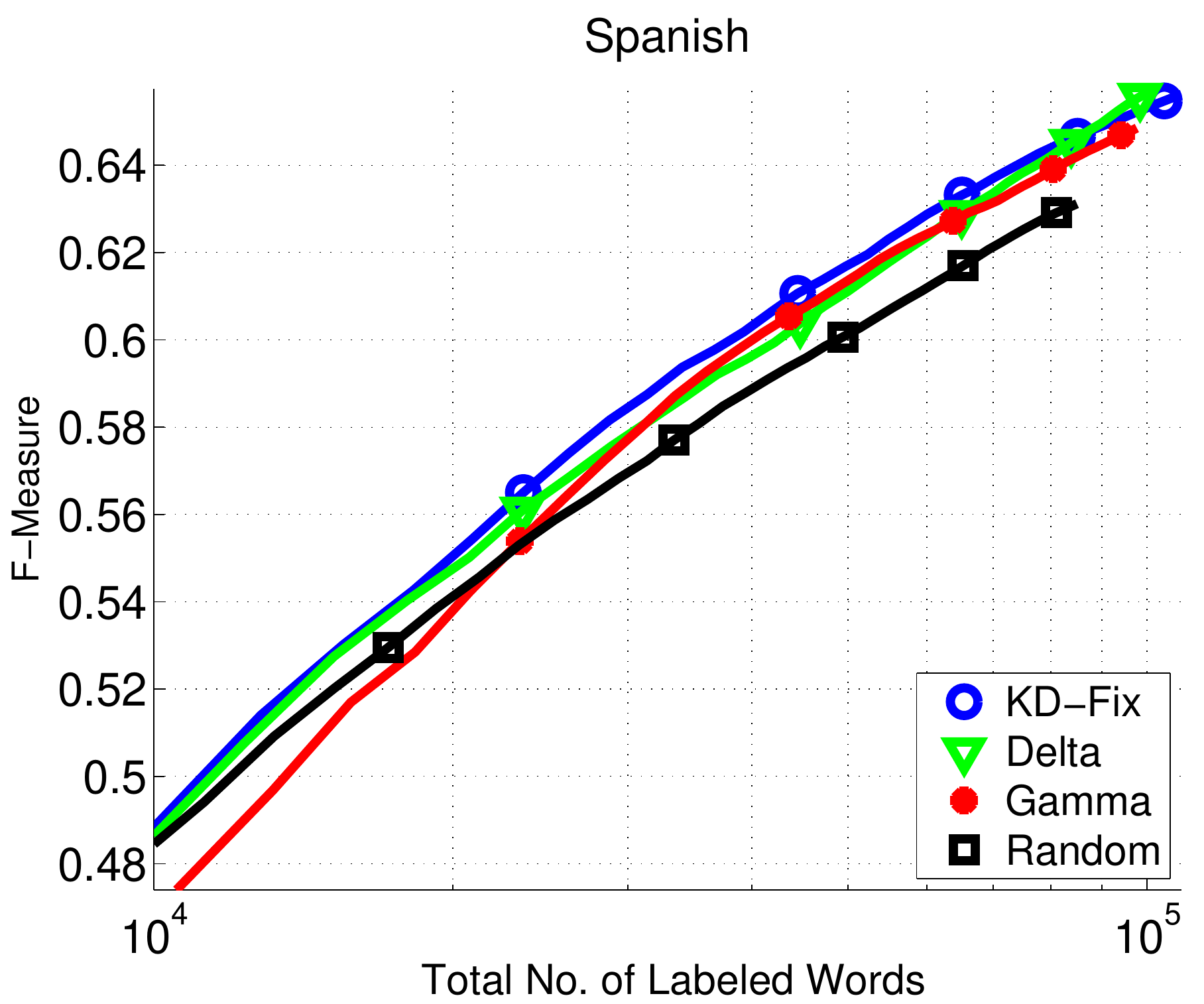}}\\
{\includegraphics[width=0.44\textwidth]{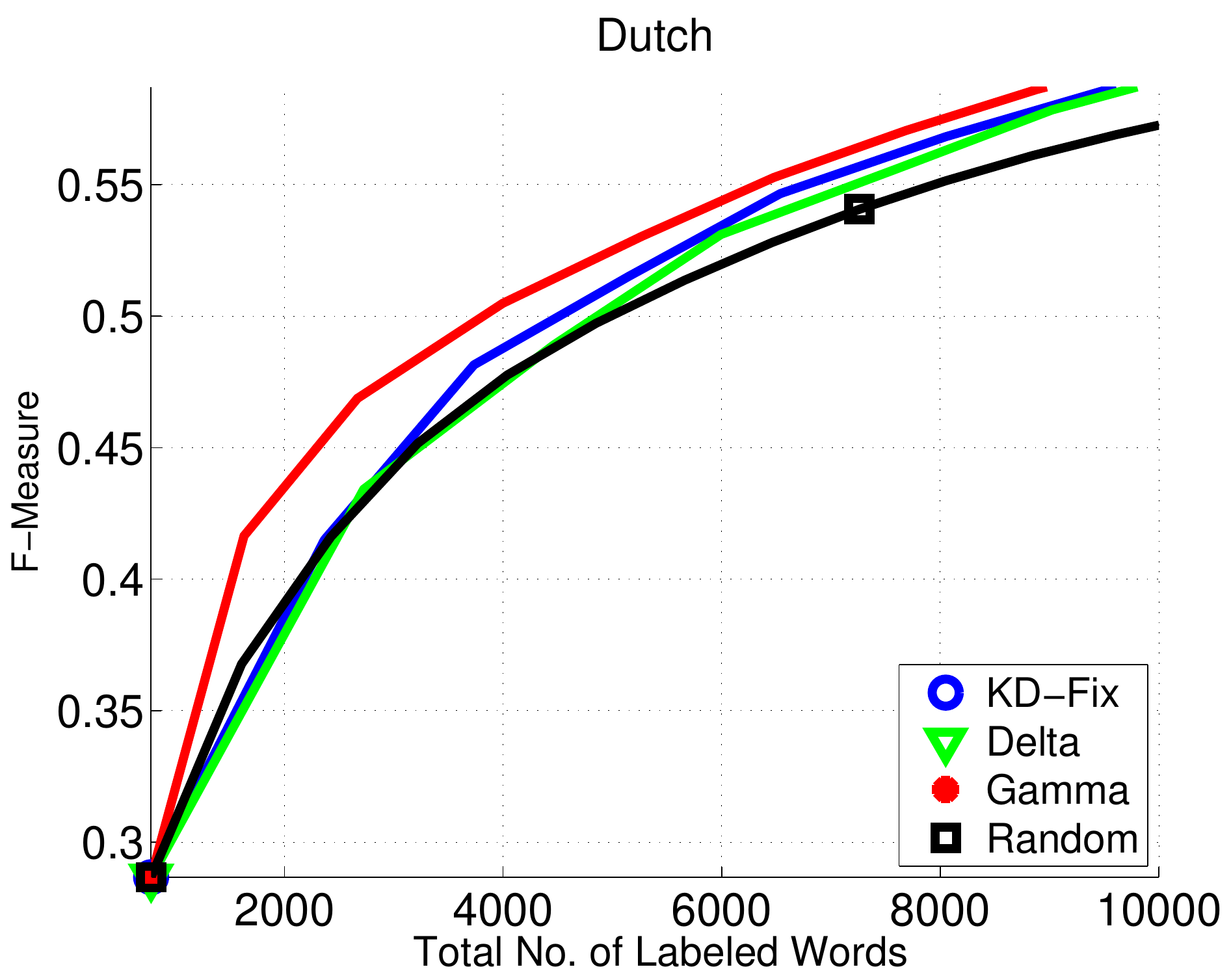}}&
{\includegraphics[width=0.44\textwidth]{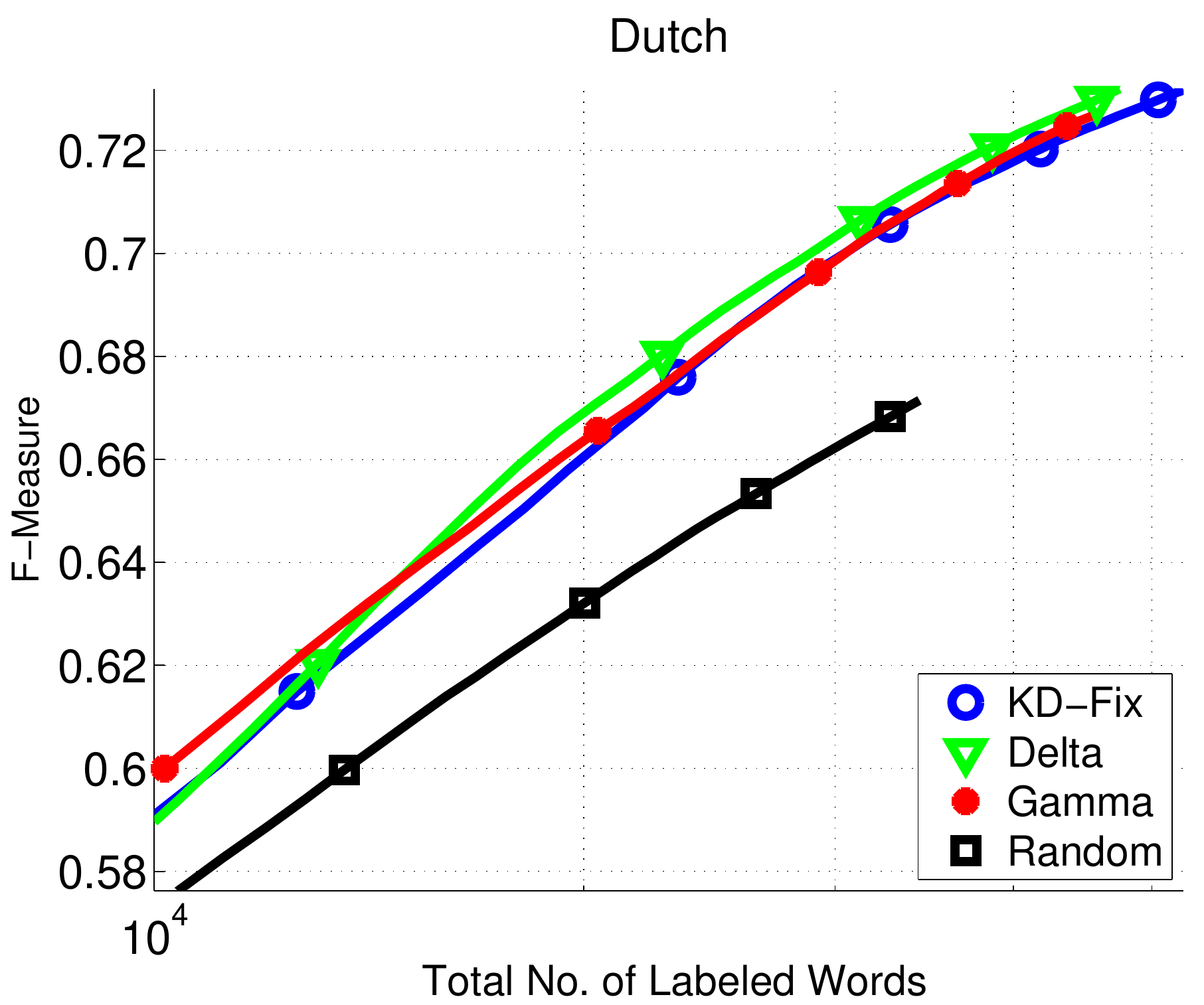}}\\
{\includegraphics[width=0.44\textwidth]{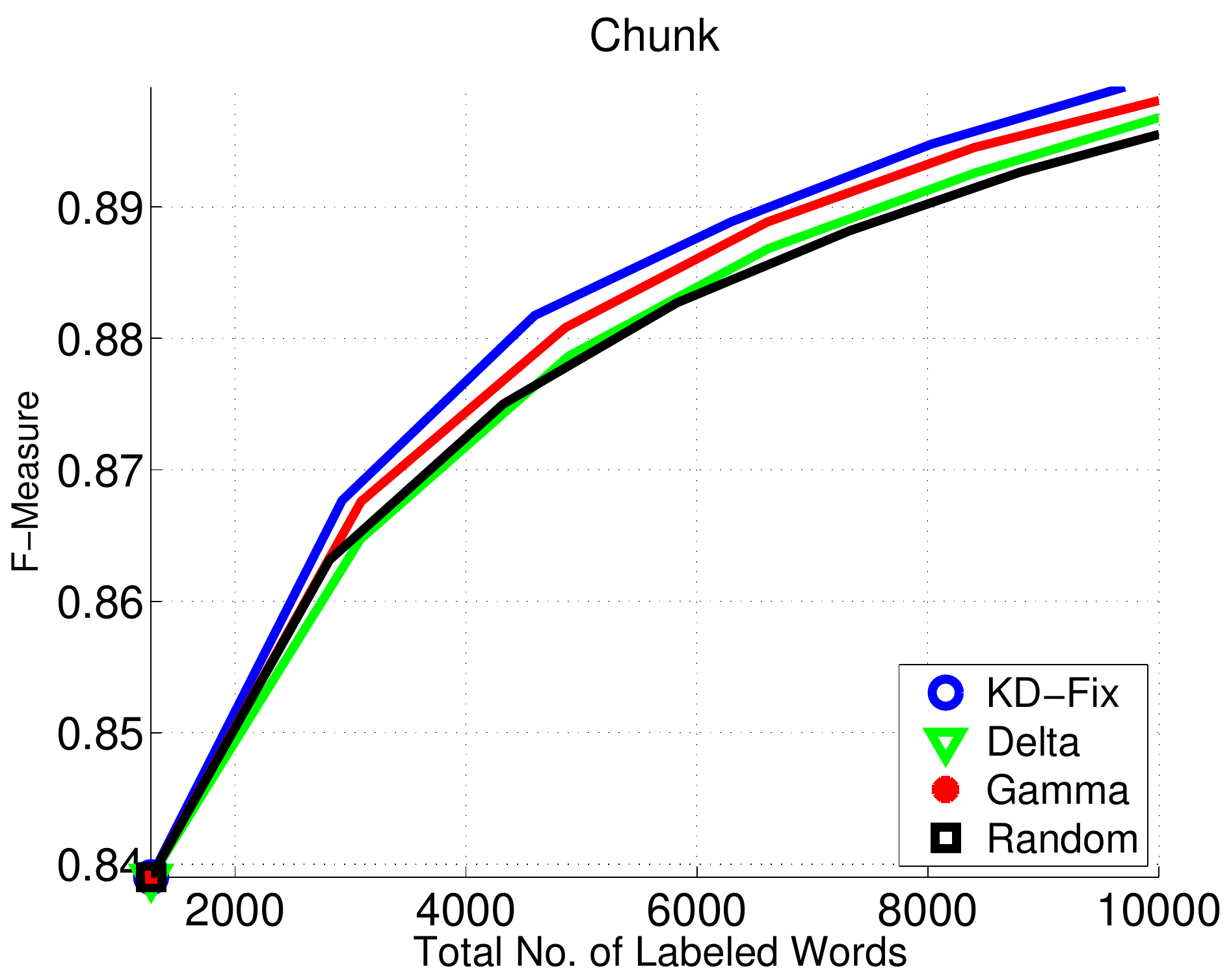}}&
{\includegraphics[width=0.44\textwidth]{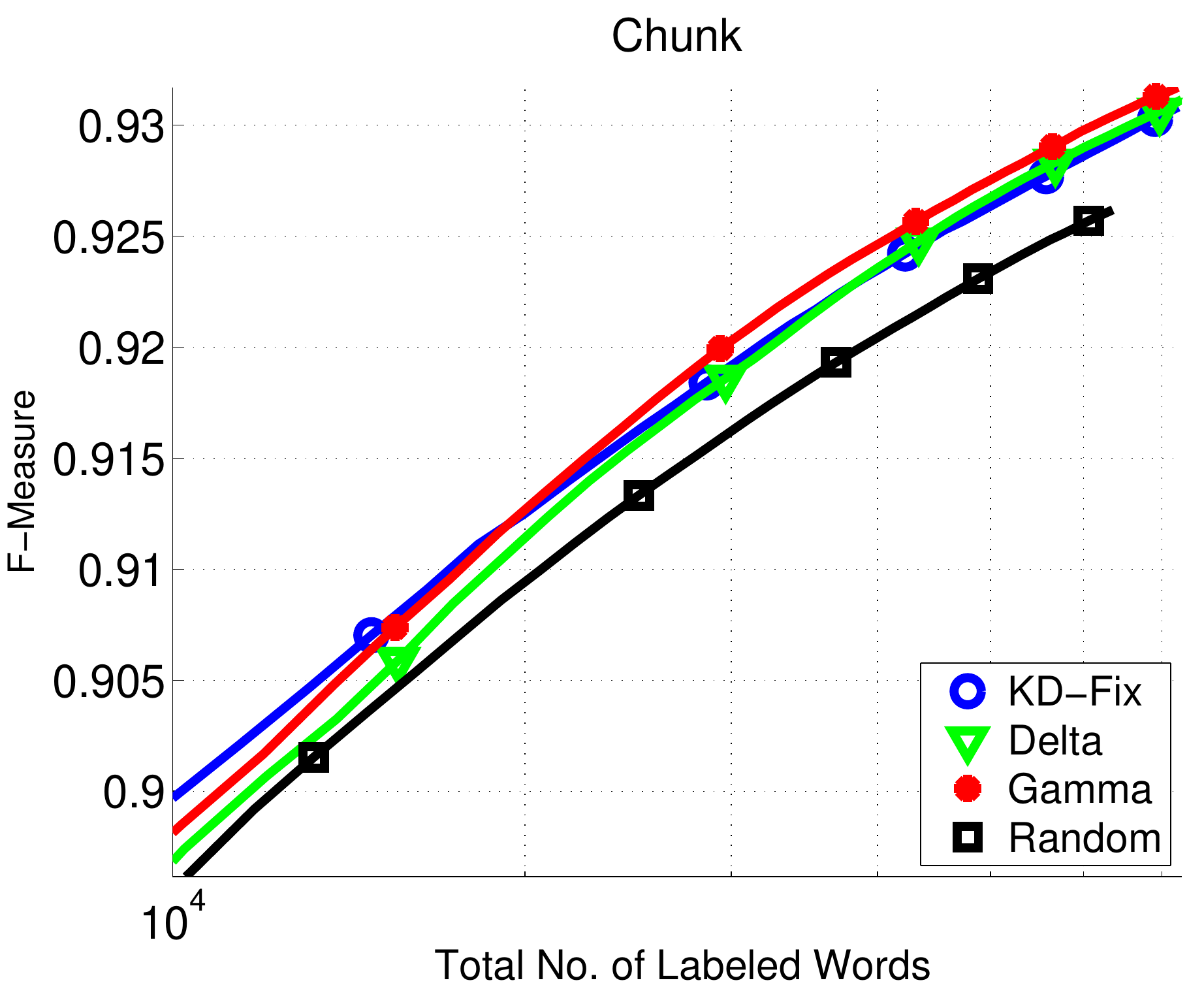}}
\end{tabular}
\caption{Averaged cumulative F-score vs. total number of words
  labeled. The left panels show the results for the first $10,000$ labeled
  words, the right panels show the results for more than
   $10k$ labeled words.}
\label{fig:sequences_active_learning_plots_CW}
\end{centering}
\vspace{-0.6cm}
\end{figure}

\begin{table}[!t!]
  \begin{center}
    {
      \begin{tabular}{|l|c|c|c|}
        \hline
        Dataset & Random  & {\tt KD-Fix}  & Effort reduction \tabularnewline
        \hline
        \hline
        NER English & 38K  & 25K  & 34\% \tabularnewline
        \hline
        NER Spanish & 80K  & 60K  & 25\% \tabularnewline
        \hline
        NER Dutch & 35K  & 23K  & 34\% \tabularnewline
        \hline
        NP chunking & 62K  & 43K  & 30\% \tabularnewline
        \hline
      \end{tabular}
      \label{table:sequences_active_learning_effort_reduction}
      \caption{Effort reduction in number of required word annotations when
        using active learning based on {\tt KD-Fix} confidence scores compared to random sentence annotation.}
}
\end{center}
\end{table}

In active learning, algorithms choose which example would be labeled
and be included in the training set.  The rational is that the
algorithm would choose to label examples which their label
contribute the most to the learning process. Other examples, often
the easy ones, will not be chosen for labeling as they are not adding
to the learning process.

In a typical active learning setting there exists a large set of
unlabeled data and a small set of labeled data.  Learning is performed
in iterations. On each iteration the current set of labeled data used
to build a model, which is then used to choose a subset of examples to
be labeled from remaining unlabeled examples. Many active learning
algorithms differ in the way this subset is chosen. We use the
confidence estimation methods for this purpose.

The experimental protocol we used is as follows. The initial set of
labeled examples contains $50$ annotated sentences, and the initial
set of unlabeled examples contains $9K$ sentences. Evaluation is
performed using an additional test set of $3K$ sentences. Learning is
performed using CW and then the resulting model and the confidence
methods are applied to the set of remaining unlabeled sentences,
yielding a rank over these sentences. Indeed, in standard binary or
even multiclass prediction, many active learning algorithms are making
a prediction for each and every example that is not labeled yet. For
sequence prediction problems, labeling sentences is time
consuming. Thus, on each iteration the algorithm first selects random
$1K$ sentences and a prediction (and confidence estimation) is
performed only to this set. The algorithm then picks the $10$
sentences with the least confident value, which are then annotated and
accumulated to the set of labeled data examples. Every such $10$
iterations (that is adding $100$ sentences) the performance of the
resulting model is evaluated using the evaluation set. The process is
repeated until $5K$ sentences are labeled.

The experiments reported above were focused in estimating confidence
for {\em individual words}. If for example, the output is given to a
human, it is realistic to ask the human to verify the tagging of
individual words as indeed most words are labeled correctly by the
model. When moving to active learning, the situation changes. Now, all
the words in a sentence are not labeled, thus if a human is required
to label a single word, she may need to label adjacent words as well,
having additional effort. We thus focused in active learning in the
sentence level rather than the word level. We defined the confidence
in a prediction of a sentence to be the minimal confidence score over
words in that sentence. Then the algorithm is ranking {\em sentences}
according to their confidence, breaking ties by favoring short
sentences (assuming short sentences contains a larger fraction of
informative words to be labeled than long sentences).

We evaluated scoring sentences using the confidence score of few
methods, {\tt KD-Fix}, {\tt KD-PC}, {\tt Delta} and {\tt
  Gamma}\footnote{We~\cite{MejerCr10} used two additional methods
  before, yet their performance was lower or about the same, and thus we omit them here for
  clarity.}. As a baseline we used random sentences selection.

The averaged cumulative f-measure vs. number of words labeled is
presented in \figref{fig:sequences_active_learning_plots_CW} (as above
{\tt KD-PC} curves are close to {\tt KD-Fix} curves and omitted for
clarity). Left panels summarizes the results for a short horizon
(small number of total-words). In two datasets random selection has
the lowest performance, while {\tt Gamma} is worse in
Spanish.  There is not clear winner among the other methods: {\tt
  KD-Fix} is the best for NP-Chunking while {\tt Gamma} is
best for Dutch. The right panels show the results for more than $10k$
training words, in log scale. In this scale, random selection performs
the worst in all cases, and then again all the other
methods performs about the same, except NER English where {\tt
  Gamma} achieves lower performance.

As the goal of active learning is reducing the annotation effort, we
compare the number of words required to be annotated when applying the
proposed selection method using {\tt KD-Fix} in order to achieve the
same performance level as by random picking of $5K$ sentences. The
results are summarized in
Table~12 
For NER English dataset only $25K$ words are required to achieve the same
level of performance obtained by training on $38K$ random words - a
$34\%$ effort reduction. Similarly for Spanish, Dutch and Chunking
about $12-20K$ words have to be annotated to achieve the same
performance of random labeling, a $25-34\%$ effort reduction.

\section{Related Work}

Confidence estimation methods for structured predication in NLP were
investigated in previously and applied to tasks such as POS tagging,
information extraction, machine translation, and automatic speech
recognition (ASR).

\namecite{1614012} propose several methods for
confidence estimation for single and multi-word fields, and for entire
records in a CRF based information extraction system. They describe a constrained
forward-backwards algorithm that defines confidence as the ratio of
total probability of any labeling for a given sentence with the total
probability given a labeling constraint which is taken as the
extracted field. The constrained forward-backwards yields same
confidence scores as {\tt Gamma} a for single token yet when extended
to multi-word field it performed better. Additionally, they used an
external maximum entropy classifier to classify fields as correct or
incorrect based of set of features describing the extracted fields.
The resulting posterior probability of the "correct" label is used as
the confidence measure. This method performed equally to the
constrained forward-backwards method. \namecite{1597216} shows how to improve interactive information
extraction system accuracy by high-lighting low confidence fields,
computed using these confidence scores, for the user to inspect and
correct.

\namecite{Scheffer2001} estimate confidence of
single token label in HMM based information extraction system by a
method similar to the {\tt Delta} method we use. The label confidence
score is the difference in the marginal probabilities of the best and
second best label to the token. They use the single token confidence
score to rank the labels and use this ranking for active learning.

\namecite{ChangxueMa2001} describe HMM based ASR system where word marginal score and difference in scores between the two best alternatives are used to compute per-word confidence measures, similar to {\tt Gamma} and {\tt Delta} methods we use. They train SVM based on these features for deciding which word candidates to reject.

\namecite{1245137} propose several methods for word level
confidence estimation for the task of machine translation based on
K-best translations. These methods are similar to the weighted and
non-weighted K-best methods ({\tt KB} and {\tt WKB}) we use. They show
the utility of the confidence scores for detecting incorrect
translated words and for re-scoring candidate translations among the
K-best alternatives and show improvement in translation quality.

\namecite{Gandrabur:2006:Cen:1177055.1177057}
describe using a neural network as
a dedicated confidence estimation layer and evaluate it for ASR and
for machine translation tasks. Both the ASR and translation systems
provide a probabilistic score to their outputs, yet the dedicated
confidence estimation layer can combine features and account for
addition information that is not available to the prediction system
and thus provide more useful confidence scores.  They show the
advantage of using a dedicated layer in ASR for the single word level and
in the concept-level (semantic phrases) by rejecting incorrect
recognitions. In context of interactive machine translation system
with sequences of 1-4 words they gain advantage by avoiding incorrect
translation suggestions.

About a decade ago, \namecite{Argamon-Engelson99committee-basedsample}
used committee voting to evaluate the uncertainty in an example in a
HMM based POS tagger. The committee of HMM taggers are sampled from
the trained model in a manner similar to {\tt KD-PC} method we
use. Each HMM parameter is independently sampled from normal
distribution with mean value equal to the trained parameter value and
with per-parameter standard deviation according to the number of
examples the parameter value was estimated by and multiplied by a
"temperature" coefficient.  The committee of taggers were then used to
produce a set of POS predictions and the uncertainty in each label is
computed according to the disagreement among the committee predictions
regarding that label.  They used the uncertainty measure for guiding
sample selection (active learning) and show that indeed this method
reduces the required labeling effort compared to random or entire
corpus labeling. However they did not evaluate directly the relative
or absolute per-label confidence score for purpose such as detecting
incorrect labels.

Confidence estimation was used in various additional setting of NLP applications such as binary and multi-class classification. For example,
\namecite{DelanyCDZ05} describe confidence estimation method in an email spam filtering system based on combination of k-Nearest Neighbors features such as agreement among the neighbors and distance to closest same and opposite prediction.
\namecite{Platt99} describes how to fit a sigmoid function to compute class posterior probabilities, that can serve as confidence scores, from the non-calibrated output score provided by the SVM in text classification tasks.
\namecite{XuLMMW2002} use confidence estimation in a question answering system by mixture of few heuristic correctness indicators and show that these confidence scores are useful for selecting between alternative answer sources.
\namecite{Bennett02} use confidence scores, referred to as reliability-indicators, for combining predictions of different types of document classifiers. They show that confidence scores allows better aggregation of the prediction votes compared to other alternatives.

Finally, there has been much work on active learning for NLP
applications including structured prediction tasks. For example
\namecite{tcm-alnlpie-99} use minimal confidence in the extraction
rules of a rules-based information extraction system to select
instances for annotation. \namecite{conf/acl/ShenZSZT04} use minimal
margin of SVM along with considerations of instances
representativeness and diversity in a NER system, and
\namecite{Baldridge2004} (see also \cite{Osborne2004}) apply active learning for
parsers.

\section{Summary and Conclusions}

We have studied few methods to estimate the \emph{per-word} confidence in structured predictions. These algorithms were evaluated both in a
relative settings and in a absolute setting. We also used the confidence score of these methods in two applications: increasing
precisions while decreasing recall, and active learning. All-in-all we
found that the two methods using sampling alternative models {\tt
  KD-PC} and {\tt KD-Fix} yield the best results. The former is induced from the CW
algorithms, while the later can be used with any linear model.
Generally {\tt KD-Fix} performed a little better except for sequential labeling in absolute setting where {\tt KD-PC} was better.

These methods were also found useful in increasing precision vs recall
in sequence labeling tasks.
Additionally, we showed that when combining these methods with active learning, one can reduce
annotating effort by at least $25\%$ compared to random sampling.
Understanding the theoretical properties of the problem and our
algorithms remains an open problem to be addressed in the future.

\section*{acknowledgments}
  This research was supported in part by the Israeli Science
  Foundation grant ISF-1567/10. Koby Crammer is a Horev Fellow, supported by
  the Taub Foundations.

\clearpage
\bibliographystyle{fullname}
\bibliography{bib}

\end{document}